%% file: main.tex
\RequirePackage{fix-cm}
\documentclass[smallcondensed]{svjour3}     
\smartqed  
\usepackage{amsfonts}
\usepackage{amsmath}

\usepackage{amsthm}

\usepackage{bbm}
\usepackage{comment}
\usepackage{graphicx}
\usepackage{tikz}
\usepackage{algorithmicx,algorithm}
\usepackage[noend]{algpseudocode}
\usepackage[caption=false]{subfig}

\usepackage[utf8]{inputenc}
\usepackage[english]{babel}

\usepackage[round]{natbib}
\usepackage[toc,page]{appendix}
\usepackage{mathtools}

\usepackage{marvosym}
\usepackage{ifsym}
\newcommand{\assign}{\leftarrow}

\newtheorem{prop}{Proposition}

\begin{document}

\title{Unsupervised Discretization by Two-dimensional MDL-based Histogram}

\author{Lincen Yang \and Mitra Baratchi \and
        Matthijs van Leeuwen 
}
\authorrunning{L. Yang et al.} 

\institute{Lincen Yang \Letter \at
              Leiden Institute of Advanced Computer Science, Leiden University \\
              \email{l.yang@liacs.leidenuniv.nl}           
           \and
           Mitra Baratchi \at
          Leiden Institute of Advanced Computer Science, Leiden University \\
          \email{m.baratchi@liacs.leidenuniv.nl}
          \and
          Matthijs van Leeuwen \at 
          Leiden Institute of Advanced Computer Science, Leiden University\\
          \email{m.van.leeuwen@liacs.leidenuniv.nl}
}
\date{Received: date / Accepted: date}
\maketitle
\begin{abstract}
Unsupervised discretization is a crucial step in many knowledge discovery tasks. The state-of-the-art method for one-dimensional data infers locally adaptive histograms using the minimum description length (MDL) principle, but the multi-dimensional case is far less studied: current methods consider the dimensions one at a time (if not independently), which result in discretizations based on rectangular cells of adaptive size. Unfortunately, this approach is unable to adequately characterize dependencies among dimensions and/or results in discretizations consisting of more cells (or bins) than is desirable.

To address this problem, we propose an expressive model class that allows for far more flexible partitions of two-dimensional data. We extend the state of the art for the one-dimensional case to obtain a model selection problem based on the normalized maximum likelihood, a form of refined MDL. As the flexibility of our model class comes at the cost of a vast search space, we introduce a heuristic algorithm, named PALM, which \underline{p}artitions each dimension \underline{al}ternately and then \underline{m}erges neighboring regions, all using the MDL principle. Experiments on synthetic data show that PALM 1) accurately reveals ground truth partitions that are within the model class (i.e., the search space), given a large enough sample size; 2) approximates well a wide range of partitions outside the model class; 3) converges, in contrast to the state-of-the-art multivariate discretization method IPD. Finally, we apply our algorithm to three spatial datasets, and we demonstrate that, compared to kernel density estimation (KDE), our algorithm not only reveals more detailed density changes, but also fits unseen data better, as measured by the log-likelihood. 
\end{abstract}



\section{Introduction}
\label{sec:intro}

Discretization, i.e., the transformation of continuous variables into discrete ones, is part of numerous data analysis workflows, making it a crucial step for a wide variety of applications in knowledge discovery and predictive modeling. However, many different discretization methods exist and it is often not easy to determine which method should be used. As a result, na\"ive methods such as equal-length and equal-frequency binning are still widely used, often with the number of bins chosen more or less arbitrarily, which can lead to suboptimal discretization.

A good discretization strikes a balance between the amount of preserved information and the complexity of the representation of the discretized data, so as to avoid discretizations that are either too coarse---resulting in severe loss of information---or too fine-grained---resulting in a bin per data point in the extreme case.

Achieving an optimal balance has been thoroughly studied for supervised discretization, i.e., discretization using additional information from a target variable. Optimal discretizations have been formalized using 1) statistical quantities, e.g., Pearson's chi-square \citep{boulle2004khiops}, 2) information-theoretic scores based on entropy or the minimum description length (MDL) principle \citep{jin2009data, fayyad1993multi}, and 3) Bayesian approaches \citep{boulle2006modl}.

In contrast, unsupervised discretization, which does not assume a target variable, has long been understudied \citep{kotsiantis2006discretization}. It serves a different purpose: supervised discretization aims to reduce the loss of information about the distribution of the target variable conditioned on the features \citep{boulle2004khiops, fayyad1993multi, kerber1992chimerge}, whereas unsupervised discretization aims to preserve information about the probability distribution of the variable to be discretized \citep{schmidberger2005unsupervised, biba2007unsupervised}.

This makes histograms well-suited to unsupervised discretization, and particularly adaptive histograms. An adaptive histogram is a probabilistic model that approximates probability density by piecewise constant densities, partitioning the data into bins such that 1) the probability density within each bin is approximately uniform (otherwise finer bins are needed), and 2) probability densities of neighboring bins are significantly different (otherwise they should be merged). 
\citet{kontkanen2007mdl} formalized this goal for one-dimensional adaptive histograms based on the minimum description length (MDL) principle \citep{rissanen1978modeling}, which is now considered to be the state-of-the-art univariate discretization method \citep{kameya2011time, nguyen2014unsupervised, marx2021estimating}.

The MDL principle \citep{rissanen1978modeling, grunwald2019minimum} is arguably one of the best off-the-shelf approaches for model selection tasks such as selecting a histogram model for given data, as it provides a means to naturally trade-off goodness-of-fit with model complexity. It achieves this by defining the ``best" probabilistic model for given data as the model that results in the best \emph{compression} of data and model together, which has been widely used in data mining and machine learning tasks \citep{galbrun2020minimum}.

\smallskip
\noindent \emph{Flexible multi-dimensional discretization}.
Traditional discretization methods are defined for one-dimensional (or univariate) data, and multi-dimensional (or multivariate) data is typically discretized by separately and independently discretizing each dimension, which ignores any dependencies between the dimensions. Multivariate discretization methods aim to take such dependencies into account, but they suffer from two problems. First, most methods focus on supervised discretization \citep{ferrandiz2005multivariate, bay2001multivariate, kwedlo1999evolutionary, kurgan2004caim}. Second, existing methods produce an adaptive grid based on the Cartesian product of the discretization results of individual dimensions. This approach ignores that the density of one dimension may change more drastically for certain values of another dimension; hence, appropriate binning of one dimension may depend on the values of the other dimensions. 

\begin{figure*}[tb]
\centering
	\begin{tabular}{c c c c}
		\includegraphics[width = 0.22\textwidth, height=3cm]{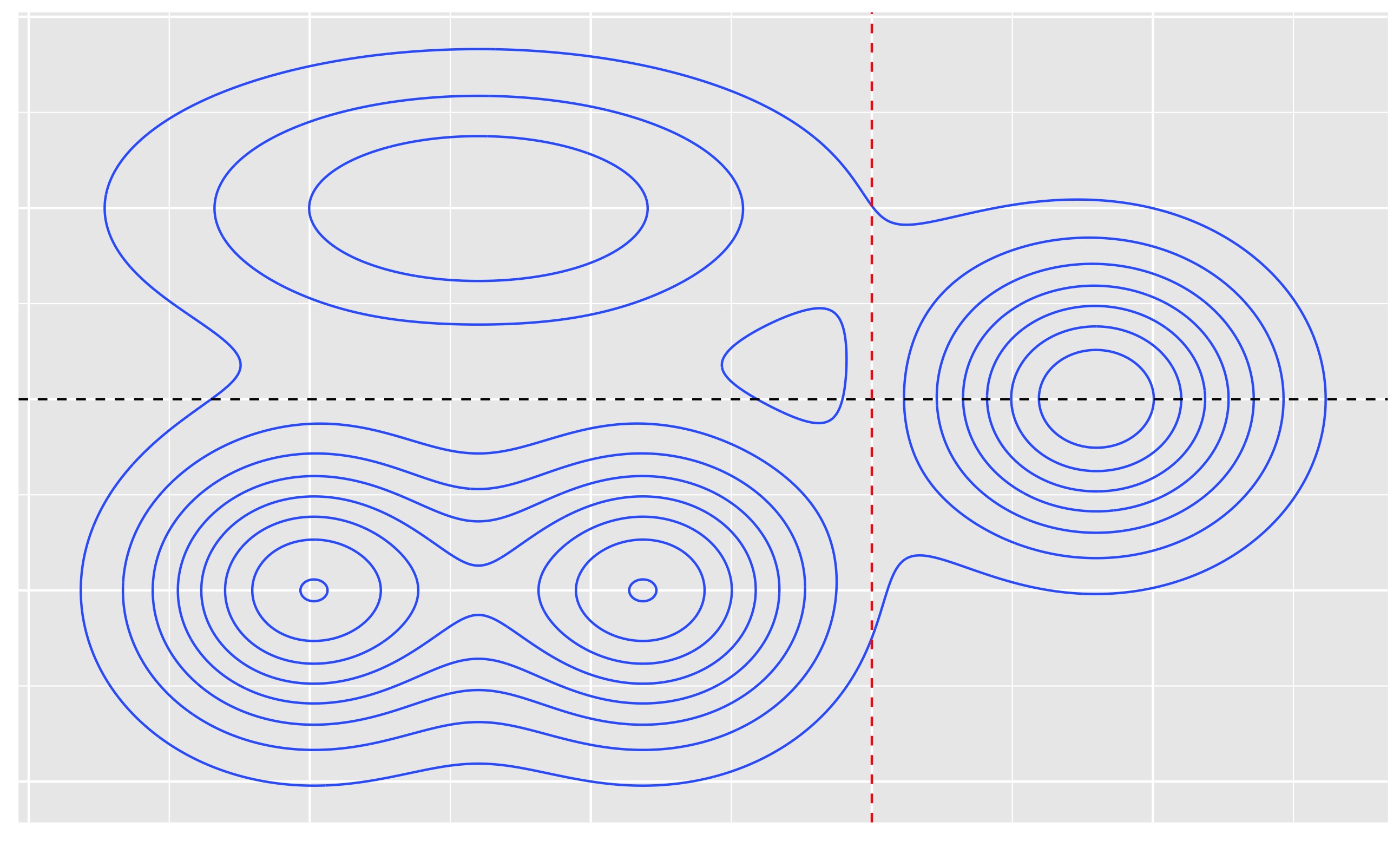} & \includegraphics[width = 0.22\textwidth, height=3cm]{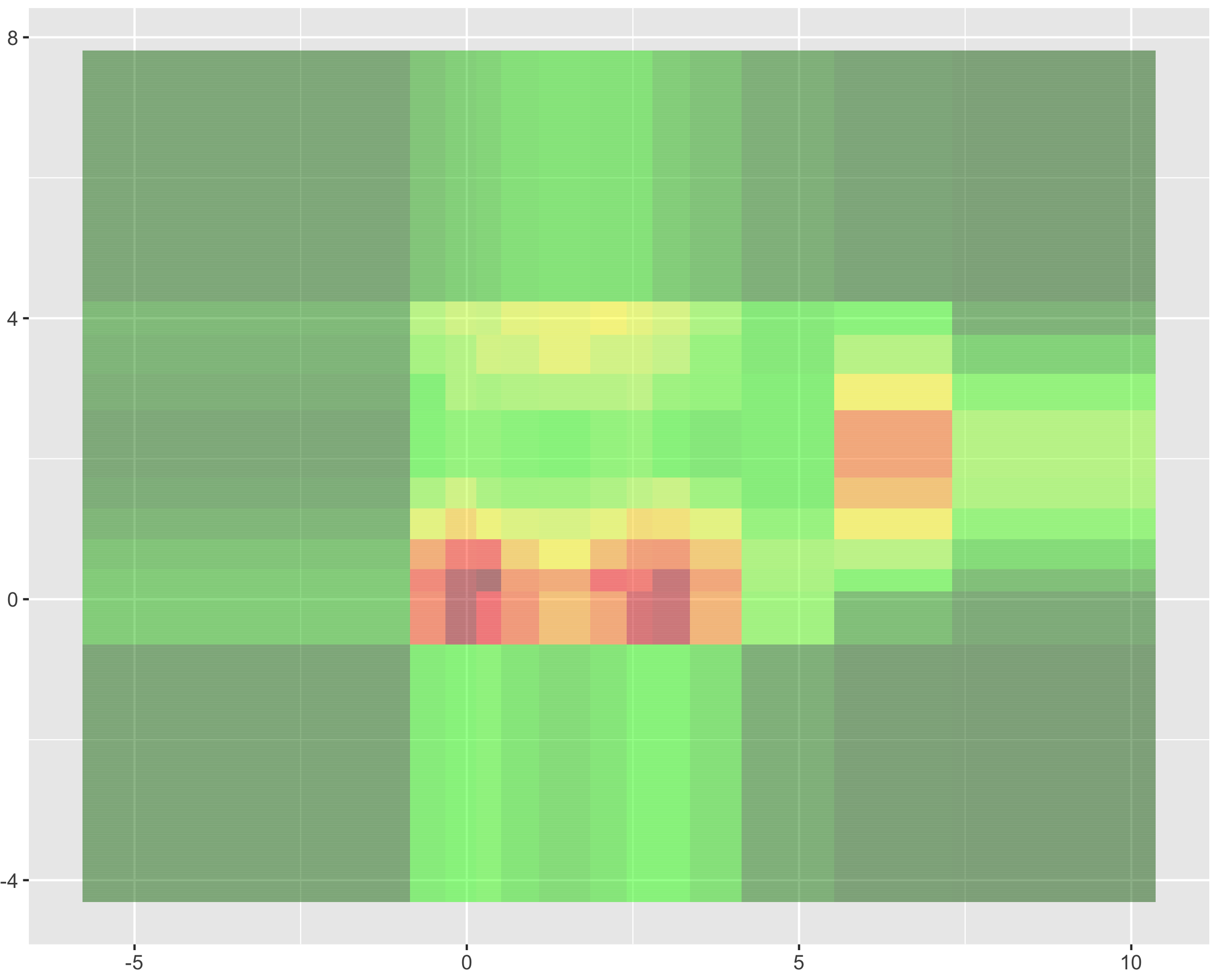} & 
		\includegraphics[width = 0.22\textwidth, height=3cm]{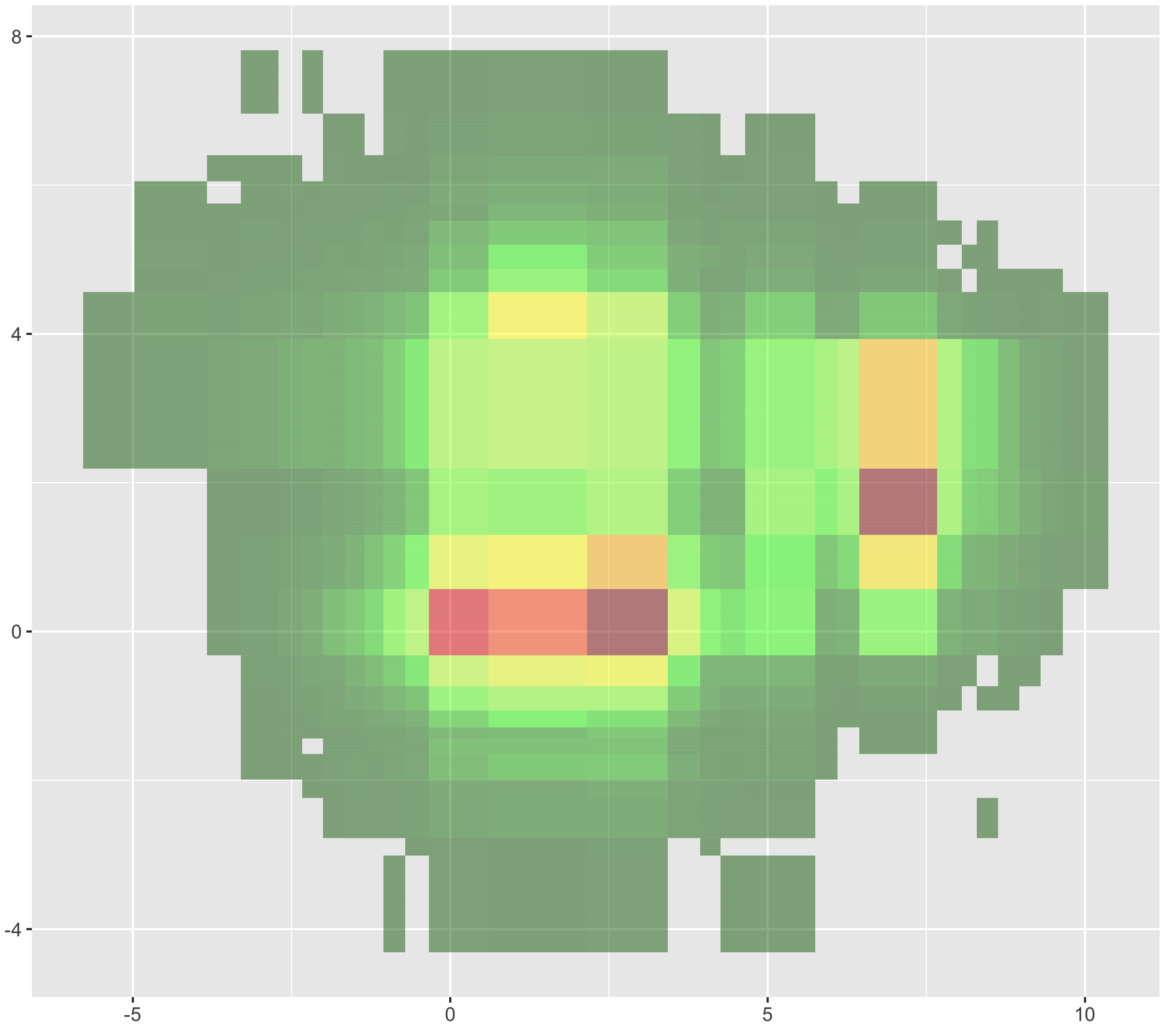} & \includegraphics[width = 0.22\textwidth, height=3cm]{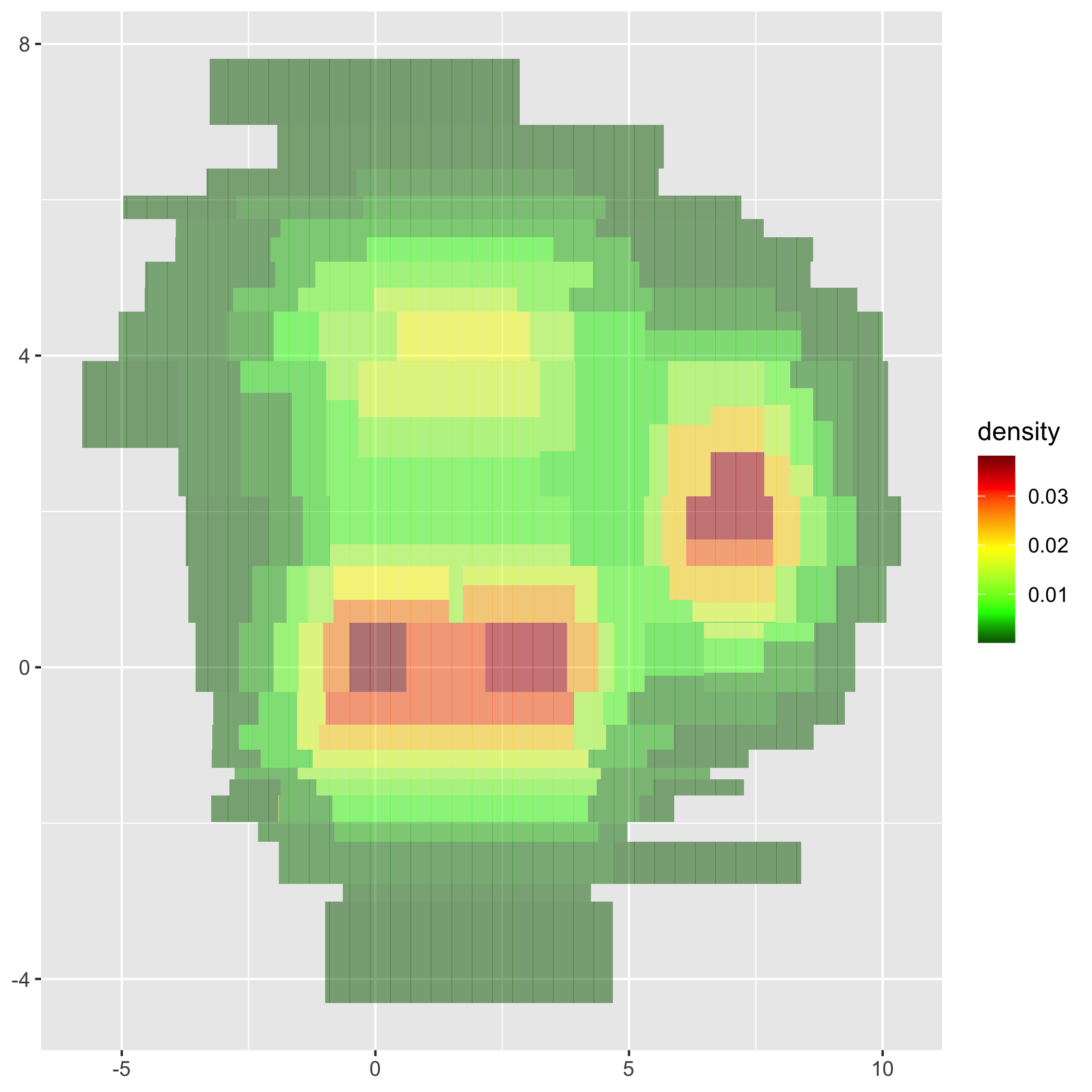}
	\end{tabular}
\caption{Distributions of a two-dimensional dataset simulated from a mixture of Gaussian distributions; from left to right: 1) true probability density contour, 2) partitioning by IPD \citep{nguyen2014unsupervised}, 3) partitioning by separately discretizing each dimension with the MDL histogram \citep{kontkanen2007mdl}, 4)  flexible partitioning by PALM, our algorithm.}	
	\label{fig:intro}
\end{figure*}

For instance, consider a two-dimensional synthetic dataset sampled from a mixture of Gaussians as shown in Figure~\ref{fig:intro} (leftmost)\footnote{For reproducibility, the the data is generated by the mixture of $N[\begin{psmallmatrix}0\\ 1\end{psmallmatrix}, \begin{psmallmatrix}1 & 0\\ 0 & 1\end{psmallmatrix}]$, $N[\begin{psmallmatrix}1.5\\ 4\end{psmallmatrix},\begin{psmallmatrix}2 & 0\\ 0 & 1\end{psmallmatrix}]$, $N[\begin{psmallmatrix}3\\ 0\end{psmallmatrix},\begin{psmallmatrix}1 & 0\\ 0 & 1\end{psmallmatrix}]$, $N[\begin{psmallmatrix}7\\ 2\end{psmallmatrix},\begin{psmallmatrix}1 & 0\\ 0 & 1\end{psmallmatrix}]$ with all mixing coefficients 0.25; sample size is $40 \, 000$.}. To adequately discretize data from this distribution, the binning of the x-axis should be different depending on whether $y$ is above or below the black dashed line, in order to capture the different density changes for the Gaussian distribution (above) and the Gaussian mixture (below). Similarly, the binning of the y-axis should be different depending on whether $x$ is left or right to the red dashed line. This motivates us to consider partitions that are more flexible than adaptive grids: we consider all partitions that can be obtained by clustering the ``cells" of a fine-grained fixed-grid. The remaining three plots in Figure~\ref{fig:intro} show the density plots obtained by 1) IPD \citep{nguyen2014unsupervised}, the state-of-the-art multivariate unsupervised discretization method, 2) the one-dimensional MDL-based histogram method \citep{kontkanen1997bayesian} applied independently on each dimension, and 3) our method. Our method produces the density estimation that most resembles the shape of the original contour, as we allow the bins of one dimension to depend on the value of another dimension. 

\smallskip
\noindent \emph{Approach and contributions}. 
We consider the problem of learning two-dimensional histogram models that enable far more flexible partitions than regular adaptive grids. That is, we allow any partition that can be obtained by iteratively merging adjacent cells of a fixed grid, which allows for learning models that provide accurate density estimates while not having more bins than strictly necessary (thereby avoiding overfitting and providing clear region boundaries, i.e., adjacent bins must have different density estimates).

We formalize the two-dimensional histogram construction problem as a model selection task using the MDL principle. For this we build on the one-dimensional MDL-based histogram selection problem as introduced in the seminal work by \citet{kontkanen2007mdl}, because it is both theoretically elegant and practically fast. Specifically, it adopts the \emph{normalized maximum likelihood} (NML) encoding scheme, a form of \emph{refined MDL} \citep{grunwald2007minimum, grunwald2019minimum} that provides minimax regret, and employs a fast dynamic programming algorithm to find the optimal solution. 

The existing approach for one-dimensional histograms cannot be trivially extended to multiple dimensions though, hence we make a number of technical contributions. 

First, we solve the challenge of computing the so-called \emph{parametric complexity} \citep{grunwald2019minimum} for the multi-dimensional case. 

Second, we observe that efficiently finding the MDL-optimal two-dimensional histogram is infeasible and propose PALM, a heuristic algorithm for learning two-dimensional histograms. PALM combines top-down (partition) and bottom-up (merge) search strategies by 1) first partitioning the data by iteratively splitting regions, and 2) then iteratively merging neighboring regions if their densities are similar. In each step, the MDL principle is used as decision criterion; as a result, our algorithm requires neither hyper-parameters\footnote{The precision with which the data is recorded can be used to set the granularity of the initial base grid.} nor any pre-defined stopping criterion to be specified. It automatically adapts to both local density structure, as shown in the example in Figure~\ref{fig:intro} and, later, in Sections~\ref{sec::exp} and \ref{sec:case}. 

Third, we make several improvements to the dynamic programming algorithm used for the one-dimensional MDL histogram, which we use as a building block for our algorithm. Specifically, as described in Section~\ref{sec:prelim}, we 1) correct a minor theoretic flaw related to computing the code length that is needed to encode the histogram model, and 2) reduce the time complexity by simplifying the dynamic programming recursion. 

We perform extensive experiments to show that our algorithm 1) accurately recovers ground truth histograms, 2) approximates well ground truth partitions that are not within the model class, and 3) outperforms IPD \citep{nguyen2014unsupervised}, the state-of-the-art algorithm for unsupervised multi-dimensional discretization. Further, case studies on spatial data show that, compared to kernel density estimation (KDE), our algorithm not only reveals more detailed density changes, but also fits unseen data better, as measured by the log-likelihood.

We restrict the scope of this paper to two-dimensional data for three reasons. First, two-dimensional discretization methods have many potential applications in the domain of spatial data analysis, e.g., using GPS data, where ad-hoc discretization methods are still widely used (Cao et al., 2014). The case studies demonstrate that our method can successfully reveal interesting patterns from GPS data. Second, as our approach uses more flexible partitions than adaptive grids, the search space is very large even for two-dimensional data. Our algorithm for the two-dimensional case should be regarded as a step towards solving the algorithmic challenge for higher dimensions, but does not solve it completely. Third, focusing on the two-dimensional case allows us to more easily examine the results empirically, e.g., to verify desired properties such as adaptivity to sample size and local density structure. 

\section{Related work}
\label{sec:related}

We briefly review previous work concerning discretization methods, histogram models, and tree-based models for density estimation.

\paragraph{Unsupervised univariate discretization.} 
Most unsupervised univariate discretization methods are rather straightforward and concern equal-width binning, equal-frequency binning, which in practice usually involve ad-hoc choices for the number of bins or the number of frequency in each bin. 

 Clustering techniques such as k-means \citep{friedman2001elements} or Bayesian clustering \citep{kontkanen1997bayesian} are also used in discretization; however, they ignore the possible heterogeneity within the cluster and choices of hyper-parameters are usually required. 
 
More advanced criteria rely on density estimation and specifically constructing adaptive histograms. Apart from the MDL-based histogram \citep{kontkanen2007mdl} already mentioned in Section \ref{sec:intro}, \citet{schmidberger2005unsupervised} proposed to construct adaptive histograms by recursive binary partition with cross-validation. A local heuristic is used to decide the cut point, and cross-validation is used to choose the number of intervals; in contrast, the MDL-based histogram \citep{kontkanen2007mdl} uses a global score with a dynamic algorithm that optimizes the cut points and the number of bins simultaneously. Moreover, an adaptive histogram can also be selected as the one whose density estimation result is closest to the result of \emph{kernel density estimation} \citep{biba2007unsupervised}, where cross-validation is used to prevent overfitting. As the true density is apparently not known, cross-validation is performed by Monte Carlo sampling-based methods. However, cross-validation is known to be computationally expensive, and the influence of choosing different kernels on discretization is not reported. 

Bayesian approaches have been widely used in adaptive histograms \citep{scricciolo2007rates, liu2014multivariate, van2017bayesian, gasparini1996bayesian, lu2013multivariate}. These methods treat all possible histograms as the model class and put a prior distribution on it, and the resulting posterior distribution is directly used for density estimation (by calculating the marginal distribution). Therefore, although these Bayesian approaches often provide theoretic guarantees as density estimation methods, they do not provide an individual adaptive histogram that can be used for discretization. 

\paragraph{Unsupervised multivariate discretization.}
Since discretizing each dimension of multivariate data independently will ignore the dependencies among different dimensions, some methods attempt to reduce the dependencies by PCA- or ICA-based methods \citep{mehta2005toward, kang2006ica}\footnote{Note that the ICA-based method \citep{kang2006ica} is designed for  supervised discretization, but we noticed that the ICA transformation there is not restricted to supervised discretization only.}. However, as both methods are based on \emph{linear transformation} of the random vector, they may fail to eliminate nonlinear dependencies. Note that extending these methods to nonlinear PCA or nonlinear ICA may not be suitable for unsupervised discretization tasks, as the uniform distribution is not invariant under nonlinear transformation, and hence we cannot obtain an adaptive histogram of the original data by inversely transforming the adaptive histogram constructed on the nonlinearly transformed data.

\citet{lud2000relative} proposed the so-called ``relative unsupervised discretization". The core of this method is to perform clustering on an individual dimension, using different subsets of values. These different subsets are obtained by filtering the dataset using other dimensions, in order to keep the dependency among different dimensions. However, this method does not control the information loss about the probability distribution of the dimension that is to be discretized.

Further, methods trying to optimize the discretization of all dimensions simultaneously exist. One approach is to start from a very fine grid, and merge neighboring subintervals for each dimension if the multivariate probabilities of the data within these two consecutive subintervals are similar \citep{nguyen2014unsupervised, bay2001multivariate}. These methods are based on certain choices of similarity metrics, and require explicit specification of the similarity threshold. We empirically show in Section~\ref{sec::exp} that IPD, the method by \citet{nguyen2014unsupervised} that is also based on the MDL principle and is considered the state-of-the-art multivariate discretization method, does not converge in practice. 

Finally, Kameya extended the the one-dimensional MDL-histogram \citep{kameya2011time} specifically for time series data, who proposed to discretize time series data by iteratively adjusting the cut points on each dimension until convergence, using the coordinate descent optimization approach.

All these multivariate discretization methods all try to optimize the adaptive grid and produce (hyper)rectangular regions. Our method, in contrast, is proposed to produce far more flexible segmentation, which allows the binning of one dimension to be dependent on the values of other dimensions.  

\paragraph{Density estimation tree.}
Algorithmically, our method is very similar to methods using tree models for density estimation \citep{ram2011density, liu2014multivariate, yang2014density}, as partitioning the data space by iteratively partitioning each dimension is identical to growing a tree. However, these density estimation trees were developed by adapting the scores used in growing, stopping, and pruning (supervised) decision and regression trees. That is, while our algorithm employs a consistent MDL-based framework for selecting the best model, these density estimation trees use separate optimization scores respectively to fit the model and to control the model complexity, often with user-specified hyper-parameters and/or computationally expensive cross-validation. 

Moreover, these density estimation trees, as is like most supervised tree models, only do binary partitioning in a greedy manner. On the contrary, our method can split a dimension into multiple bins (from 1 to a pre-determined $K_{max}$) instead of just two, which is not only more flexible, but also more interpretable, as after partitioning on a certain dimension, within each bin the data points on that dimension can be regarded as approximately uniform.

Finally, our method has an additional merging step, which creates much more flexible partitions of data, resulting in models that are more informative for pattern mining and exploratory data analysis. 

\paragraph{Supervised discretization.}
When discretization is needed for a supervised task such as classification, we can use \emph{supervised discretization}, which means that the target variable is used to assess how much information on the target the discretization maintains. Several criteria can be put in this category, which are mostly based on statistical hypothesis testing or entropy, as summarized in the survey paper by \citet{kotsiantis2006discretization}. The MDL principle has also been used for supervised discretization \citep{fayyad1993multi, pfahringer1995compression, zhang2007discretization, ferrandiz2005multivariate, gupta2010clustering}, but all of them use the so-called \emph{crude MDL} principle \citep{grunwald2007minimum}, which is theoretically suboptimal.

\section{Problem Statement}
\label{sec:problem}

Informally, we consider the problem of inferring the best two-dimensional histogram for a given sample of continuous data. To make this problem precise, we start off by introducing our notation and definitions. Note that all $\log(\cdot)$ should be read as $\log_2(\cdot)$ unless specified otherwise.  

\subsection{Notation and definitions of data, model, and model class} \label{subsec:notations}

Consider as data a vector of length $n$, i.e., $x^n = (x_1, ..., x_n)$, sampled independently from a \emph{random variable} $X$. 

The \emph{sample space} of $X$, denoted as $S$, is a \emph{bounded} subset of $\mathbb{R}^2$. Although the sample space of a random variable, e.g., a Gaussian, can be infinite in theory, we always assume it to be a bounded ``box" when dealing with a given dataset. The task of estimating $S$ from the data directly is another research topic, usually referred to as ``support estimation" in statistical literature \citep{cuevas1997plug}, and hence is out of the scope of our main focus in this article.

Conceptually, a histogram---no matter whether it is one- or multi-dimensional---is a \emph{partition} of the sample space $S$, denoted by $\widetilde{S}$ and parametrized by a vector $\Vec{f} = (f_1, \ldots, f_K)$. A partition $\widetilde{S}$ is defined as a set of \emph{disjoint} subsets of $S$, and the union of all these subsets is $S$ itself, i.e., $\widetilde{S} = \{S_1, S_2, \ldots, S_K\}$, where $\forall j \in \{1,\ldots,K\}$, $S_j \subseteq S$, $\bigcup_{j = k}^K S_j  = S$, and $\forall j, k \in \{1,\ldots,K\}$, $S_j \cap S_k = \emptyset$. We also call these subsets, i.e., elements of $\widetilde{S}$, as \emph{regions}.

Next, we assume that the probability density of $X$, denoted by $f(X)$, is given by
\begin{equation}
    f(X) = \sum_{j \in \{1,\ldots,K\}}\mathbbm{1}_{S_j}(X) f_j,  \label{eq1}
\end{equation}
where $\mathbbm{1}_{\{\cdot\}}(\cdot)$ is the \emph{indicator function}. Each $f_j$ is a \emph{constant} and $\Vec{f}$ satisfies $\sum_{i=1}^K f_j|S_j| = 1$, where $|S_j|$ denotes the geometric area of $S_j$, i.e., when $X \in S_j$, $f(X) = f_j$.
We refer to any partition $\widetilde{S}$ as a \emph{histogram model} that contains a family of probability distributions; i.e., $\forall \Vec{f} \in \mathbb{R}^K$, we denote a single probability distribution by $\widetilde{S}_{\Vec{f}}$.

We denote the model class as $\mathbb{M}$, representing all possible partitions with $K$ regions that can be obtained by clustering cells of a fixed grid covering $S$, where $K \in \{1,\ldots, K_{max}\}$. The granularity of the grid, denoted as $\epsilon$, and $K_{max}$ are fixed in advance, but note that they can be set arbitrarily small and large, respectively. 

Geometrically, this is equivalent to drawing \emph{inner boundaries} within $S$ along the fixed grid. In practice, $\epsilon$ can represent the precision up to which the data is recorded or that is useful for the given task. Although the model class we consider only has inner boundaries consisting of \emph{line segments}, we will show that such a model class is flexible enough to approximate curved inner boundaries in Section~\ref{sec::exp}. 


\subsection{Histogram model selection by the MDL principle} \label{subsec:hist}
We now formally define the task of two-dimensional data discretization as an MDL-based model selection task, using histogram models as the model class.
 
The MDL principle is arguably one of the best off-the-shelf model selection methods and has been successfully applied to many machine learning tasks \citep{grunwald2007minimum, hansen2001model}. It has solid theoretical foundations in information theory and naturally prevents overfitting as the optimization criterion always includes the model complexity, defined as the code length (in bits) needed to encode that model \citep{grunwald2007minimum}. 

The basic idea is to \emph{losslessly} \emph{encode} the model and data together, by firstly encoding the model and then compressing the data using that model. The model resulting in the shortest total \emph{code length} is defined to be MDL-optimal, i.e.,
\begin{equation}
    \widetilde{S^*} = \arg \min_{\widetilde{S} \in \mathbb{M}} L(x^n, \widetilde{S}) 
     = \arg \min_{\widetilde{S} \in \mathbb{M}} (L(\widetilde{S}) + L(x^n|\widetilde{S})), \label{eq:mdl_score_general}
\end{equation}
where $L(\widetilde{S})$ and $L(x^n|\widetilde{S})$ are respectively the code length of the model and the code length of the data compressed by that model. Note that $L(\cdot|\cdot)$ denotes the \emph{conditional} code length \citep{grunwald2007minimum}; informally, $L(A|B)$ represents the code length of the message a \emph{decoder} needs to receive in order to be able to losslessly reconstruct message $A$ after having already received message $B$. 

We will show in Section \ref{sec: code} that properly encoding the model and calculating its corresponding code length $L(\widetilde{S})$ turns out to be very difficult. As a result, we unfortunately cannot regard our model selection task simply as an optimization problem.

To alleviate this, we divide the model selection task into two steps, namely 1) partitioning alternately and 2) merging. 
\begin{figure} 
	\begin{tikzpicture} [scale = 1.1] 
	\draw (0,0) -- (1,0) -- (1,1) -- (0,1) -- (0,0);
	
	\draw[->] (1.2,0.5) -- (1.8,0.5);
	
	\draw (2,0) -- (3,0) -- (3,1) -- (2,1) -- (2,0);
	\draw (2.3,0) -- (2.3,1);
	\draw (2.6,0) -- (2.6,1);
	
	\draw[->] (3.2,0.5) -- (3.8,0.5);
	
	\draw (4,0) -- (5,0) -- (5,1) -- (4,1) -- (4,0);
	\draw (4.3,0) -- (4.3,1);
	\draw (4.6,0) -- (4.6,1);
	\draw (4.0,0.3) -- (4.3,0.3);
	\draw (4.0,0.6) -- (4.3,0.6);
	
	\draw[->] (5.2,0.5) -- (5.8,0.5);
	
	\draw (6,0) -- (7,0) -- (7,1) -- (6,1) -- (6,0);
	\draw (6.3,0) -- (6.3,1);
	\draw (6.6,0) -- (6.6,1);
	\draw (6.0,0.3) -- (6.3,0.3);
	\draw (6.0,0.6) -- (6.3,0.6);
	\draw (6.3,0.5) -- (6.6,0.5);
	
	\draw[->] (7.2,0.5) -- (7.8,0.5);
	
	\draw (8,0) -- (9,0) -- (9,1) -- (8,1) -- (8,0);
	\draw (8.3,0) -- (8.3,1);
	\draw (8.6,0) -- (8.6,1);
	\draw (8.0,0.3) -- (8.3,0.3);
	\draw (8.0,0.6) -- (8.3,0.6);
	\draw (8.3,0.5) -- (8.6,0.5);
	\draw (8.5,0.5) -- (8.5,1);
	
	\draw[->] (9.2,0.5) -- (9.8,0.5);
	
	\draw (10,0) -- (11,0) -- (11,1) -- (10,1) -- (10,0);
	\draw (10.3,0) -- (10.3,0.3);
	\draw (10.3,0.5) -- (10.3,1);
	\draw (10.6,0) -- (10.6,0.5);
	\draw (10.0,0.3) -- (10.3,0.3);
	\draw (10.0,0.6) -- (10.3,0.6);
	\draw (10.3,0.5) -- (10.6,0.5);
	\draw (10.5,0.5) -- (10.5,1);
	
	\end{tikzpicture}
\caption{An illustration of the partitioning and merging steps. From left to right: alternatively partitioning each region until compression cannot be further improved, and finally merging some of the neighboring regions to further improve compression.}
\label{fig:illustrationPALM}
\end{figure}

First, we alternately split each region within partition $\widetilde{S}$ (initially $\widetilde{S} = \{S\}$) in one of the two dimensions, then update $\widetilde{S}$ accordingly, and repeat the process. In other words, in each iteration we further split each region within $\widetilde{S}$ in one dimension (i.e., horizontally or vertically), which is equivalent to selecting the best set of horizontal or vertical \emph{cut lines}. 

Denote the subset of data points within a certain region $S' \in \widetilde{S}$ as $\{x^n \in S'\}$. We formally define the task of selecting the set of MDL-optimal cut lines set as 
\begin{equation}
\begin{split}
    {C^*_{S'}} &= \arg \min_{{C_{S'}} \in \mathbb{C}_{S'}} L(\{x^n \in S'\}, {C_{S'}})  \\
&= \arg \min_{{C_{S'}} \in \mathbb{C}_{S'}} (L({C_{S'}}) + L(\{x^n \in S'\} | {C_{S'}})), \label{eq:score_split}
\end{split}
\end{equation}
where $\mathbb{C}_{S'}$ are all possible sets of cut lines, containing $K = \{0,1,\ldots,K_{max}\}$ cut lines, for the certain region $S' \in \widetilde{S}$ in one certain dimension (i.e., horizontal or vertical), and $K_{max}$ is predetermined a priori to be ``large enough" given the task at hand. 

In Section \ref{sec:prelim}, we will show that searching for the MDL-optimal cut lines for (a subset of) two-dimensional data is the same as searching for the MDL-optimal cut points for the one-dimensional data that is the projection of the two-dimensional data onto the x- or y-axis. 

The partitioning step will automatically stop once for each region the MDL-optimal set of cut lines is the null set, i.e., no further partitioning is needed. 

Second, we search for all possible clusterings of neighboring regions gained in the previous partitioning step, in a greedy manner. In other words, we consider all possible clustering of regions of the partition gained by the previous partitioning step, which is actually a subset of the full model class $\mathbb{M}$ as defined in Section \ref{subsec:notations}. We denoted this constrained model class by $\mathbb{M}_c$, and we formally define the merging step as selecting the MDL-optimal model within $\mathbb{M}_c$, i.e., 
\begin{equation}
    \widetilde{S}_{merge}^* = \arg \min_{\widetilde{S} \in \mathbb{M}_c} L(x^n, \widetilde{S}) 
     = \arg \min_{\widetilde{S} \in \mathbb{M}_c} (L(\widetilde{S}) + L(x^n|\widetilde{S})). \label{eq:score_merge}
\end{equation}
Figure \ref{fig:illustrationPALM} shows an illustrative example of the partitioning and merging process.




\section{Calculating the code length}
\label{sec: code}
We now discuss the details of the code length (in bits) needed to encode the data and the model. 

We first show the calculation of code length of data given a histogram model, encoded by the normalized maximum likelihood (NML) code \citep{grunwald2007minimum, grunwald2019minimum}. Specifically, we show that the \emph{parametric complexity} term in the code length is independent of data dimensionality, which is an important observation that makes it feasible to compute the NML code length.

Next, we discuss in detail the difficulties of encoding all possible models $\widetilde{S} \in \mathbb{M}$ if we would want to directly optimize over the full model class $\mathbb{M}$ using Equation (\ref{eq:mdl_score_general}), which motivates our (more pragmatic) solution of dividing the model selection task into two separate steps. 

Finally, we discuss the calculation of the code length of a model in the partitioning and merging step respectively, i.e., $L({C_{S'}})$ and $L(\widetilde{S})$ of Equations (\ref{eq:score_split}) and (\ref{eq:score_merge}).

\subsection{Code length of the data} 
\label{subsec:code_length_for_data}

Extending the work that was previously done for the one-dimensional case \citep{kontkanen2007mdl}, we use the same code---i.e., the \emph{Normalized Maximum Likelihood} (NML) code---to encode the two-dimensional data. This code has the desirable property that it is theoretically optimal because it has minimax regret. The code length of the NML code consists of two terms, namely the maximum likelihood and the parametric complexity (also referred to as \emph{regret}), and is given by
\begin{equation}
    L(x^n|\widetilde{S}) = -\log\left(\frac{P(x^n|\widetilde{S}_{\hat{\Vec{f}}(x^n) })}{\text{COMP}(n, \widetilde{S})}\right), \label{eq:NML_def}
\end{equation}
where $P(x^n|\widetilde{S}_{\hat{\Vec{f}}(x^n)})$ is the probability of the data given $\widetilde{S}_{\hat{\Vec{f}}(x^n)}$, i.e., the parameters $\Vec{f} = (f_1, ..., f_K)$ are estimated by the \emph{maximum likelihood estimator} given dataset $x^n$, denoted as $\hat{\Vec{f}}(x^n) = (\hat{f}_1, \ldots, \hat{f}_K)$. The term $\text{COMP}(n, \widetilde{S})$ is the so-called \emph{parametric complexity}, which is defined as 
\begin{equation}
    \text{COMP}(n, \widetilde{S}) = \sum_{y^n \in S^n} P(y^n|\widetilde{S}_{\hat{\Vec{f}}(y^n)}), \label{eq10}
\end{equation}
where $\sum_{y^n \in S^n}$ is the sum over \emph{all possible sequences $y^n$} within the Cartesian product of sample space $S$ that can be \emph{generated} by the histogram model $\widetilde{S}$, i.e., the order of individual values within vector $y^n$ \emph{does} matter.  

We will now first describe the calculation of $P(x^n|\widetilde{S}_{\hat{\Vec{f}}(x^n) })$, and then the calculation of $\text{COMP}(n, \widetilde{S})$. 

For any single data point $x_i \in x^n$, let 
$x_i = (x_{i1}, x_{i2})$ denote the pair of values for its two dimensions. We then have
\begin{equation}
    P(x^n|\widetilde{S}_{\hat{\Vec{f}}(x^n)}) = \prod_{i=1}^n P(x_i|\widetilde{S}_{\hat{\Vec{f}}(x^n)}) = \prod_{j = 1}^K \left(\prod_{x_i \in S_j} P(x_i|\widetilde{S}_{\hat{\Vec{f}}(x^n)})\right) ,\label{eq:data_prob_1}
\end{equation}
as the data points are assumed to be independent. Note that $K$ represents the number of regions of $\widetilde{S}$. 

Since we assume our data to have precision $\epsilon$, we can define the probability of the data, also referred to as its \emph{maximum likelihood}, as
\begin{equation}
    P(x_i|\widetilde{S}_{\hat{\Vec{f}}(x^n)}) = P(X \in [x_{i1} - \frac{\epsilon}{2}, x_{i1} + \frac{\epsilon}{2}] \times 
    [x_{i2} - \frac{\epsilon}{2}, x_{i2} + \frac{\epsilon}{2}] \,\, | \,\, \widetilde{S}_{\hat{\Vec{f}}(x^n)}) 
     = \hat{f_j} \epsilon^2 \text{.} \label{eq:data_prob_2}
\end{equation}
The maximum likelihood estimator for the histogram model \citep{scott2015multivariate} is 
\begin{equation}
    \hat{f}_j = \frac{h_j}{n \, |S_j|} \text{,} \,\, \forall j \text{,} \label{eq:ml_hist}
\end{equation}
where $h_j$ is the number of data points within $S_j$, and $|S_j|$ is the area of $S_j$. Thus, following Equations (\ref{eq:data_prob_1}),(\ref{eq:data_prob_2}), and (\ref{eq:ml_hist}),
\begin{equation}
    P(x^n|\widetilde{S}_{\hat{\Vec{f}}(x^n)}) = \prod_{j = 1}^K (\hat{f_j} \, \epsilon^2)^{h_j} = \prod_{j = 1}^K (\frac{h_j \, \epsilon^2}{n \, |S_j|})^{h_j}. \label{eq:mle_xn}
\end{equation}

Next, we describe the calculation of $\text{COMP}(n, \widetilde{S})$. Although it may be surprising at first glance, we show that
\begin{prop} \label{prop1}
	The parametric complexity $\text{COMP}(n, \widetilde{S})$ of a histogram model is a function of sample size $n$ and the number of bins $K$. Given $n$ and $K$, $\text{COMP}(n, \widetilde{S})$ is independent of the dimensionality of the data. 
\end{prop} 
We leave the formal proof to Appendix~A, but the proposition is based on the following important observations. First, as \citet{kontkanen2007mdl} proved, $\text{COMP}(n, \widetilde{S})$ is a function of sample size $n$ and the number of bins $K$ for one-dimensional histograms. The remaining question is whether this holds for two (and higher) dimensional histograms as well. Observe that the maximum likelihood given a two-dimensional histogram model for any data is a function of $h_j$ and $|S_j|/\epsilon^2$, respectively representing the number of data points in each region, and the total number of \emph{possible positions} of data points in each region, which are both some form of ``counts" and hence are ``dimensionality free". Finally, $\text{COMP}(n, \widetilde{S})$, as defined in Equation (\ref{eq10}), is just the sum of maximum likelihoods. Based on these observation, it is trivial to prove that $\text{COMP}(n, \widetilde{S})$ has the same form for one- and multi-dimensional histograms.

Therefore, for both one- and multi-dimensional histogram models, we can denote $\text{COMP}(n, \widetilde{S})$ as $\text{COMP}(n,K)$, and as shown by \citet{kontkanen2007mdl},
\begin{equation}
    \text{COMP}(n, K) = \sum_{h_1 + ... + h_K = n} \frac{n!}{h_1! ... h_K!} \prod_{j=1}^K (\frac{h_j}{n})^{h_j} \text{,} \label{eq:comp}
\end{equation}
which turns out to be the same as the parametric complexity for the multinomial model \citep{kontkanen2007linear}. We can calculate $\text{COMP}(n, K)$ in linear time \citep{kontkanen2007linear} by means of the following recursive formula:
\begin{equation}
    \text{COMP}(n,K) = \text{COMP}(n,K-1) + \frac{n}{K-2}\text{COMP}(n,K-2) \text{.} \label{eq:comp_rec}
\end{equation}

\subsection{Code length of the model}
\label{subsec: cl_model}
We first discuss in detail why properly encoding all models in the model class is difficult, and then describe the code length of model in the partitioning step and the merging step respectively. 
 
\subsubsection{Encoding all models in the model class is difficult}
According to Kraft's inequality, encoding all models in the model class is equivalent to assigning a prior probability distribution to all models \citep{grunwald2007minimum}. This prior distribution should reflect the model complexities \citep{grunwald2004tutorial}, especially when there exists some hierarchical structure in the model class. For models with similar model complexity, the prior distribution should be non-informative. Particularly, a common practice is to divide the model class into sub-classes according to the hierarchical structure, and then assign the prior distribution to each model by first assigning some prior to all the sub-classes and then assigning a uniform prior to all models within each sub-class. 

The model class of all histogram models (i.e., all partitions of $S$) has an apparent hierarchical structure with respect to model complexity. That is, the model class could be divided into sub-classes based on a combination of two factors: 1) the number of regions, and 2) the number of line segments composing the inner boundaries. Nevertheless, it is extremely challenging to assign a proper (or even an intuitively ``natural") prior distribution based on this complexity hierarchical structure, because of the following two reasons. 

First, it is difficult to specify a \emph{joint} prior distribution on the number of regions and the number of line segments, as they are dependent on each other, though specifying marginal prior distributions for each of the factors may be feasible. 

Second, given the number of regions, denoted by $K$, and the number of line segments composing the inner boundaries, denoted by $T$, it is challenging to count the number of models with $K$ regions and $T$ line segments. Hence, the prior probability of each model (with the uniform prior) within this sub-class is also difficult to obtain. On one hand, there is no analytical formula to obtain such count (to the best of our knowledge). On the other hand, to count this number algorithmically, we would first need to decide how many line segments each region has, i.e., to assign positive integers to $\{T_1, \ldots, T_K\}$ such that $T_1 + \ldots + T_K = T$. The number of possible values of $\{T_1, \ldots, T_K\}$ grows exponentially as $K$ increases. Further, we would need to decide where to put these line segments to form $K$ regions. The number of possible positions is enormous if $\epsilon$ is reasonably small. Finally, we would need to go over all individual cases to check for repeated counting for $T$, since regions can share line segments, which makes the counting computationally infeasible. 

\subsubsection{Code length of the model in the partitioning and merging steps}
As properly encoding all possible models within $\mathbb{M}$ turns out to be too difficult, we now discuss how to calculate the code length of the model separately for the partitioning and merging step.

\paragraph{Partitioning.} For a region $S' \in \widetilde{S}$, assume that there are $E$ candidate positions for cut lines, either horizontally or vertically. To encode the set of cut lines, we first encode the number of regions $K \in \{1,\ldots,K_{max}\}$, where $K_{max}$ is predetermined. We assign a uniform prior to $K$, and thus the code length needed to encode $K$ becomes a constant, which has no effect on the result of the partitioning step. Given $K$, we then encode the positions of $(K-1)$ cut lines, with again a uniform prior to all possible sets of $(K-1)$ cut lines. The code length needed in bits is
\begin{equation}
	L(C_{S'}) = \log {E \choose K-1} \label{eq:cl_split_model}
\end{equation}

\paragraph{Merging.} Next we discuss the code length of encoding all models in the constrained model class $\mathbb{M}_c$, which contains all possible models that can be obtained by merging neighboring regions of the partition after the partitioning step. 

We argue that we should have a non-informative prior on $\mathbb{M}_c$. First, as discussed before, it is challenging to specify a joint prior to both the number of line segments and the number of regions. Second, if neighboring regions are merged, the partition of the sample space tends to have fewer regions but more geometric complexity. Hence, there exists no obvious ways to compare model complexities, even in an intuitive manner. 

Thus, we treat the model complexities to be roughly equivalent and we assign a uniform prior to all models in $\mathbb{M}_c$. As a result, the code length of all models within $\mathbb{M}_c$ is a constant and has no effect on the result of the merging step. In other words, we only consider the code length of data in the merging step. 


\section{Revisiting MDL histograms for one-dimensional data}
\label{sec:prelim}
In this section, we elaborate the link of our work to the MDL-based histograms to one-dimensional data. 

We first show that searching for the best cut lines on one certain dimension of given two-dimensional data is equivalent to searching for the best cut points for the corresponding one-dimensional data. We then review the algorithm for inferring MDL histograms for one-dimensional data as proposed by \citet{kontkanen2007mdl}, and describe how we improve it both theoretically and practically. 

\paragraph{Notation and relation to our problem.} To be able to distinguish it from two-dimensional data $x^n$, we denote one-dimensional data as $z^n = (z_1, ..., z_n)$, with precision equal to $\epsilon$. Further, we define the sample space of $z^n$ as $[\min z^n, \max z^n]$.

We define the one-dimensional histogram model with $K$ bins as a set of \emph{cut points}, denoted as $C^K = \{C_0 = \min z^n, C_1, ..., C_{K} = \max z^n\} \subseteq C_a$, with $K \in \{0,1,\ldots,K_{max}\}$, where $K_{max}$ is pre-determined and $C_a$ is defined as 
\begin{equation}
    C_a = \{\min z^n, \min z^n + \epsilon, ..., \min z^n + E \cdot \epsilon, \max z^n\}, \label{eq:oned-candidate-points}
\end{equation}
with $E = \lfloor \frac{\max z^n - \min z^n}{\epsilon} \rfloor$. Note that we assume all subintervals to be closed on the left and open on the right, except that the rightmost subinterval is closed on both sides.

The code length needed to encode the model $C^K$ is 
\begin{equation}
L(C^K) = \log {E \choose K-1}, \label{eq:oned-cl-model}
\end{equation}
which is the same as Equation (\ref{eq:cl_split_model}). Further, based on the calculation of maximum likelihood given any histogram model (Section~\ref{subsec:code_length_for_data}) and Proposition~\ref{prop1}, the code length needed to encode $z^n$ given $C^K$ by the NML code is 
\begin{equation}
\begin{split}
	L(z^n|C^K) & = -\log P(z^n | C^K) + \log{ \text{COMP}(n,K) } \\
	 & = -\log \prod_{j = 1}^K (\frac{h_j \, \epsilon}{n \, (C_{j+1} - C_j)})^{h_j} +  \log{ \text{COMP}(n,K) }.
\end{split}
\end{equation}
If we compare $L(z^n|C^K)$ and $L(C^K)$ with Equations (\ref{eq:mle_xn}) and (\ref{eq:cl_split_model}), we can see that the definition of the two-dimensional MDL-optimal cut lines and the one-dimensional MDL-optimal cut points only differ by a constant. Thus, given a two-dimensional dataset $x^n = \{(x_{11},x_{21}),\ldots,(x_{1n},x_{2n})\}$, the optimization task of searching for the MDL-optimal vertical (or horizontal) cut lines is equivalent to the task of searching for the MDL-optimal one-dimensional cut points based on one-dimensional dataset $z^n = \{x_{11},\ldots,x_{1n}\}$ (or $z^n = \{x_{21},\ldots,x_{2n}\}$). That is, $z^n$ is the projection of $x^n$ on the x- or y-axis. 

In other words, the algorithm for constructing MDL-based one-dimensional histograms proposed by \citet{kontkanen2007mdl} can be directly applied to the partitioning step of our model selection task. We now briefly review this algorithm and show how we improve it both theoretically and practically. 

\paragraph{Improved one-dimensional MDL-based histograms. } We improve the one-dimensional algorithm proposed by \citet{kontkanen2007mdl} in two ways. First, in their previous work, the candidate cut points, denoted as $C'_a$, are chosen based on the data $z^n$, i.e., $C'_a = \bigcup_{i=1}^n\{z_i \pm \epsilon\}$, and hence the code length of model is calculated \emph{dependent} on given dataset, i.e., $L(C^K|z^n)$ is calculated instead of $L(C^K)$, which is theoretically sub-optimal, because generally
\begin{equation}
	L(z^n,C^K) = L(z^n|C^K) + L(C^K) \neq L(z^n|C^K) + L(C^K|z^n) \text{.}  
\end{equation}
In practice, this will cause significantly worse results when the sample size is very small. In such cases, the size of the set $C'_a$ will be very small, and hence the code length of model
will be significantly underestimated, leading to serious overfitting. We fix this problem by encoding the model independent of the data, as defined by Equations (\ref{eq:oned-candidate-points}) and (\ref{eq:oned-cl-model}).

Further, we show that we do not need to consider \emph{all} candidate cut points within $C_a$, but just those cut points with a data point near it from left or right, without other cut points in between. That is, we have the following.
\begin{prop} \label{prop2}
For any two cut points $C_i, C_k \in C_a$, suppose $C_i < C_k$ and no data points exist in the interval $[C_i, C_k]$, then any cut point $C_j \in [C_i, C_k]$ would not be in the MDL-optimal set of cut points, i.e., we can skip all such $C_j$ during the search process.
\end{prop}
This reduces the search space to a subset of $C_a$, and hence reduces the computational requirements. We include the proof in Appendix~B. 

Finally, we simplify the recursion formula for the dynamic programming proposed by \citet{kontkanen2007mdl} in their original paper, which significantly reduces empirical computation time. 

\paragraph{Dynamic programming algorithm.} 
\citet{kontkanen2007mdl} derived the recursion formula based on the total code length $L(z^n, C^K)$, i.e., 
\begin{equation}
\begin{split}
L(z^n, C^K) &= L(z^n|C^K) + L(C^K) \\
&=-\log(P(z^n|C^K) + \log \text{COMP}(n, K) + \log {E \choose K-1}.
\end{split}
\end{equation}

We show that we can simplify the recursion by only including the probability of the data, i.e., $P(z^n|C^K)$, instead of $L(z^n, C^K)$. Observe that when the number of bins $K$ is fixed, $L(C^K)$ and $\text{COMP}(n,K)$ become constant. Then, for fixed $K$, minimizing $L(z^n, C^K)$ is equivalent to minimizing $\{-\log(P(z^n|C^K)\}$, i.e., maximizing the likelihood. 

Therefore, minimizing $L(z^n, C^K)$, for all $K \in  \{1,\ldots,K_{max}\}$, can be done in two steps: 1) find the maximum likelihood cut points with fixed $K$, denoted as $\hat{C}^K$, for each $K$, using the following dynamic algorithm; and 2) calculate $L(z^n|\hat{C}^K)$ for each $K$, and find the $\hat{K} \in  \{1,\ldots,K_{max}\}$ that minimizes $L(z^n, \hat{C}^K)$. Then, 
\begin{equation}
    \hat{C}^{\hat{K}} = \arg \min_{K \in  \{1,\ldots,K_{max}\}, C^K \in C_a} \,\,\,\,\,\, L(z^n, C^K). \label{eq18}
\end{equation}
Now we describe the dynamic programming algorithm for finding $\hat{C}^K$ for each $K \in \{1,\ldots,K_{max}\}$. The (log) probability of $z^n$ given any cut points is
\begin{equation}
\begin{split}
    \log P(z^n|C^K) &= \sum_{i = 1}^n \log P(z_i | C^K) \\
    & = \sum_{j=1}^K \sum_{z_i \in [C_{j-1},C_j)} \log P(z_i | C^K) \\
    & = \sum_{j=1}^{K-1} \sum_{z_i \in [C_{j-1},C_j)} \log P(z_i | \{C^{K} \setminus C_K \}) +\sum_{z_i \in [C_{K-1},C_K]} \log P(z_i|C_K) \\
    & = \log P(z^n_{C_{K-1}} | \{C^{K} \setminus C_K \}) + \sum_{z_i \in [C_{K-1},C_K]} \log P(z_i|C_K)
\end{split}
\end{equation}
where $z^n_{C_K}$ is a constrained dataset containing all data points smaller than $C_K$, i.e.,
\begin{equation}
    z^n_{C_{K-1}} = \{z \in z^n | z < C_{K-1}\}. \label{eq20}
\end{equation}
Given the previous, the recursion formula is given by
\begin{equation}
\begin{split}
    \max_{C^K \subseteq C_a} \log P(z^n| C^K)= 
     \max_{C_K \in C_a} [&\max_{ \{C^K \setminus C_K\} \subseteq C_a} \log P(z^n_{C_{K-1}}|\{C^K \setminus C_K\}) \\
     & + \sum_{z_i \in [C_{K-1},C_K]} \log P(z_i|C_K) ]
\end{split}
\end{equation}
and hence a dynamic programming algorithm can be applied to search all $K \in \{1,\ldots,K_{max}\}$. In practice, $K_{max}$ is pre-determined, and larger $K_{max}$ should be investigated if $\hat{K} = K_{max}$.

The disadvantage of implementing the dynamic programming algorithm based on $L(z^n, C^K)$, $\forall K \in \{1,\ldots,K_{max}\}$, is that we would need to calculate the parametric complexity $\text{COMP}(\cdot)$ for every constrained dataset. Our improved version, in contrast, involves only $P(z^n|C^K)$, and thus we only need to calculate $\text{COMP}(\cdot)$ for the full dataset $z^n$ when calculating $L(z^n, \hat{C}^K)$ for each $K$, which will be much faster in practice.

The essential component of the dynamic programming algorithm is to construct the constrained dataset $z^n_{C_{K-1}}$, $\forall K \in \{1,\ldots,K_{max}\}$. These constrained datasets are easy to construct in the one-dimensional case with a natural order, but infeasible for two or higher dimensional cases. Hence we resort to the heuristic algorithm presented in the next section.
 

\section{The PALM Algorithm for Partitioning and Merging}
\label{sec:alg}

We propose a heuristic algorithm named PALM, which infers histogram models for two-dimensional data by decomposing the overall model selection problem into two steps: 1) \underline{p}artition space $S$ \underline{al}ternately based on the discretization result from previous iterations until it stops automatically; and then 2) \underline{m}erge neighboring regions if their densities are very similar. Both steps use the MDL principle as the decision criterion, with the code length defined in Section~\ref{sec: code}.

The PALM algorithm is given in Algorithm~\ref{al:partition}. Specifically, we first initiate $\widetilde{S}=\{S\}$ and choose the starting direction (line 1); then we iterate over all regions in $\widetilde{S}$ and partition each of them by searching for the MDL-optimal cut lines in the chosen direction (lines 3--5), and update $\widetilde{S}$ accordingly (lines 8--10); then, we keep iterating until $\widetilde{S}$ is no longer updated (lines 2 and 6--7), which completes the partitioning step. 

Next, the merging step searches, in a greedy manner, for the MDL-optimal partition of $S$ over all possible partitions that can be obtained by merging any two neighboring regions of the partition that is obtained in the partitioning step. That is, we list all the neighboring pairs of regions in $\widetilde{S}$, i.e., two regions that share part of their boundaries (line 15); then, we merge the pair that compresses the data most (or equivalently, decreases the MDL score most) and update the neighboring pairs list (lines 21--23); finally, we stop the merging step when no better compression can be obtained by merging any neighboring two pairs in $\widetilde{S}$ (lines 19--20). 

\begin{algorithm}[h]
\caption{PALM} 
\hspace*{0.02in} {\bf Input:} 
data $x^n$, data precision $\epsilon$, sample space $S$, maximum number of splits per partitioning step $K_{max}$\\
\hspace*{0.02in} {\bf Output:} 
$\widetilde{S}$, a partition of $S$
\begin{algorithmic}[1]

\State $dir \assign 0$ or $1$   \Comment{Initial partitioning direction: 0 and 1 represent horizontal and vertical}
\While{\textbf{true}} \Comment{Partitioning step.}
\For{$S_k \in \widetilde{S}$}
	\State Partition $S_k$ as $\widetilde{S_k}$ by finding the optimal cut lines for $S_k$ in direction $dir$ 
	\State $C^*_{S_k} = \arg \min_{{C_{S_k}} } L(\{x^n \in S_k\}, {C_{S_k}})$ 
\EndFor
\If{$\widetilde{S_k} = \{S_k\}$, for all $S_k \in \widetilde{S}$}
	\State \textbf{break}
\Else
    \State $\widetilde{S} \assign \bigcup\widetilde{S_k}$
    \State $dir \assign 1 - dir$
\EndIf
\EndWhile

\State

\State $\widetilde{S}_{merge} \assign \widetilde{S}$ \Comment{Merging step.}
\State $K_{merge} \assign$ the number of regions of $\widetilde{S}_{merge}$
\While{\textbf{true}} 
\State Get all neighboring pairs of regions of $\widetilde{S}_{merge}$, $Pairs \assign \{(S_j, S_k),\ldots\}$
\For{$(S_j, S_k) \in Pairs$}
    \State $\widetilde{S'_{j,k}} \assign$  merge the pair $(S_j, S_k)$  in $\widetilde{S}_{merge}$ 
    \State Calculate $L(x^n, \widetilde{S'_{j,k}}) = -\log \left(P(x^n|\widetilde{S'_{j,k}})\right) + \log \text{COMP}(n, K_{merge}-1)$
\EndFor
\If{$\min_{S'_{j,k}} L(x^n, \widetilde{S'_{j,k}}) > L(x^n, \widetilde{S}_{merge})$}
    \State \Return $\widetilde{S}_{merge}$
\Else
    \State $\widetilde{S}_{merge} \assign \arg \min_{\widetilde{S'_{i,j}}} L(x^n, \widetilde{S'_{i,j}})$
    \State $K_{merge} \assign K_{merge} - 1$
\EndIf
\EndWhile
\end{algorithmic}
\label{al:partition}
\end{algorithm}

\paragraph{Algorithm complexity.} We now discuss the worst-case algorithm complexity for the partitioning and merging step respectively, and we will show the empirical runtime in Section~\ref{subsec:time}.

For the first iteration of the partitioning step (i.e., when $\widetilde{S} = \{S\}$), the algorithm has a complexity of $\mathcal{O}(K_{max}E^2)$, the same as the one-dimensional case \citep{kontkanen2007mdl}, where $E$ is the number of possible locations for vertical (or horizontal) lines within the whole sample space $S$, based on the fixed grid with granularity $\epsilon$. The second iteration has a worst-case time complexity of $\mathcal{O}(K_{max}^2E^2)$ when the first iteration produces exactly $K_{max}$ regions. Following this line, the worst-case time complexity of the partitioning step is $\mathcal{O}(K_{max}^IE^2)$, where $I$ is the number of iterations.

As for the merging step, the time complexity is bounded by $K_pK_0$, where $K_0$ denotes the number of regions after the partitioning step, and $K_p$ denotes the number of neighboring pairs. That is, we can merge at most $(K_p-1)$ times, and each merging requires going over all the neighboring pairs.

Although the worst-case time cost for the partitioning step is exponential, and $K_0$ and $K_p$ could be large in practice, we will show in Section~\ref{subsec:time} that the empirical runtime may scale much better than exponential growth. 

\paragraph{Choosing the hyper-parameter settings.} 
We now briefly discuss how to choose $\epsilon$ and $K_{max}$ in practice. First, we should set $\epsilon$ to be the same as the precision of the data by default; data is always recorded up to a precision in practice. Further, when prior knowledge exists given a specific task, $\epsilon$ may be larger than the recording precision, because the domain expert or data analyst may decide that the data is only meaningful up to a more coarse precision.

Second, theoretically we should set $K_{max}$ to be sufficiently large, and hence in practice $K_{max}$ is a ``budget" rather than a hyper-parameter like the threshold or stopping criterion in other discretization methods (e.g., Nguyen 2014, Kerber 1992). That is, unlike these hyper-parameters, which can be either too large or too small and hence need to be carefully tuned, $K_{max}$ can be simply picked to be as large as possible. 

This makes our method practically hyper-parameter-free, in the sense that---given the guidelines above---no tedious hyper-parameter tuning should be necessary to obtain the best possible results.

\section{Experiments}
\label{sec::exp}

In this section, we investigate the performance of PALM using synthetic data, after which we will apply it to real-world data in the next section. We show that PALM can construct two-dimensional histograms that are adaptive to both local densities and sample size of the data. 

We start off by defining the ``loss" that we use for quantifying the quality of the ``learned" partitions. We then present experiment results on a wide variety of synthetic data. Although our algorithm relies on heuristics, we show that it has a number of desirable properties, as follows.

First, if the data is generated by a histogram model within our model class $\mathbb{M}$, PALM is able to identify the ``true" histogram given a large enough sample size. The results are discussed in Section \ref{subsec:exp1}.

Second, in Section \ref{subsec:exp_sine} we show that PALM has the flexibility to approximate histogram models outside the model class $\mathbb{M}$. Specifically, we study the behavior of PALM on a dataset generated as follows: we set the sample space $S = [0,1]\times[0,1]$, and partition it by a sine curve; we then generate data points uniformly distributed above and below the sine curve, with different densities.  

Third, we study the performance of PALM on data generated by two-dimensional Gaussian distributions in Section \ref{subsec:gaussian}. We show that it inherits the property of the one-dimensional MDL histogram method \citep{kontkanen2007mdl} that the bin sizes of the histogram are self-adaptive: the two-dimensional bin sizes become smaller locally where the probability density changes more rapidly. 

Fourth, in Section \ref{subsec:exp3} we compare PALM with the IPD algorithm \citep{nguyen2014unsupervised}, using a simple synthetic dataset that is almost identical to what has been used to study the performance of IPD \citep{nguyen2014unsupervised}.

Note that we always set $\epsilon = 0.001$, and all simulations are repeated $500$ times unless specified otherwise\footnote{The code is available at: https://github.com/ylincen/PALM}. 
The initial partitioning direction is fixed as vertical, to make the visualizations of the inferred partitions comparable. 

\subsection{Measuring the difference between two-dimensional histograms}
\label{exp:measure}
As PALM produces a histogram model and can be regarded as a density estimation method, one of the most intuitive ``loss" functions is the \emph{Mean Integrated Squared Error (MISE)} \citep{scott2015multivariate}, defined as
\begin{equation}
    \text{MISE}(\hat{f}) = \mathop{\mathbb{E}} [\int_S (f(x) - \hat{f}(x))^2 dx], \label{eq:MISE}
\end{equation}
where $f$ is the true probability density and $\hat{f}$ is the histogram model density estimator. We report the empirical MISE by calculating the integral numerically, and estimating $\mathbb{E[\cdot]}$ by the empirical mean of results over all repetitions of the simulation.

As MISE cannot indicate whether there are more ``bins" than necessary, we also propose two ``loss" functions that directly quantify the distances between the inner boundaries of the learned and true partitions of a sample space $S$. We first break up the line segments of the inner boundaries into \emph{pixels} with a precision set to $0.01 = 10\epsilon$ (merely to speed up the calculation). Then we introduce two loss functions based on the idea of \emph{Hausdorff distance}, considering \emph{false positives} and \emph{false negatives} respectively. Namely, we propose $L_{\text{learn}}$, based on the learned partition, and $L_{\text{true}}$, based on the true partition:
\begin{equation}
L_{\text{learn}} = \sum_{p \in P}{min_{q \in Q} ||p-q||^2}; L_{\text{true}} = \sum_{q \in Q}{min_{p \in P} ||p-q||^2} \label{eq27}
\end{equation}
where $||\cdot||$ denotes the Euclidean distance and $P$ and $Q$ are the sets of \emph{pixels} on the line segments of the learned partition and the true partition, respectively. 

The intuition for $L_{\text{learn}}$ is that, for a given pixel on a line segment of the learned partition, we find on the line segments of the true partition the pixel closest to it, and measure their distance; for $L_{\text{true}}$ it is the other way around. Thus, if $L_{\text{learn}}$ is large, the learned partition must have unnecessary extra line segments, whereas if $L_{\text{true}}$ is large, the learned partition fails to identify part of the line segments that actually exist. 

\subsection{Revealing ground truth two-dimensional histograms}
\label{subsec:exp1}

We describe the settings for simulating the data and then our experiment results, to empirically show that our algorithm can identify the ``true" histogram model if the data is generated by it.
\paragraph{Experiment settings.}
To randomly generate the ``true" partitions, we use a generative process that is very similar to the search process of our algorithm: we fix a rectangular region, $S = [0,1] \times [0,1]$, randomly generate vertical cut lines to split it into $K_1$ regions, and randomly generate horizontal cut lines to split each of the $K_1$ regions into $(K_{21}, ..., K_{2,K_1})$ regions respectively. Then, for each pair of neighboring regions, we merge them with a pre-determined probability $P_{merge}$.


We set these hyper-parameters as follows:
\begin{equation}
K_1  = K_{21} = K_{22} = ... = K_{2,K_1} = 5; P_{merge} = 0.4; \epsilon = 0.001.
\end{equation}
With these hyper-parameters, our generative process is able to generate a diverse subset of $\mathbb{M}$, as $P_{merge}$ is chosen delicately to be not too small or too large. Figure~\ref{fig:exp1_example} shows four random examples of the true partitions and learned partitions. These learned partitions are produced with the sample size set as $10\,000$. 

After the partition is fixed, we generate ``true" density parameters for the histogram model using a uniform distribution, i.e.,
\begin{equation}
        f_j \sim \text{Uniform}(0,1), \forall i = 1, 2, ..., K;  \label{eq25}
\end{equation} 
and normalize them such that $\sum_{j = 1}^{K} f_j |S_j| = 1$, where $K$ is the number of regions in total and $|S_j|$ is the geometric area of $S_j$. Note that we do not force the $f_j$ to be different from each other.
\begin{figure*}
\tabcolsep=0.11cm
\begin{tabular}{cccc}
\input{tex_pic_PALMexampleOnHistogramSimulated.tex}
\end{tabular}
\caption{Random examples of true (black solid) and learned partitions (red dashed) of the experiment in Section \ref{subsec:exp1}, mainly to show that our experiment settings can produce very flexible partitions of $[0,1]\times[0,1]$. Note that the sample size is set as 10\,000, which is \emph{not} enough for MISE (Equation \ref{eq:MISE}) to converge to almost 0, but the learned partitions by PALM already look promising: it can partly identify the true partitions.}
\label{fig:exp1_example}
\end{figure*}

\paragraph{Results.}
Figure~\ref{fig:MISE} shows that MISE is already small for small sample size, and converges to almost 0 as the sample size increases. We also show, in Figure~\ref{fig:loss_1}, that  $L_{\text{learn}}$ and $L_{\text{true}}$ converge to almost zero except for some outliers. 

The outliers of $L_{\text{learn}}$ are due to sampling variance when generating data points, the number of which decreases significantly as the sample size grows.

The outliers of $L_{\text{true}}$, however, are due to the random generation of the density parameters $f_j$. As we do not force all $f_j$'s to be different, they could accidentally turn out to be very similar. In that case, some of the ``true" inner boundaries are actually unnecessary, and our algorithm will ``fail" to discover them. Table 1 confirms that this is the cause of outliers when the sample size is large ($\geq1e5$): when PALM fails to identify part of the ``true" inner boundaries and $L_{true} > 1$, the learned histogram still estimates the density very accurately. The only explanation is then that some regions of the true partition accidentally have very similar $f_j$'s.

Moreover, when the sample size is moderate, e.g., 5000, $L_{\text{learn}}$ is already small, meaning that PALM can partly identify the true partition quite precisely, and rarely produces unnecessary extra regions. As the sample size increases, $L_{\text{true}}$ decreases, indicating that the learned partition becomes more and more complex; i.e., it is shown that the model selection process is self-adaptive to sample size. 
\begin{figure}[h]
\resizebox {\textwidth} {!} {
\input{fig_MISE.tex}
}
    \vspace{-6mm}
    \caption{Sample size vs MISE: MISE converges to almost 0 when the sample size becomes larger than $100\,000$. The range between the $5$th and $95$th percentiles is shown in blue. }
    \label{fig:MISE}
\end{figure}
\begin{figure}[h]
\centering
\input{fig_loss_1.tex}
    \vspace{-8mm}
    \caption{Boxplots showing the sample size versus $L_{\text{learn}}$ and $L_{\text{true}}$ as defined in Equation (\ref{eq27}). Note that the y-axis has a logarithmic scale. $L_{\text{learn}}$ is generally much smaller than $L_{\text{true}}$, meaning that it is very rare that PALM produces unnecessary extra regions. When the sample size is large enough for MISE to converge ($n \geq 1e5$), outliers of $L_{\text{true}}$ are due to sampling variance when generating the true parameters $f_j$ defined in Equation (\ref{eq25}), see Table 1; the number of outliers for  $L_{\text{learn}}$ decreases rapidly as the sample size becomes larger, as they are due to sampling variance when generating the data. }
    \label{fig:loss_1}
\end{figure}
\begin{table}[ht]
\centering
\begin{tabular}{rrr}
  \hline
Sample size & MISE for subgroup: $L_{true} > 1$ & overall MISE \\
\hline
 100 000 & 0.00148 & 0.00148 \\ 
  300 000 & 0.00055 & 0.00074 \\ 
  500 000 & 0.00051 & 0.00065 \\ 
  700 000 & 0.00019 & 0.00069 \\ 
  1 000 000 & 0.00023 & 0.00058 \\ 
  3 000 000 &  0.00017 & 0.00055 \\ 
  5 000 000 & 0.00006 & 0.00051 \\
  \hline
\end{tabular}
\caption{The average MISE of cases when $L_{true} > 1$, and the overall mean of MISE. We show that, when PALM fails to identify part of the true partitions, the learned histogram model still estimates the probability density accurately. The only explanation for these cases is that some neighboring regions in the true partitions have very similar ``true" $f_j$ as defined in Equation (\ref{eq25}), as a result of which PALM does not deem it necessary to further partition these regions.}
\end{table}

\subsection{Approximating histogram models outside model class $\mathbb{M}$}
\label{subsec:exp_sine}

We now investigate the case where the true model is not within model class $\mathbb{M}$, while the data is still generated uniformly within each region.

We show that, although the model class $\mathbb{M}$ is based on a grid, it is indeed flexible and expressive: in practice, the learned partitions can approximate true partitions outside $\mathbb{M}$, and the approximation becomes more and more accurate as the sample size increases. 

\paragraph{Experiment settings.} As an illustrative example, we partition $S = [0,1] \times [0,1]$ by several sine curves, defined as
\begin{equation}
    g(x) = \frac{1}{4}\sin{2m\pi x}+\frac{1}{2}\label{eq30} 
\end{equation}
and where $m$ is a hyper-parameter.

We randomly generate data from a uniform distribution above and under the sine curve, and we set the probability density above $g(x)$ to be twice as large as below $g(x)$, i.e., we uniformly sample $\frac{2}{3}n$ data points above $g(x)$, and $\frac{1}{3}n$ data points below $g(x)$, where $n$ is the total sample size. 

\paragraph{Results.}
\begin{figure}[h]
    \centering
    \includegraphics[width = 0.49\textwidth]{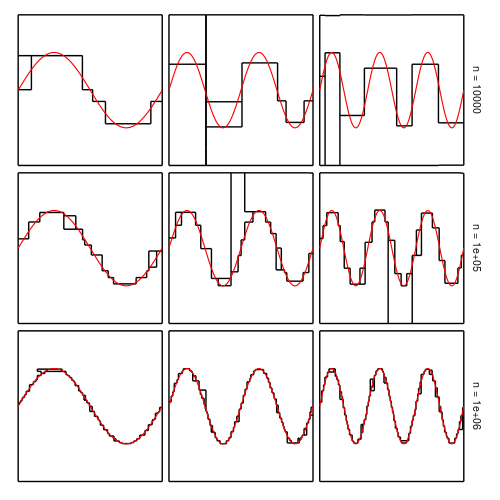}
    \includegraphics[width = 0.49\textwidth]{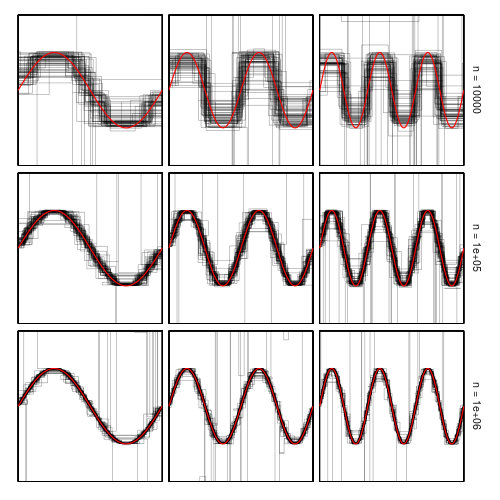}    
    \caption{(Left) Sine curve defined in Equation (\ref{eq30}) (red), with $m\in\{2,4,6\}$ from left to right on each row, and the learned partition by PALM (black). Data is randomly generated by uniforms distribution above and below the sine curve, within $S = [0,1] \times [0,1]$. Densities above and below the since curve are 2:1. From top to bottom, the sample sizes of the simulated data are $n \in \{1e4, 1e5, 1e6\}$. (Right) $50$ partition results of $50$ different simulated datasets are plotted \emph{together}. It shows that PALM is not guaranteed to be absolutely stable, as it occasionally produces undesired extra line segments, but the line segments of the learned partitions mostly gather around the true sine curve.}
    \label{fig:exp_sine}
\end{figure}
We empirically show that the learned partitions approximate the sine curves quite precisely, though occasionally a few extra undesired regions are produced. 
Figure~\ref{fig:exp_sine} (left) shows the learned partitions on single simulated datasets, with $m \in \{2,4,6\}$ to control the degree of oscillation, and sample size $n \in \{1e4,1e5,1e6\}$. We see that, as the sample size grows, our approximation becomes more and more accurate. 

However, since our algorithm is greedy in nature, there is no guarantee to find the partition with the global minimum score. In practice, PALM will occasionally produce undesired, extra line segments. Thus, to investigate the stability of the learned partitions, we repeat the simulation 50 times for each combination of $m$ and $n$, and plot \emph{all} partition results in one single plot in Figure~\ref{fig:exp_sine} (right).

Figure~\ref{fig:exp_sine} (right) shows that the undesired extra regions are produced more frequently as $m$ increases, but seems independent of sample size $n$. However, as sample size increases, the learned partitions become indeed more stable as they gather around the sine curves more closely. 

\subsection{Gaussian random variables}
\label{subsec:gaussian}
In this section, we show the performance of our algorithm on data generated from a two-dimensional Gaussian distribution. Specifically, we consider two of them, i.e., $N[\begin{psmallmatrix}0\\ 0\end{psmallmatrix},\begin{psmallmatrix}1 & 0\\ 0 & 1\end{psmallmatrix}]$ and $N[\begin{psmallmatrix}0\\ 0\end{psmallmatrix},\begin{psmallmatrix}1 & 0.5\\ 0.5 & 1\end{psmallmatrix}]$, of which the key difference is whether the two dimensions are independent. We assume $S = [-5,5]\times[-5,5]$, as the true Gaussian density outside such $S$ is negligible. 

\begin{figure}[h]
\centering
    \includegraphics[width = 0.31\textwidth, height = 0.4\textwidth]{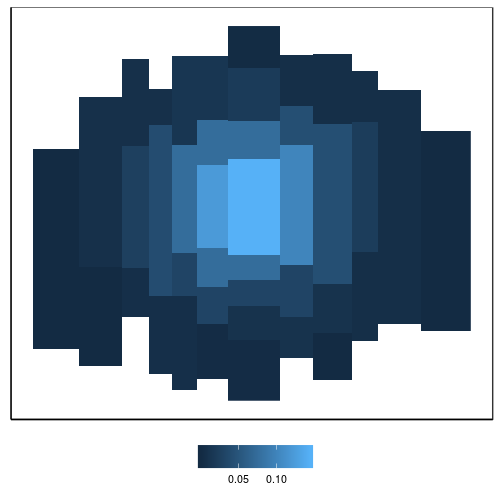}
    \includegraphics[width = 0.31\textwidth, height = 0.4\textwidth]{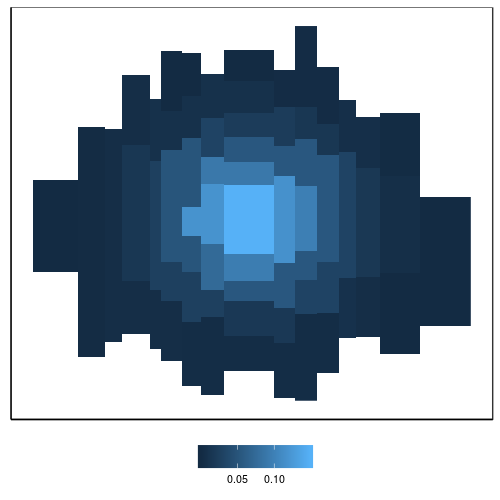}
    \includegraphics[width = 0.31\textwidth, height = 0.4\textwidth]{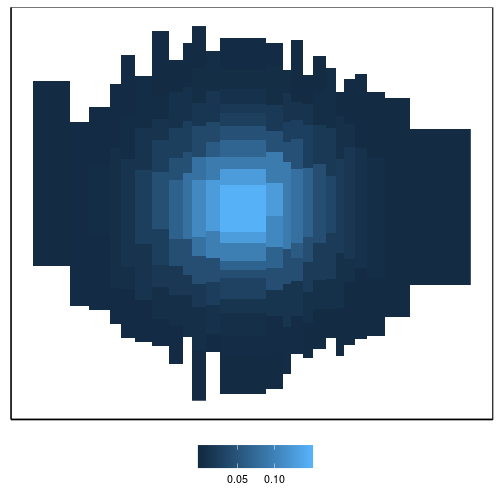} \\
    \includegraphics[width = 0.31\textwidth, height = 0.4\textwidth]{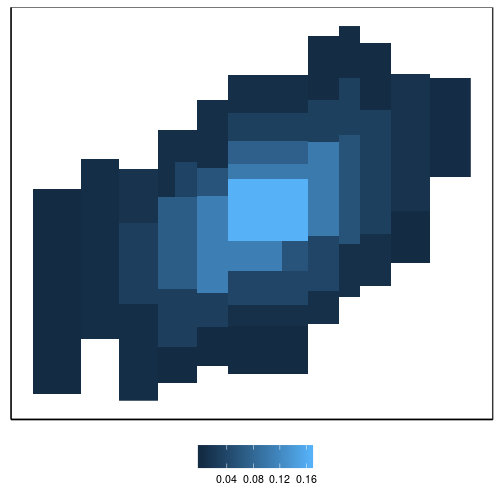}
    \includegraphics[width = 0.31\textwidth, height = 0.4\textwidth]{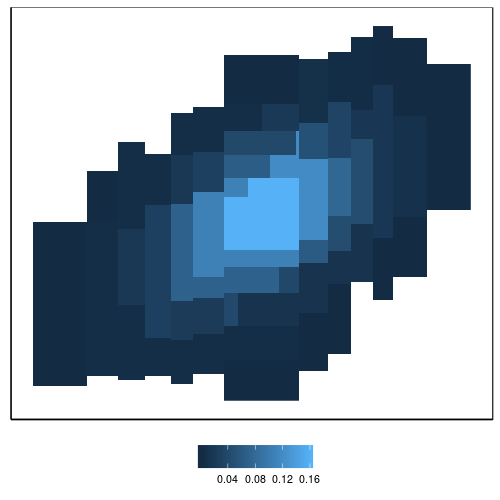}
    \includegraphics[width = 0.31\textwidth, height = 0.4\textwidth]{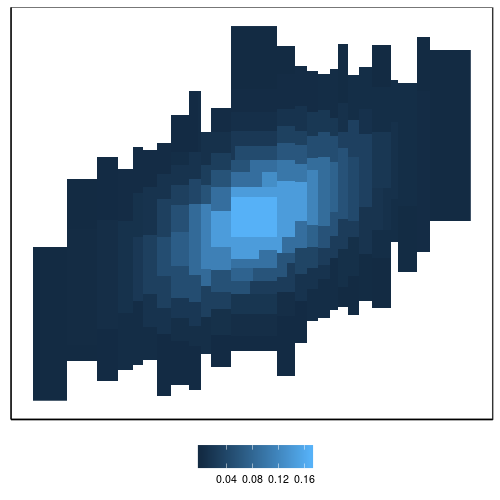}
    \caption{Learned partitions and estimated densities by PALM. The data is generated from two-dimensional Gaussian distributions, with sample size $n \in \{5\,000,10\,000,50\,000\}$, from left to right. The top and bottom row is respectively generated from independent and dependent two-dimensional Gaussian distributions.}
\label{fig:gauss}
\end{figure}

Figure~\ref{fig:gauss} shows the learned partitions as well as the learned empirical densities from a random simulated dataset with different sample sizes, $n \in \{5\,000, 10\,000, \\ 50\,000\}$. Note that bin size is self-adaptive with regard to sample size and local structure of the probability density. We also mention that the empirical runtime for a single dataset generated by such Gaussian distributions is at most a few minutes, for all $n \leq 50\,000$. 

To quantify the quality of the learned partitions by PALM, we compare the MISE of PALM to the MISE of fixed equally-spaced grid partitions with different granularities. Figure~\ref{fig:gauss_mise} shows the mean and standard deviation of MISE for different cases, and we conclude that, to achieve roughly the same level of MISE with a fixed grid, a fixed grid needs to have five times as many regions as a partition learned by PALM.

\begin{figure}[h]
    \centering
\input{fig_gauss.tex}
    \caption{For data generated from a two-dimensional Gaussian distribution, described in Section~\ref{subsec:gaussian}, the mean and standard deviation of MISE is calculated for different partitions: (from left to right) PALM, fixed grid with the same number of regions as PALM (denoted as `1x'), fixed grid with two times number of regions as PALM (denoted as `2x'), ..., and fixed grid with the same number of regions before the merging step of PALM (denoted as `*1x'). We assume $S = [-5,5]\times[-5,5]$, as the true Gaussian density outside $S$ is negligible.}
    \label{fig:gauss_mise}
\end{figure}

\subsection{Comparison with IPD}
\label{subsec:exp3}
Since---to the best of our knowledge---no existing discretization method can produce partitions as expressive as PALM, it seems not so meaningful to compare with any existing algorithm. However, we do include a comparison with the IPD algorithm \citep{nguyen2014unsupervised}, mainly to show that our algorithm not only can produce more flexible partitions by definition, but also beats this state-of-the-art algorithm on a ``simple" task, i.e., when the ``true" partition is an adaptive two-dimensional grid. 

We use simple synthetic data, similar to one of the synthetic datasets used to study the performance of IPD \citep{nguyen2014unsupervised}. The data is generated to be uniform within four regions in $S = [0,1] \times [0,1]$. These regions are produced by partitioning $S$ by one vertical line $x = V_x$ and one horizontal line $y = H_y$, where $V_x, H_y \sim \text{Uniform}(0,1)$. The number of data points within each region is equal.

We compare the loss, as defined in Equation (\ref{eq27}), and we show in Figure~\ref{fig:exp4} that 1) PALM has better performance on small datasets, and 2) as the sample size gets larger, PALM converges but IPD partitions $S$ into more and more regions, as can be witnessed from an increasing $L_{\text{true}}$. 

\begin{figure}[h]
    \centering
\input{fig_exp4.tex}
    \caption{Comparison of PALM and IPD, using the box-plot and the mean of $L_{\text{learn}}$ and $L_{\text{true}}$, as defined in Equation (\ref{eq27}). PALM not only performs better when the sample size is small, but also converges as the sample size increases, while IPD does not converge.}
    \label{fig:exp4}
\end{figure}

\subsection{Empirical runtime}
\label{subsec:time}
We next discuss the empirical runtime with respect to $K_{max}$, the maximum number of bins to search, and $E$, the number of candidate cut points. 

\begin{figure}[h]
	\centering
    \includegraphics[width = 0.48\textwidth]{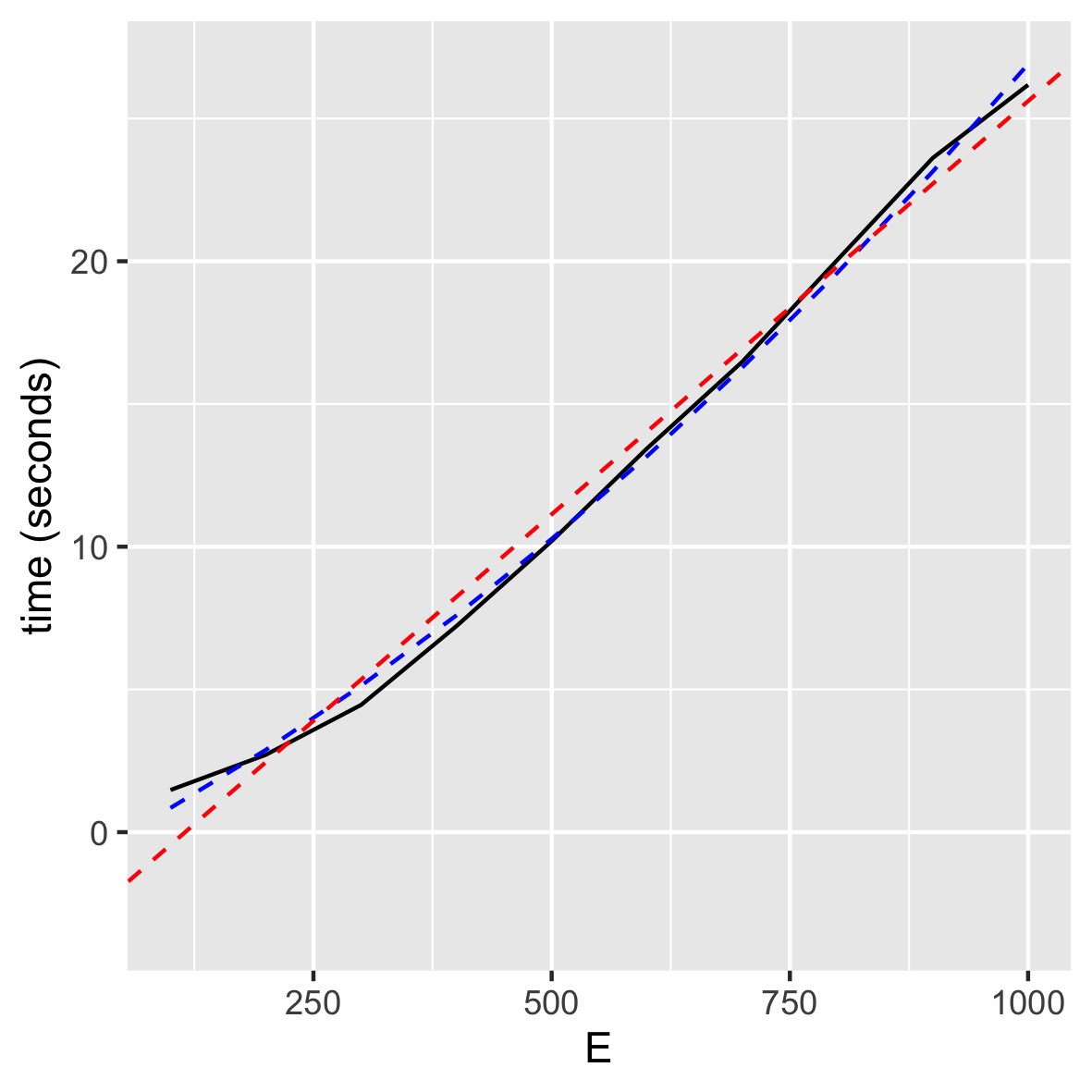}
    \includegraphics[width = 0.48\textwidth]{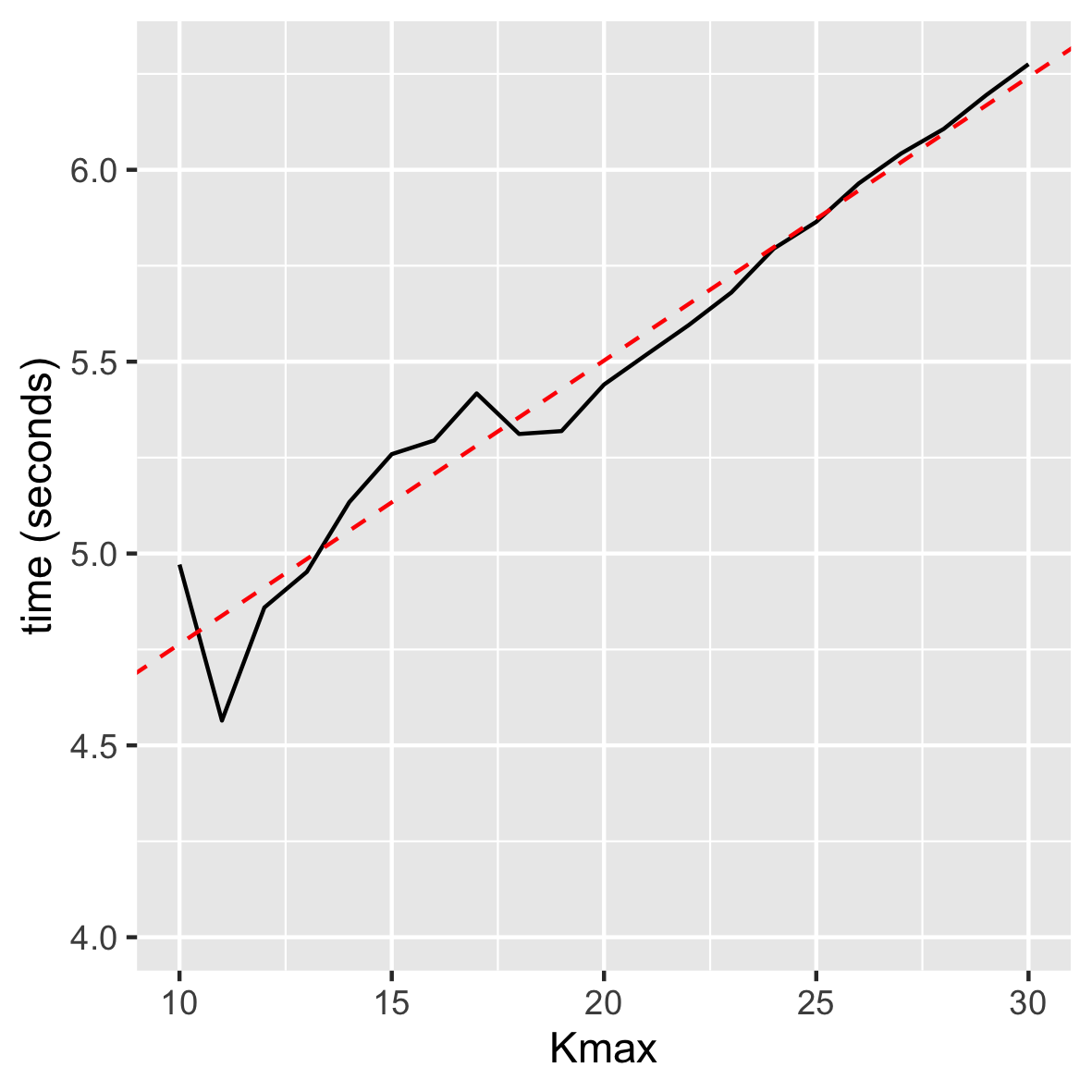}
    \caption{Empirical time complexity on simulated two-dimensional Gaussian data, with respect to $E$, the number of candidate cut points, and the runtime, and $K_{max}$, the maximum number of bins we search.}
    \label{fig:timecomplexity}
\end{figure}

Specifically, we use two-dimensional datasets simulated from independent standard Gaussian distributions to examine the relationship between $K_{max}$ and runtime, with fixed sample size equal to $500$ and $\epsilon = 0.001$. The results are illustrated in Figure~\ref{fig:timecomplexity}, showing that the runtime increases linearly with $K_{max}$. Further, to investigate the relationship between $E$ and the runtime, we again simulate from two-dimensional Gaussian distributions with different variance $\sigma^2$ to control the $E$ \footnote{For reproducibility, we first simulate $10\,000$ data points from $N[\begin{psmallmatrix}0\\ 0\end{psmallmatrix}, \begin{psmallmatrix}0 \,\, \sigma^2\\ \sigma^2\,\, 0\end{psmallmatrix}]$, where $\sigma^2 = E \epsilon / 2$, where $E$ is the desired number of candidate cut points. Since the corresponding desired data range with $E$ candidate cut points is $[-E \epsilon / 2, E \epsilon / 2]$, we next remove the data points outside this desired data range, and we finally randomly select $1\,000$ data points from the remaining data points. }. We fixed the sample size to be $1\,000$ and $\epsilon = 0.001$. The results show that, the runtime grows quadratically with $E$ (as shown by the blue dashed curve), but the second-order coefficient is quite small (as it is very close to the red dashed line with a linear trend). We report the runtime based on $500$ repetitions.

\section{Case study}
\label{sec:case}
We now show the results of applying our algorithm to real-world spatial datasets. We start with describing the three datasets we use in Section~\ref{subsec:case_data}. Next, we describe our case study tasks in Section~\ref{subsec:case_task}. Specifically, we inspect the results by visualizing the histograms, to illustrate that our method can be used as an explanatory data analysis (EDA) tool. We also compare with kernel density estimation (KDE), arguably the most widely used EDA method for spatial datasets, both for the visualizations and the goodness-of-fit on unseen data. In Section~\ref{subsec:case_res}, we report our results and show that 1) PALM can produce partitions that characterize more detailed density changes than KDE, and 2) PALM fits better on unseen data (i.e., a test dataset), in the sense that the partition of PALM has larger log-likelihood on the test dataset than KDE. Finally, we report the runtimes and detailed algorithm settings, respectively in Section~\ref{subsec:case_time} and \ref{subsec:case_algset}. 

\begin{figure}[!h]
    \centering
    \includegraphics[width = 0.48\textwidth]{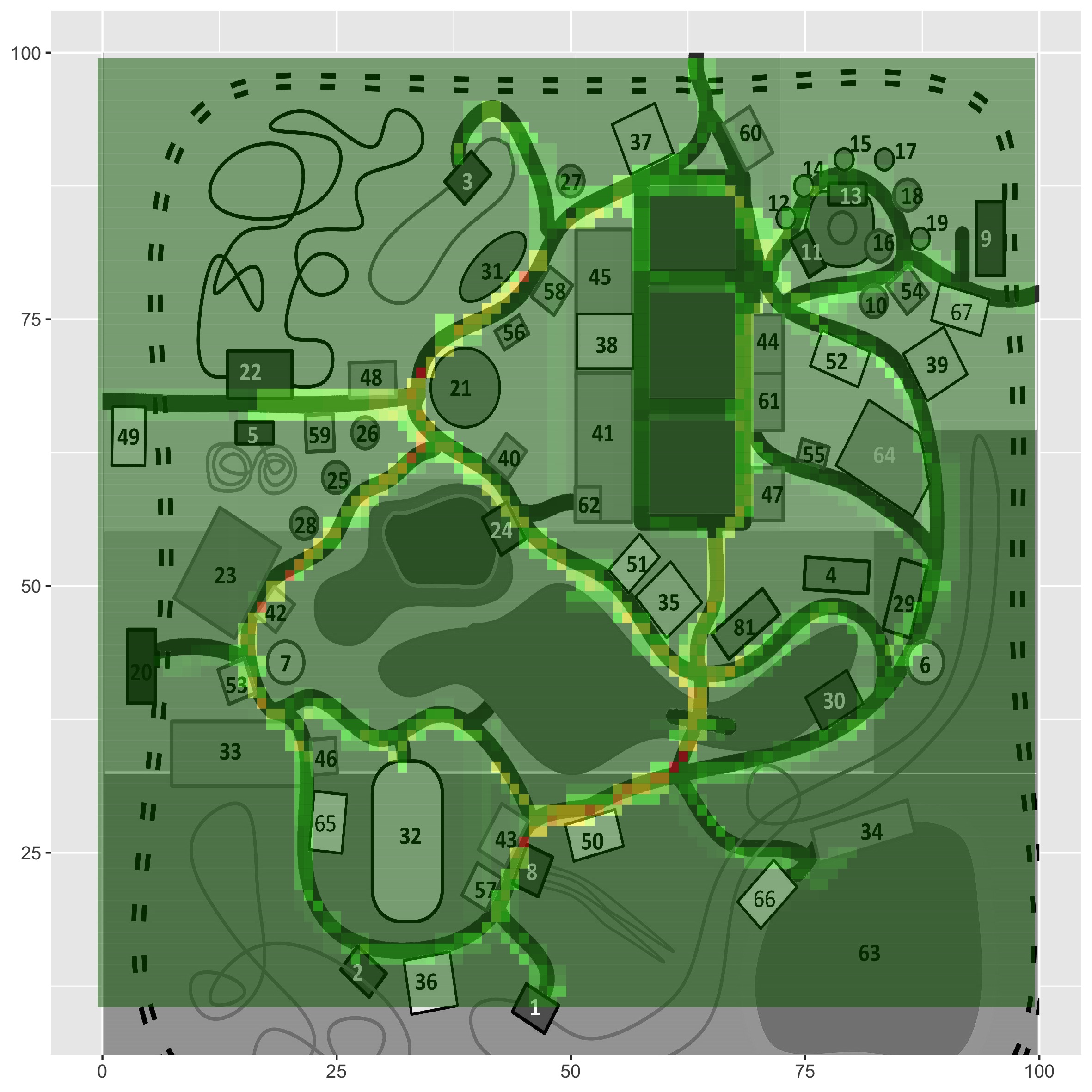}	
    \includegraphics[width = 0.48\textwidth]{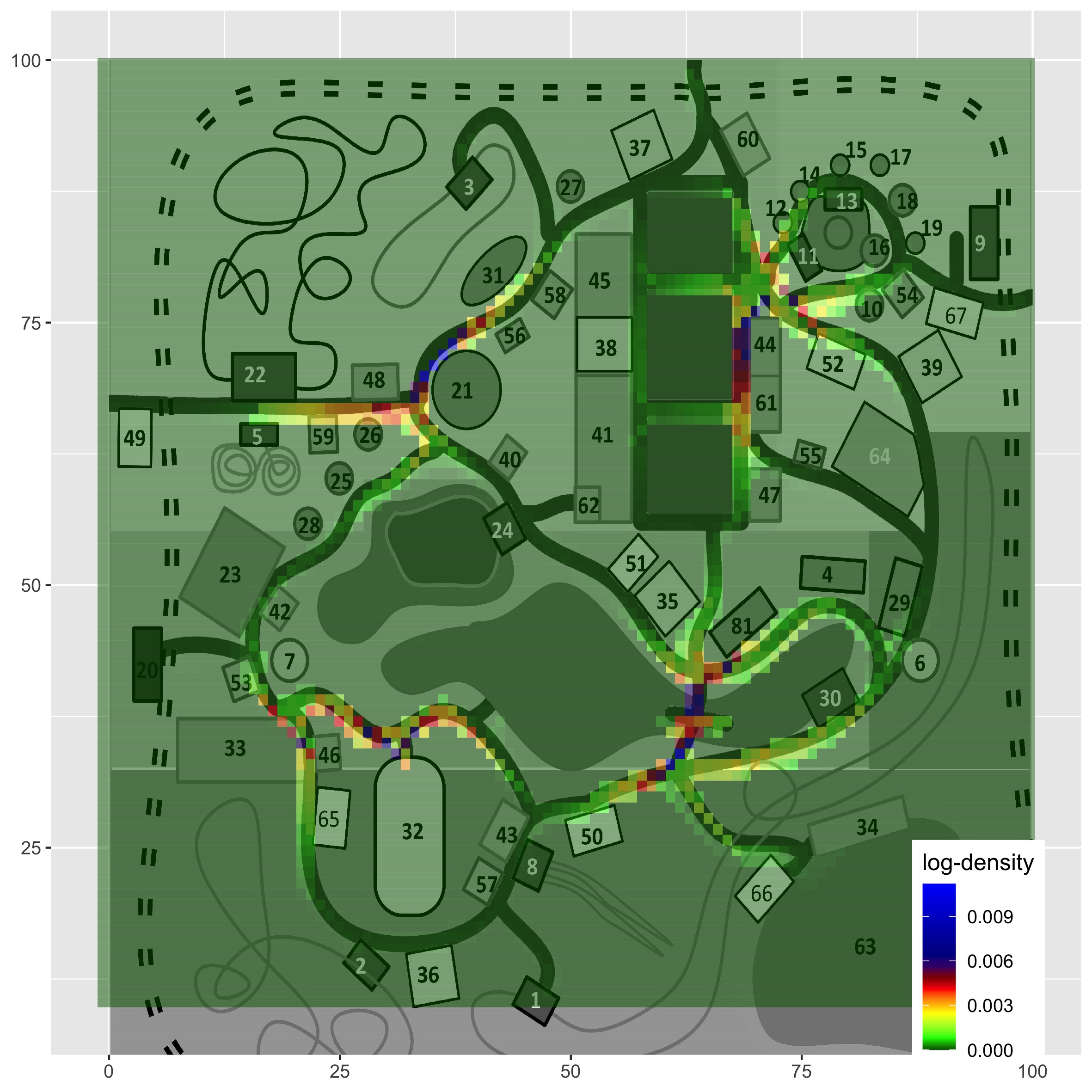}	
    \centering
    \includegraphics[width = 0.48\textwidth]{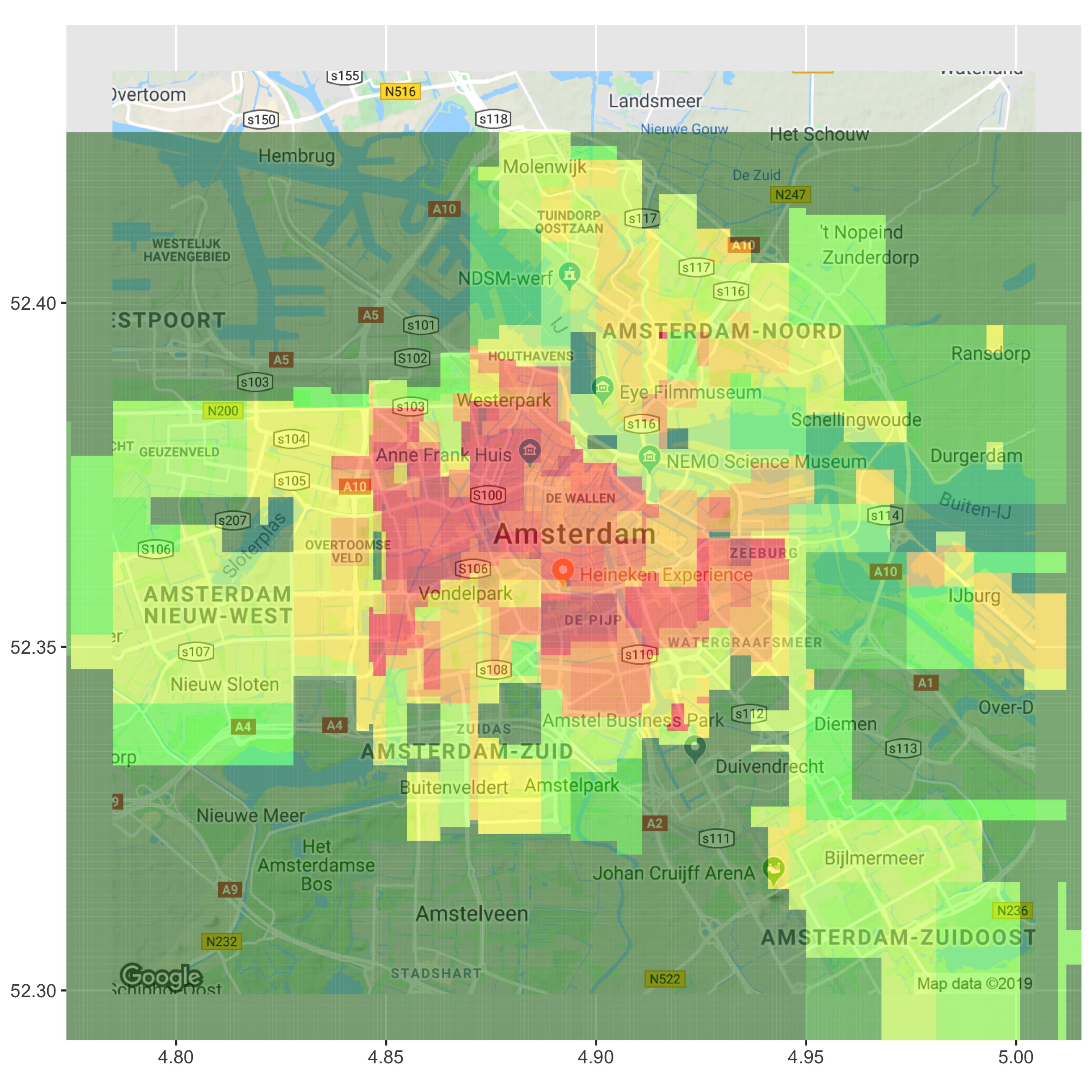}	
    \includegraphics[width = 0.48\textwidth]{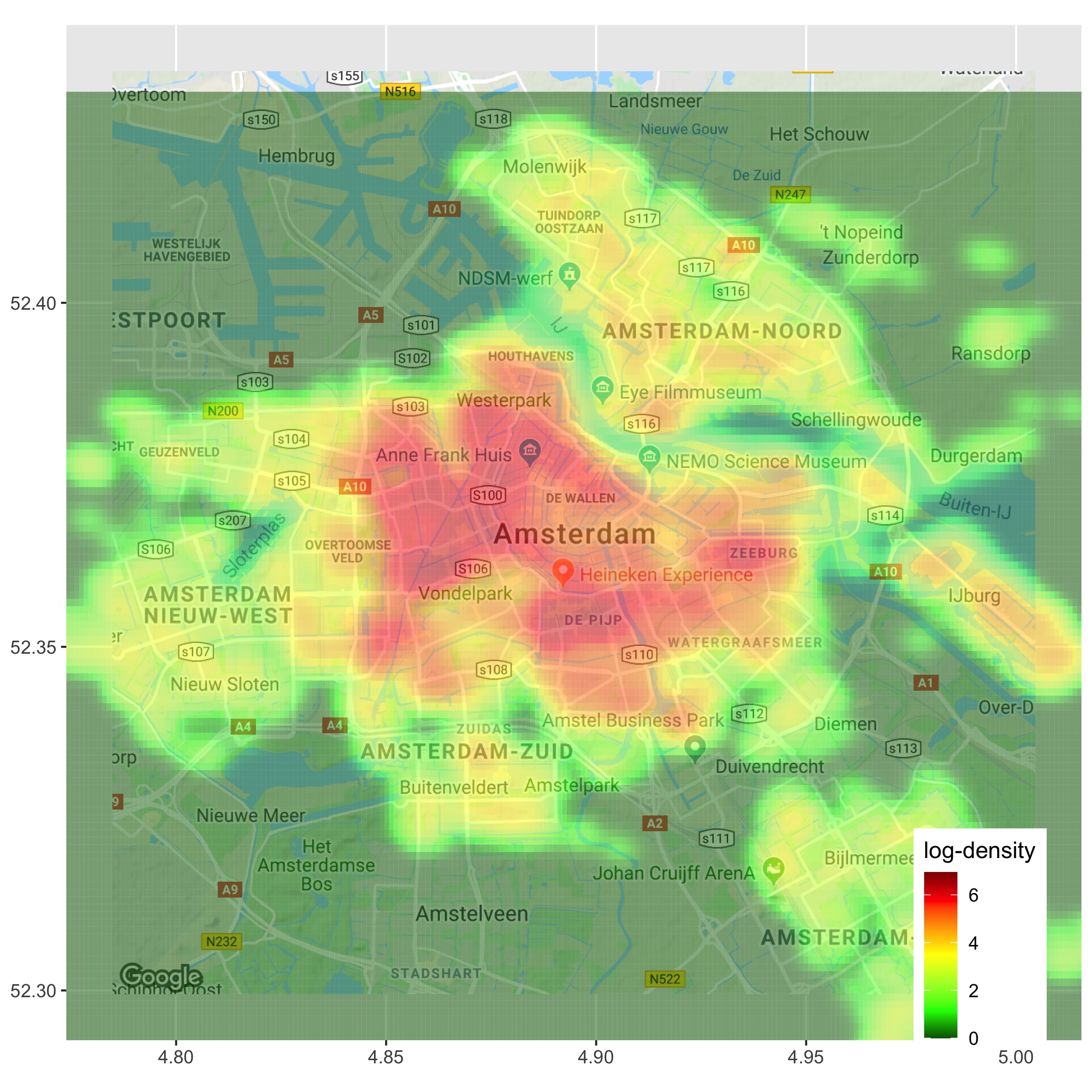}	
        \centering
    \includegraphics[width = 0.48\textwidth]{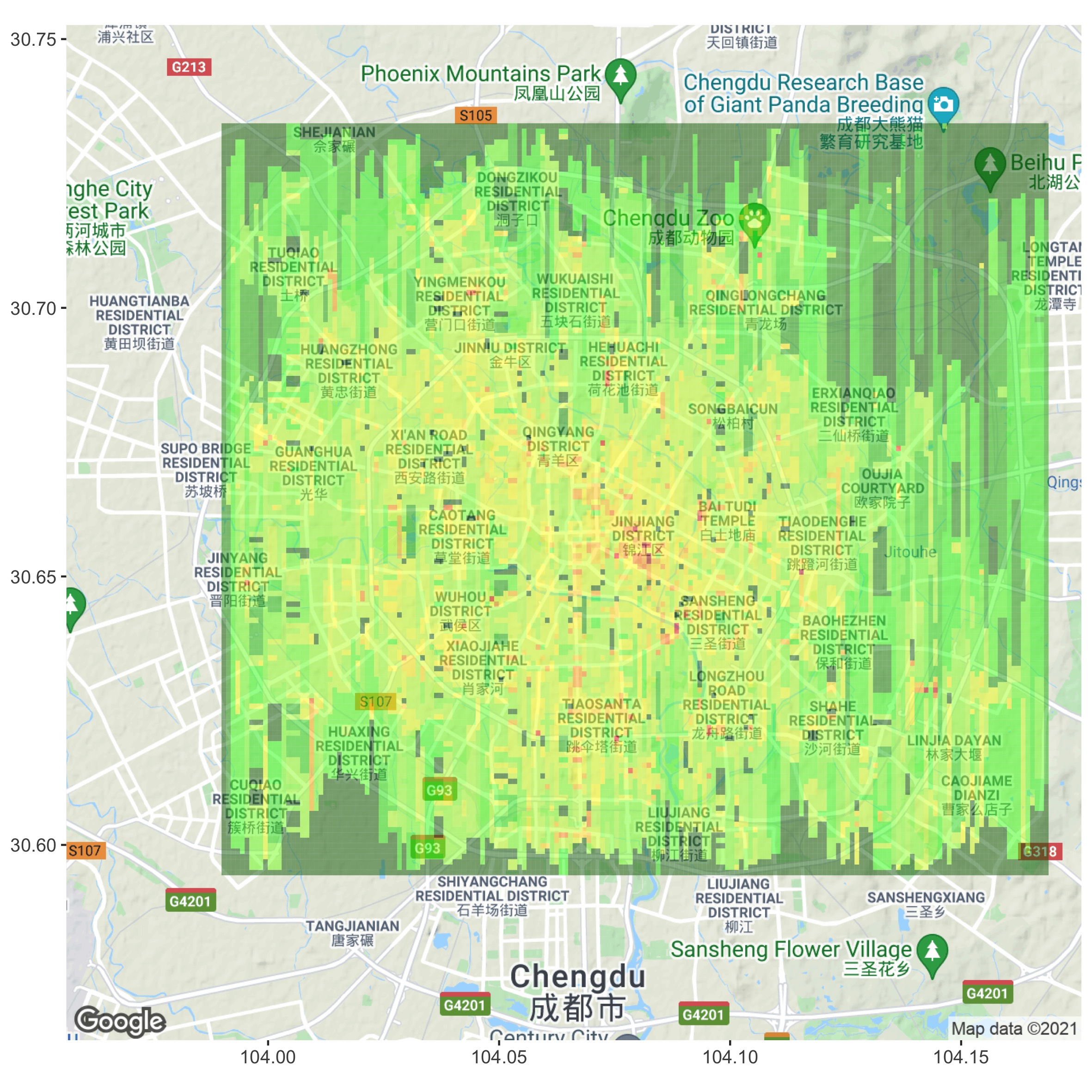}	
    \includegraphics[width = 0.48\textwidth]{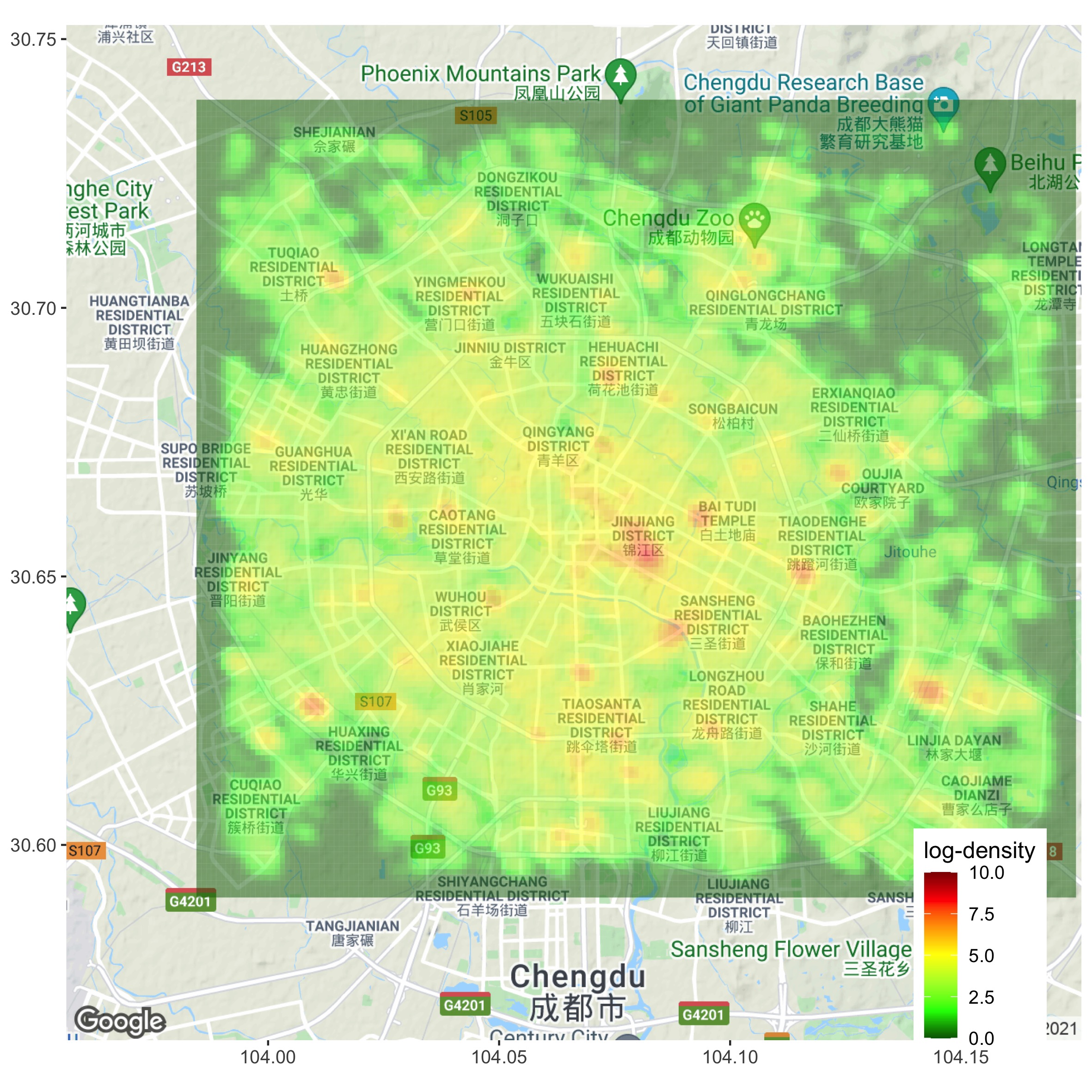}	
    \caption{Estimated densities on three real-world datasets using PALM (left) and KDE (right); from top to bottom: DinoFun amusement park, Amsterdam Airbnb housing, and taxi destinations in Chengdu.}
    \label{fig:case_res}
\end{figure}

\subsection{Datasets} 
\label{subsec:case_data}
We consider three diverse real world datasets: locations of Airbnb housing in Amsterdam\footnote{http://insideairbnb.com}, GPS locations of destinations of DiDi taxi queries in Chengdu, China\footnote{https://gaia.didichuxing.com}, and GPS recordings of visitors' movement in an amusement park\footnote{http://vacommunity.org/VAST+Challenge+2015}. 

\smallskip \noindent
\textbf{Visitors movement data in the DinoFun amusement park.} All visitors at the amusement park need to carry a device or use a smartphone app to check in at different attractions (e.g., roller coasters). Further, the amusement park is segmented into $100 \times 100$ cells (all of them are roughly 5 meters $\times$ 5 meters), and each cell has a sensor which can track the position of each visitor. The device (or the mobile app), together with the sensors, checks the position of the visitor every few seconds and records the position \emph{if the visitor moves to another cell}. Thus, applying PALM on this dataset will reveal the densities of places that people have been in the amusement park. This data has a sample size of $9\,078\,623$, in which every row represents a single position that one individual visitor visited (or passed by), with information like visitor's ID, timestamp, and location. 

\smallskip \noindent
\textbf{Amsterdam airbnb locations.} This data has a sample size of $20\,244$, and the location of each house is recorded by its longitude and latitude. Applying PALM on this dataset shows the distribution of Airbnb housing in Amsterdam. 

\smallskip \noindent
\textbf{DiDi taxi data in Chengdu.} The sample size of the data is $107\,573$, which collects the longitude and latitude of taxi destinations. Applying PALM on this dataset shows the densities of different regions that people visited by taxi in Chengdu, China.  

\subsection{Case study tasks}
\label{subsec:case_task}
\textbf{Explanatory data analysis and visualizations.} We first partition the three two-dimensional datasets by PALM and estimate the densities of all regions, using the full datasets. We visualize the densities by the heat maps in Figure~\ref{fig:case_res}, and compare the visualizations obtained by two-dimensional kernel density estimation (KDE) \citep{duong2007ks}, also with the full dataset. We also include the visualization results of the discretization obtained by IPD~\citep{nguyen2014unsupervised} for comparison, although IPD is not primarily designed for two-dimensional datasets. The background of Figure~\ref{fig:case_res} are the map of the DinoFun amusement part (provided together with the dataset), and the map of Amsterdam and Chengdu (from Google Maps API and the R package ``ggmap"~\citep{KahleGGmap}). 

\smallskip \noindent
\textbf{Comparison of KDE and PALM on the goodness-of-fit.} Further, to quantitively compare how KDE and PALM fit unseen data, and thus generalize to the underlying data distribution, we randomly split each dataset into training and testing set, obtain the PALM and KDE result from the training set, and compare the log-likelihoods on the testing dataset. We repeat the random splitting 100 times\footnote{To speed up the process, we randomly sampled a subset of the Chengdu taxi dataset that contains 10\% of the full sample size; also, for the amusement park movement dataset, we only use the subset of the data that is between 4 hours and 5 hours after the opening of the park, with sample size $713\,846$. Note that this is only for the comparison of goodness-of-fit but not for the visualizations and empirical runtime evaluation}.

\subsection{Case study results} 
\label{subsec:case_res}
We first analyze the result on each dataset respectively, based on which we give our concluding remarks for the case study at the end of this section. 

\smallskip \noindent
\textbf{Visitors movement data in the DinoFun amusement park.} As shown in Figure~\ref{fig:case_res}, both KDE and PALM reveal the walking path of the amusement park purely from the movement data (i.e., without knowing the map as additional information). Although KDE seems to capture more density changes, we show that it fits unseen data much worse than PALM, measured by the log-likelihood on the test dataset, shown in Table 2. Thus, we conclude that KDE may overfit on this dataset. 

\smallskip \noindent
\textbf{Amsterdam airbnb locations.} 
The visualizations of PALM and KDE look generally similar: if we treat red and orange regions in the center as the ``dense region", the rigid boundary between the dense region and the rest obtained by PALM approximates well the corresponding curve boundary obtained by KDE. However, note that more density changes are captured within the dense region, and PALM revealed two dense spots outside the central areas that KDE neglects, respectively on the top right and the bottom right of the map\footnote{The top right dense spot is close to the ``AMSTERDAM NOORD" text on the map (on the ``T"), and bottom right dense spot is near ``Amstel Business Park".}. Further, as shown in Table 2, the (average) log-likelihood of PALM and KDE on the test set is almost the same, which indicates that PALM does not overfit on this dataset, i.e., the dense spots revealed by PALM are valid. 

\smallskip \noindent
\textbf{DiDi taxi data in Chengdu.} The visualizations of PALM and KDE lead to different understandings of this dataset: while KDE reveals several hot clusters of taxi destinations, PALM shows that the density can change drastically within very small range of areas. As PALM fits better on the testing dataset, we conclude that PALM does not overfit but KDE may over-smooth this dataset.

\begin{table}[ht]
\centering
\begin{tabular}{rlrrrr}
 & Dataset & $l_{palm}$ & $l'_{palm}$& $l_{kde}$ & $(l_{palm}-l_{kde}) / l_{kde}$ \\ 
  \hline
1 & Amsterdam housing  & 29976.36 & 29983.31 & 30069.78 & -0.00 \\ 
  2 & Amusement park & 270.56 & 262.0688 & 227.22 & 0.19 \\ 
  3 & Chengdu taxi & 14904.28 & 14742.05 & 14073.42 & 0.06 \\ 
   \hline
\end{tabular}
\caption{The log-likelihood of PALM with partitioning vertically first, $l_{palm}$, and with partitioning horizontally first, $l'_{palm}$, and the log-likelihood of KDE, $l_{kde}$, on the test set for each of the three datasets. \label{table_case}} 
\end{table}

\smallskip \noindent
By default the PALM algorithm always starts by splitting the x-axis. Starting by splitting the y-axis leads to slightly different models, and thus somewhat different visualizations, but those differences are so minimal that they can be ignored for practical purposes. That is, the differences mostly appear in sparse regions, with very low densities, where no interesting patterns occur. To demonstrate that the differences are negligible, we compare the log-likelihoods obtained on unseen data when starting splitting on either the x-axis or the y-axis. The log-likelihoods are indeed very similar for both starting directions, as shown in Table 2, confirming that the resulting histogram models can only be different in sparse and less important regions; otherwise the log-likelihoods would be substantially different.

Based on the analysis above, we conclude that 1) although PALM partitions the dataset with rigid boundaries, PALM fits the data better than KDE when the datasets have drastic local density changes, such as the Chengdu taxi dataset and the amusement park  dataset; 2) when we have smoother two-dimensional data such as the Amsterdam housing dataset, PALM and KDE fit the data equally well; 3) when we look at the visualizations, PALM tends to capture more density changes than KDE, and PALM can reveal dense spots that KDE neglects; in other words, KDE tends to over-smooth the dataset. 

Last, we include the visualizations of the IPD discretization in Appendix~C, in which we demonstrate that the discretization results obtained by IPD have much coarser granularity. Hence, our discretization results preserve more information from the original datasets. 

\subsection{Empirical runtime}
\label{subsec:case_time}
We examine the empirical runtime on these three datasets in Table 3 (using the full datasets, without the split of training and testing set). We conclude that KDE is generally much faster, except on the amusement park dataset, which has a very large sample size but small $E$. 

\begin{table}[ht]
\label{table_case}
\centering
\begin{tabular}{rlrrr}
  \hline
 & Dataset & sample size & PALM & KDE \\ 
  \hline
1 & Amsterdam housing & $20\,244$ & 106.821 & 6.73 \\ 
  2 & Amusement park & $9\,078\,623$ & 1134.581 & 2017.215 \\ 
  3 & Chengdu taxi & $107\,573$ & 60977.285 &  29.197 \\ 
   \hline
\end{tabular}
\caption{Empirical runtime (in seconds) for the three case study datasets.} 
\end{table}

Note that the runtime of KDE highly depends on the number of evaluation points, the bandwidth selection methods, and whether to use the binned kernel estimation as an approximation to the exact kernel estimation for bandwidth selection and/or density estimation. The runtime we report here is based on the following settings: 1) the number of evaluation points is the same as the number of pixels evaluated by PALM, i.e., the pixels on the fixed grid with the granularity $\epsilon$; 2) the binned approximation  for the plug-in bandwidth selection is used; otherwise it becomes too slow\footnote{It cost more than 10 minutes for the Amsterdam housing data, and more than two days for the amusement park data, both on the full dataset (no splitting for training and testing set).}; 3) the binned approximation for the density estimation is not used. Note that we use these same settings not only for the runtime evaluation, but also for visualizations and calculating the log-likelihood on the testing datasets. 

\subsection{Algorithm settings}
\label{subsec:case_algset}
We now describe some additional algorithm settings for reproducibility for PALM and KDE. 

\smallskip \noindent
\textbf{Kernel density estimation (KDE).} We choose the Gaussian kernel for KDE, the most commonly used kernel by default. We also experiment with several bandwidth selection methods, including both plug-in methods and cross-validation methods. We find that plug-in methods are generally both more stable and much faster in these three cases, and we choose the one that is specifically designed for two-dimensional cases \citep{duong2003plug}. 

Also, we visualize the KDE results by directly plotting the density of each ``pixel"; another common practice is to use a contour function, which will further smooth the KDE results and hence hamper the straightforward comparisons with the PALM results.

\smallskip \noindent
\textbf{PALM.} We set $\epsilon = 1$ and $K_{max} = 100$ for the amusement park dataset, as the amusement park is divided into a $100 \times 100$ grid, so the data is recorded at precision of $1$ and the maximum number of bins cannot exceed $100$.  For the other two datasets, the precision of the dataset is set as $\epsilon = 0.001$, which is roughly 100 meters. During the partitioning step, we set $K_{max} = 300$ to make sure that $\hat{K} < K_{max}$.

\section{Conclusions}
\label{sec:conclusions}
We proposed to discretize two-dimensional data by histograms with far more flexible partitions than adaptive grids, as we observed that the appropriate binning of one dimension may depend on the value of the other dimension. 

Next, we formalized this task based on the MDL principle. Building upon the one-dimensional MDL histogram, we made several technical contributions so as to extend both the formulation and algorithm to the two-dimensional case. Specifically, we solved the problem of calculating the parametric complexity for multi-dimensional cases. Also, we revisited and improved the algorithm for one-dimensional dataset by 1) correcting a minor flaw related to the model encoding, and 2) simplifying the dynamic programming recursion and hence improving the time complexity. 

Further, we proposed a novel heuristic algorithm PALM, which combines the top-down and bottom-up search strategies, and we extensively examined the performance of the PALM algorithm on both synthetic and real-world datasets. That is, we verified our algorithm on various synthetic datasets, and showed that: 1) PALM reveals the ground-truth histogram and converges, in contrast to IPD that produces more and more bins as sample size increases; 2) PALM approximates well to the partitions outside the model class; 3) PALM is self-adaptive to local density structures and sample sizes.

Finally, we applied our algorithm on three diverse real-world spatial datasets, and demonstrated that PALM not only captures more densities changes than KDE, but also fits the unseen data better than KDE, as measured by the log-likelihood. 

\begin{acknowledgement}
This work is part of the research programme 'Human-Guided Data Science by Interactive Model Selection' with project number 612.001.804, which is (partly) financed by the Dutch Research Council (NWO).
\end{acknowledgement}


\section*{Declarations}

\textbf{Funding}: This work is (partly) financed by the Dutch Research Council (NWO).

\noindent 
\textbf{Conflict of interest/Competing interests}: Not applicable. 

\noindent 
\textbf{Ethics approval}: Not applicable.

\noindent 
\textbf{Consent to participate}: Not applicable. 

\noindent 
\textbf{Consent for publication}: Not applicable. 

\noindent 
\textbf{Availability of data and materials}: Two of the datasets used in the case studies are publicly available at http://insideairbnb.com and http://vacommunity.org/VA\\ST+Challenge+2015. The third dataset used in case studies is available upon request at https://gaia.didichuxing.com.
    
\noindent 
\textbf{Code availability}: The code is available at: https://github.com/ylincen/PALM; 

\noindent 
\textbf{Authors' contributions}: LY contributed to theory development, algorithm design and implementation, conducting the experiments and case studies, and writing the manuscript. MB was involved in developing initial ideas, and contributed to the case studies. MvL contributed to problem formalization, algorithm and experiment design, and writing the manuscript. All authors approved the final manuscript. 
\bibliography{main}

\newpage
\section*{Appendix A: Proof that $\text{COMP}(n, \widetilde{S})$ is independent of the number of dimensions (Section 4.1, Proposition 1)}
Assume $S \subset \mathbb{R}^l$, $\widetilde{S}$ is any partition of $S$ with $K$ regions, and $\forall S_j \in \widetilde{S}$, $|S_j|$ represents the (hyper-)volume of $S_j$; for any $y^n$ that can be generated by $\widetilde{S}$, $h_j(y^n)$ denotes the number of data points in region $S_j$. 
\begin{equation}
\begin{split}
    \text{COMP}(n, \widetilde{S})& = \sum_{y^n \in S^n} P(y^n|\widetilde{S}_{\Vec{f} = \hat{\Vec{f}}(y^n)}) \\
    & = \sum_{y^n \in S^n} [\prod_{j = 1}^K (\frac{h_j(y^n) \, \epsilon^l}{n \, |S_j|})^{h_j}] \\
     & = \sum_{h_1+ ... + h_K = n, h_j \geq 0, \forall j} \,\,\,\,\,\,\, \sum_{\{y^n: h_j(y^n) = h_j,\forall j\}} \,\, [\prod_{j = 1}^K (\frac{h_j(y^n) \, \epsilon^l}{n \, |S_j|})^{h_j}]
 \end{split}
\end{equation}
To count the elements in the set ${\{y^n: h_j(y^n) = h_j,\forall j\}}$, we observe that the number of possible ways of distributing $(h_1, ..., h_K)$ data points into each region of $\widetilde{S}$ respectively is
\begin{equation}
    {n \choose h_1} {n - h_1 \choose h_2} \ldots {n - h_1 - \ldots - h_{K-1} \choose h_K} = \frac{n!}{h_1! ... h_K!}\text{.}
\end{equation}
As we assume the precision to be $\epsilon$, for any $S_j$, the number of possible locations for those $h_j(y^n)$ points is equal to $(\frac{|S_j|}{\epsilon^l})^{h_j}$. Thus, the number of elements in the set ${\{y^n: h_j(y^n) = h_j,\forall j\}}$ is
\begin{equation}
    \frac{n!}{h_1! ... h_K!} \prod_{j=1}^K \left(\frac{|S_j|}{\epsilon^l}\right)^{h_j}
\end{equation}
Therefore, 
\begin{equation} \label{eq:regret_proof}
\begin{split}
    \text{COMP}(n, \widetilde{S}) & = \sum_{h_1+ ... + h_K = n} [\frac{n!}{h_1! ... h_K!} \prod_{j=1}^K [(\frac{|S_j|}{\epsilon^l})^{h_j}\prod_{j=1}^K(\frac{h_j \cdot \epsilon^l}{n \cdot |S_j|})^{h_j}] \\
    & = \sum_{h_1+ ... + h_K = n} [\frac{n!}{h_1! ... h_K!} \prod_{j=1}^K [(\frac{|S_j|}{\epsilon^l})^{h_j}(\frac{h_j \cdot \epsilon^l}{n \cdot |S_j|})^{h_j}] \\
     &= \sum_{h_1 + ... + h_K = n} \frac{n!}{h_1! ... h_K!} \prod_{j=1}^K \left(\frac{h_j}{n}\right)^{h_j} \text{,} 
\end{split}
\end{equation}
which completes the proof. 

Note that for continuous data $y^n$, $\text{COMP}(n, \widetilde{S})$ becomes an integral over $y^n \in S^n$, but by the definition of Riemann integral, (which always exists since $\epsilon$ cancels out), the result of $\text{COMP}(n, \widetilde{S})$ is the same as Equation (\ref{eq:regret_proof}).
\section*{Appendix B: Proof that only searching for cut points that are closest to data points is sufficient (Section 6, Proposition 2)}
\label{app:proof2}

Consider one-dimensional data $z^n$, and a partition of the data space $S$, by a set of cut points, denoted as $C^K = \{C_0 = \min z^n, C_1, ..., C_K = \max z^n\}$, the probability of data is
\begin{equation}
    P(z^n|C^K) = \prod_{j = 1}^K (\frac{h_j \, \epsilon}{n \, |S_j|})^{h_j}
\end{equation}
where $h_j$ is the number of data points within the subinterval $S_j$, and $|S_j|$ is the length of the subinterval $S_j$.

We regard $P(x^n|C^K)$ as a \emph{continuous} function of the vector $\Vec{S} = (|S_1|, ..., |S_K|)$, i.e., we forget about the granularity $\epsilon$ for now, and clearly all $h_j$'s are fixed once we fix the $\Vec{S}$. 

On the other hand, if we keep all $h_j$'s fixed, we can still ``move" all the cut points to change $\Vec{S}$ while keeping the $h_j$'s fixed, i.e., we can move a cut point $V_x$ within some closed interval, denoted as $[a,b]$, within which no data points exist.

We prove that the maximum of $P(x^n|C^K)$ will always achieved when $V_x = a$ or $V_x = b$ as we keep other cut points fixed. By doing this, we also prove that, given candidate cut points, we only need to consider cut points that are near to the data points, i.e., if for any candidate cut point, it is another two cut points that are closest to it, other than one or more data points, we can then skip this candidate cut point. 

Since when we move one single cut point, it only affects the subinterval left and right to that cut point, while all other $|S_j|$'s remain the same, it is sufficient to just prove for the case $K = 2$. 

Since now $C_0 = \min_{i \in [n]} x_{i1}$ and $C_2 = \max_{i \in [n]} x_{i1}$, $P(x^n|C^2)$ becomes a function of $C_1$, and equivalently a function of $|S_1|$, where both $C_1$ and $|S_1|$ are bounded as we need to keep $h_1$ and $h_2$ fixed, i.e., 
\begin{equation}
    \log P(x^n|C^2) = \log \left(\left(\frac{\epsilon{h_1}}{n|S_1|}\right)^{h_1} \left(\frac{\epsilon h_2}{n(|S|-|S_1|)}\right)^{h_2} \right) \label{eq:appen}
\end{equation}
where we assume $|S_1| \in [a, b]$ for some certain closed interval $[a,b]$. As we want to maximize $\log P(x^n|C^2)$, it is equivalent to \emph{minimizing}
\begin{equation}
     F(|S_1|) := h_1 \log|S_1| + h_2 \log (|S| - |S_1|)
\end{equation}
as other terms in Equation (\ref{eq:appen}) are constant. Since 
\begin{equation}
    F'(|S_1|) = \frac{h_1(|S|-|S_1|)-h_2|S_1|}{(|S|-|S_1|)|S_1|} \text{,}
\end{equation} 
by setting $F'(|S_1|) = 0$, we have
\begin{equation}
    |S_1|^* = \frac{h_1}{h_1+h_2}L .
\end{equation}
We also have
\begin{equation}
    F''(|S_1|) = \frac{-(h_1+h_2)|S_1|^2 + 2h_1|S||S_1| - h_1|S|^2}{(|S| - |S_1|)^2|S_1|^2} < 0
\end{equation}
because 1) the denominator is always positive apparently, and 2) the numerator is a simple quadratic function which is always negative. The reason is that 1) $-(h_1+h_2)|S_1| < 0$ and 2) the numerator has no real roots, since
\begin{equation}
   (2h_1|S|)^2 - 4(-(h_1+h_2))(h_1|S|^2) = -4h_2h_1|S_1|^2 < 0.
\end{equation}
Therefore, if $|S_1|^* \not\in [a, b]$, $F(|S_1|)$ is monotonic within $[a, b]$; if $|S_1|^* \in [a, b]$, $|S_1|^*$ reaches the \emph{maximum}. In both cases, the minimum of $F(|S_1|)$ will be either $a$ or $b$, which completes the proof. 


%

\section*{Appendix C: IPD visualizations on case study datasets}
\begin{figure}[!h]
    \centering
    \includegraphics[width = 0.63\textwidth]{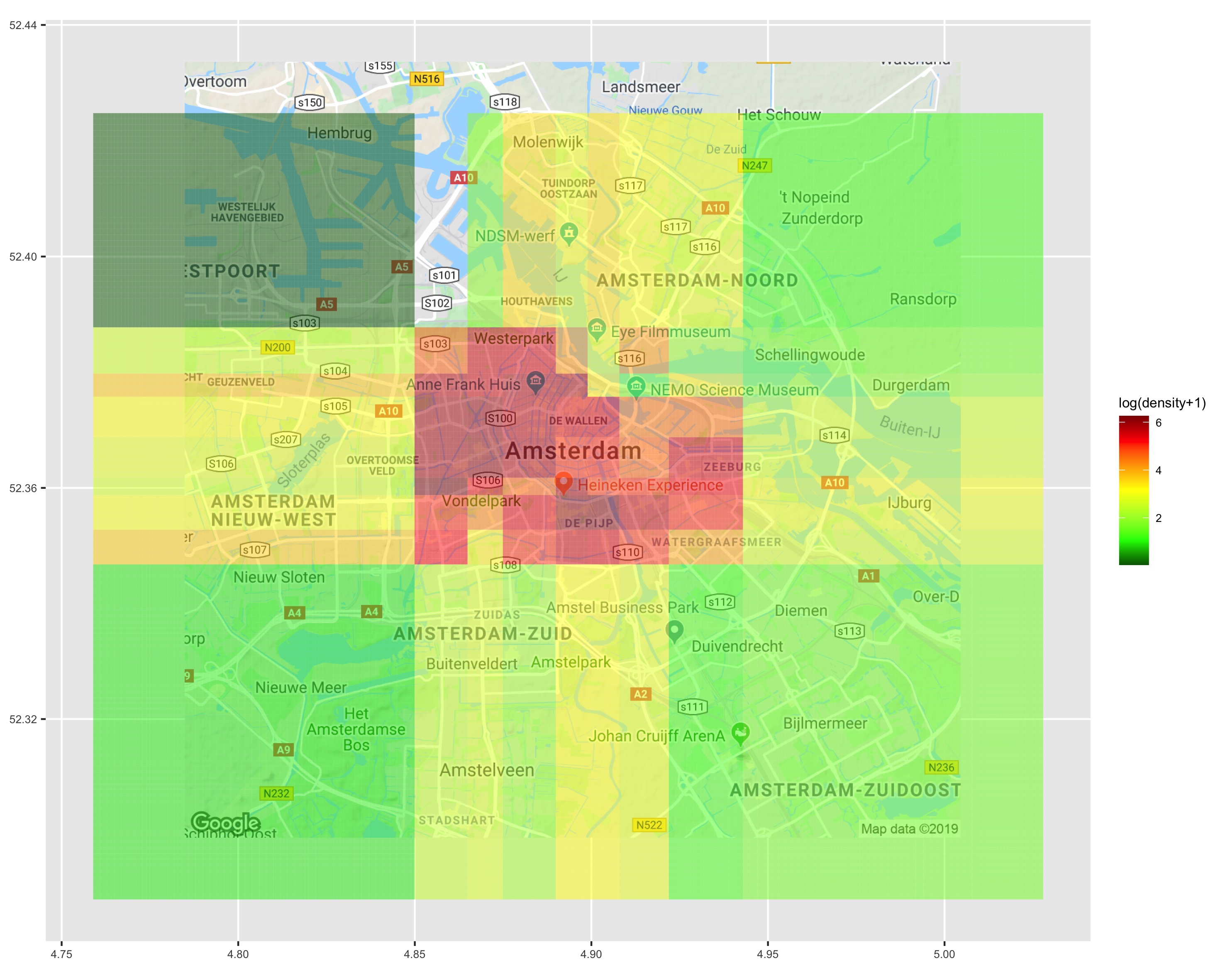}	
    \includegraphics[width = 0.68\textwidth]{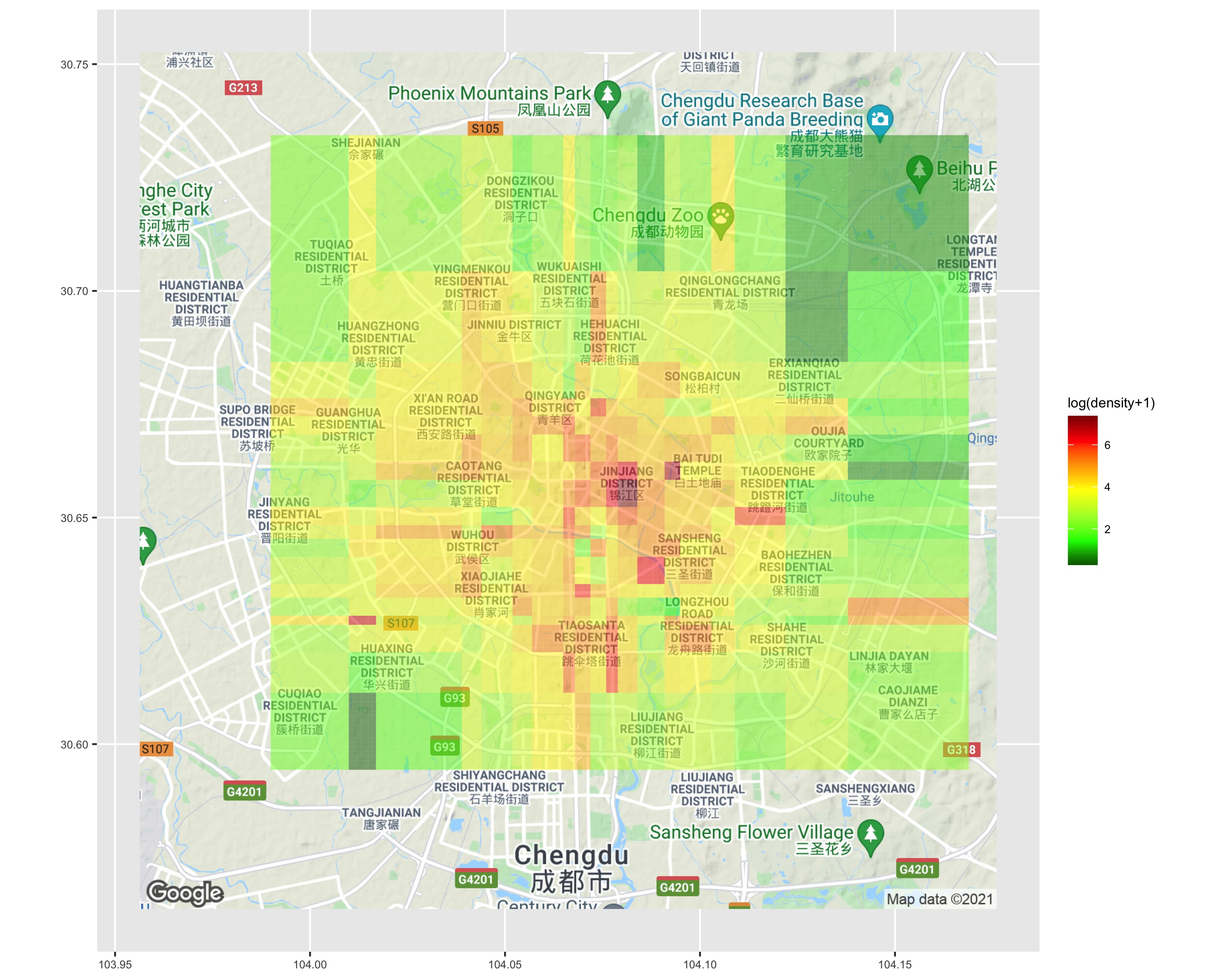}	
    \caption{Visualization of the IPD discretization results on two of the case study datasets (we fail to obtain the result of IPD on the Amusement Park data within four hours).}
    \label{fig:case_ipd}
\end{figure}

\end{document}

%% file: tex_pic_PALMexampleOnHistogramSimulated.tex
\begin{tikzpicture}[x=1pt,y=1pt]
\definecolor{fillColor}{RGB}{255,255,255}
\path[use as bounding box,fill=fillColor,fill opacity=0.00] (0,0) rectangle ( 79.50, 79.50);
\begin{scope}
\path[clip] (  0.00,  0.00) rectangle ( 79.50, 79.50);
\definecolor{drawColor}{RGB}{255,255,255}
\definecolor{fillColor}{RGB}{255,255,255}

\path[draw=drawColor,line width= 0.6pt,line join=round,line cap=round,fill=fillColor] (  0.00,  0.00) rectangle ( 79.50, 79.50);
\end{scope}
\begin{scope}
\path[clip] (  8.25,  8.25) rectangle ( 74.00, 74.00);
\definecolor{drawColor}{RGB}{0,0,0}

\path[draw=drawColor,draw opacity=0.90,line width= 0.6pt,line join=round] ( 11.24, 11.24) -- ( 11.24, 71.01);

\path[draw=drawColor,draw opacity=0.90,line width= 0.6pt,line join=round] ( 17.22, 60.85) -- ( 17.22, 65.03);

\path[draw=drawColor,draw opacity=0.90,line width= 0.6pt,line join=round] ( 17.22, 69.22) -- ( 17.22, 71.01);

\path[draw=drawColor,draw opacity=0.90,line width= 0.6pt,line join=round] ( 25.58, 65.03) -- ( 25.58, 71.01);

\path[draw=drawColor,draw opacity=0.90,line width= 0.6pt,line join=round] ( 53.38, 11.24) -- ( 53.38, 12.43);

\path[draw=drawColor,draw opacity=0.90,line width= 0.6pt,line join=round] ( 53.38, 29.47) -- ( 53.38, 30.07);

\path[draw=drawColor,draw opacity=0.90,line width= 0.6pt,line join=round] ( 53.68, 11.24) -- ( 53.68, 12.43);

\path[draw=drawColor,draw opacity=0.90,line width= 0.6pt,line join=round] ( 53.68, 29.47) -- ( 53.68, 30.07);

\path[draw=drawColor,draw opacity=0.90,line width= 0.6pt,line join=round] ( 58.46, 11.24) -- ( 58.46, 28.27);

\path[draw=drawColor,draw opacity=0.90,line width= 0.6pt,line join=round] ( 71.01, 11.24) -- ( 71.01, 71.01);

\path[draw=drawColor,draw opacity=0.90,line width= 0.6pt,line join=round] ( 11.24, 11.24) -- ( 71.01, 11.24);

\path[draw=drawColor,draw opacity=0.90,line width= 0.6pt,line join=round] ( 53.38, 12.43) -- ( 53.68, 12.43);

\path[draw=drawColor,draw opacity=0.90,line width= 0.6pt,line join=round] ( 58.46, 28.27) -- ( 71.01, 28.27);

\path[draw=drawColor,draw opacity=0.90,line width= 0.6pt,line join=round] ( 53.38, 29.47) -- ( 53.68, 29.47);

\path[draw=drawColor,draw opacity=0.90,line width= 0.6pt,line join=round] ( 53.38, 30.07) -- ( 53.68, 30.07);

\path[draw=drawColor,draw opacity=0.90,line width= 0.6pt,line join=round] ( 11.24, 60.85) -- ( 17.22, 60.85);

\path[draw=drawColor,draw opacity=0.90,line width= 0.6pt,line join=round] ( 17.22, 65.03) -- ( 25.58, 65.03);

\path[draw=drawColor,draw opacity=0.90,line width= 0.6pt,line join=round] ( 17.22, 69.22) -- ( 25.58, 69.22);

\path[draw=drawColor,draw opacity=0.90,line width= 0.6pt,line join=round] ( 11.24, 71.01) -- ( 71.01, 71.01);
\definecolor{drawColor}{RGB}{255,0,0}

\path[draw=drawColor,line width= 0.6pt,dash pattern=on 4pt off 4pt ,line join=round] ( 11.24, 11.24) -- ( 11.24, 71.01);

\path[draw=drawColor,line width= 0.6pt,dash pattern=on 4pt off 4pt ,line join=round] ( 23.67, 64.67) -- ( 23.67, 71.01);

\path[draw=drawColor,line width= 0.6pt,dash pattern=on 4pt off 4pt ,line join=round] ( 58.93, 11.24) -- ( 58.93, 27.68);

\path[draw=drawColor,line width= 0.6pt,dash pattern=on 4pt off 4pt ,line join=round] ( 71.01, 11.24) -- ( 71.01, 71.01);

\path[draw=drawColor,line width= 0.6pt,dash pattern=on 4pt off 4pt ,line join=round] ( 11.24, 11.24) -- ( 71.01, 11.24);

\path[draw=drawColor,line width= 0.6pt,dash pattern=on 4pt off 4pt ,line join=round] ( 58.93, 27.68) -- ( 71.01, 27.68);

\path[draw=drawColor,line width= 0.6pt,dash pattern=on 4pt off 4pt ,line join=round] ( 11.24, 64.67) -- ( 23.67, 64.67);

\path[draw=drawColor,line width= 0.6pt,dash pattern=on 4pt off 4pt ,line join=round] ( 11.24, 71.01) -- ( 71.01, 71.01);
\definecolor{drawColor}{gray}{0.20}

\path[draw=drawColor,line width= 0.6pt,line join=round,line cap=round] (  8.25,  8.25) rectangle ( 74.00, 74.00);
\end{scope}
\begin{scope}
\path[clip] (  0.00,  0.00) rectangle ( 79.50, 79.50);
\definecolor{drawColor}{RGB}{0,0,0}

\path[draw=drawColor,line width= 0.6pt,line join=round] (  8.25,  8.25) --
	(  8.25, 74.00);
\end{scope}
\begin{scope}
\path[clip] (  0.00,  0.00) rectangle ( 79.50, 79.50);
\definecolor{drawColor}{RGB}{0,0,0}

\path[draw=drawColor,line width= 0.6pt,line join=round] (  8.25,  8.25) --
	( 74.00,  8.25);
\end{scope}
\end{tikzpicture}
&
\begin{tikzpicture}[x=1pt,y=1pt]
\definecolor{fillColor}{RGB}{255,255,255}
\path[use as bounding box,fill=fillColor,fill opacity=0.00] (0,0) rectangle ( 79.50, 79.50);
\begin{scope}
\path[clip] (  0.00,  0.00) rectangle ( 79.50, 79.50);
\definecolor{drawColor}{RGB}{255,255,255}
\definecolor{fillColor}{RGB}{255,255,255}

\path[draw=drawColor,line width= 0.6pt,line join=round,line cap=round,fill=fillColor] (  0.00,  0.00) rectangle ( 79.50, 79.50);
\end{scope}
\begin{scope}
\path[clip] (  8.25,  8.25) rectangle ( 74.00, 74.00);
\definecolor{drawColor}{RGB}{0,0,0}

\path[draw=drawColor,draw opacity=0.90,line width= 0.6pt,line join=round] ( 11.24, 11.24) -- ( 11.24, 71.01);

\path[draw=drawColor,draw opacity=0.90,line width= 0.6pt,line join=round] ( 25.88, 11.24) -- ( 25.88, 30.96);

\path[draw=drawColor,draw opacity=0.90,line width= 0.6pt,line join=round] ( 25.88, 34.85) -- ( 25.88, 44.71);

\path[draw=drawColor,draw opacity=0.90,line width= 0.6pt,line join=round] ( 25.88, 60.55) -- ( 25.88, 68.92);

\path[draw=drawColor,draw opacity=0.90,line width= 0.6pt,line join=round] ( 46.20, 11.24) -- ( 46.20, 30.96);

\path[draw=drawColor,draw opacity=0.90,line width= 0.6pt,line join=round] ( 46.20, 68.92) -- ( 46.20, 71.01);

\path[draw=drawColor,draw opacity=0.90,line width= 0.6pt,line join=round] ( 47.10, 13.63) -- ( 47.10, 19.91);

\path[draw=drawColor,draw opacity=0.90,line width= 0.6pt,line join=round] ( 49.49, 58.76) -- ( 49.49, 68.62);

\path[draw=drawColor,draw opacity=0.90,line width= 0.6pt,line join=round] ( 59.95, 11.24) -- ( 59.95, 22.30);

\path[draw=drawColor,draw opacity=0.90,line width= 0.6pt,line join=round] ( 59.95, 58.76) -- ( 59.95, 71.01);

\path[draw=drawColor,draw opacity=0.90,line width= 0.6pt,line join=round] ( 71.01, 11.24) -- ( 71.01, 71.01);

\path[draw=drawColor,draw opacity=0.90,line width= 0.6pt,line join=round] ( 11.24, 11.24) -- ( 71.01, 11.24);

\path[draw=drawColor,draw opacity=0.90,line width= 0.6pt,line join=round] ( 59.95, 13.33) -- ( 71.01, 13.33);

\path[draw=drawColor,draw opacity=0.90,line width= 0.6pt,line join=round] ( 46.20, 13.63) -- ( 47.10, 13.63);

\path[draw=drawColor,draw opacity=0.90,line width= 0.6pt,line join=round] ( 46.20, 19.91) -- ( 47.10, 19.91);

\path[draw=drawColor,draw opacity=0.90,line width= 0.6pt,line join=round] ( 59.95, 22.30) -- ( 71.01, 22.30);

\path[draw=drawColor,draw opacity=0.90,line width= 0.6pt,line join=round] ( 11.24, 26.48) -- ( 25.88, 26.48);

\path[draw=drawColor,draw opacity=0.90,line width= 0.6pt,line join=round] ( 25.88, 30.96) -- ( 46.20, 30.96);

\path[draw=drawColor,draw opacity=0.90,line width= 0.6pt,line join=round] ( 11.24, 34.85) -- ( 25.88, 34.85);

\path[draw=drawColor,draw opacity=0.90,line width= 0.6pt,line join=round] ( 11.24, 44.71) -- ( 25.88, 44.71);

\path[draw=drawColor,draw opacity=0.90,line width= 0.6pt,line join=round] ( 49.49, 58.76) -- ( 59.95, 58.76);

\path[draw=drawColor,draw opacity=0.90,line width= 0.6pt,line join=round] ( 11.24, 60.55) -- ( 25.88, 60.55);

\path[draw=drawColor,draw opacity=0.90,line width= 0.6pt,line join=round] ( 59.95, 64.43) -- ( 71.01, 64.43);

\path[draw=drawColor,draw opacity=0.90,line width= 0.6pt,line join=round] ( 49.49, 68.62) -- ( 59.95, 68.62);

\path[draw=drawColor,draw opacity=0.90,line width= 0.6pt,line join=round] ( 25.88, 68.92) -- ( 46.20, 68.92);

\path[draw=drawColor,draw opacity=0.90,line width= 0.6pt,line join=round] ( 11.24, 71.01) -- ( 71.01, 71.01);
\definecolor{drawColor}{RGB}{255,0,0}

\path[draw=drawColor,line width= 0.6pt,dash pattern=on 4pt off 4pt ,line join=round] ( 11.24, 11.24) -- ( 11.24, 71.01);

\path[draw=drawColor,line width= 0.6pt,dash pattern=on 4pt off 4pt ,line join=round] ( 24.87, 60.25) -- ( 24.87, 68.92);

\path[draw=drawColor,line width= 0.6pt,dash pattern=on 4pt off 4pt ,line join=round] ( 26.24, 26.36) -- ( 26.24, 31.02);

\path[draw=drawColor,line width= 0.6pt,dash pattern=on 4pt off 4pt ,line join=round] ( 47.10, 11.24) -- ( 47.10, 31.02);

\path[draw=drawColor,line width= 0.6pt,dash pattern=on 4pt off 4pt ,line join=round] ( 47.10, 68.92) -- ( 47.10, 71.01);

\path[draw=drawColor,line width= 0.6pt,dash pattern=on 4pt off 4pt ,line join=round] ( 49.79, 58.99) -- ( 49.79, 68.62);

\path[draw=drawColor,line width= 0.6pt,dash pattern=on 4pt off 4pt ,line join=round] ( 49.79, 70.29) -- ( 49.79, 71.01);

\path[draw=drawColor,line width= 0.6pt,dash pattern=on 4pt off 4pt ,line join=round] ( 59.89, 58.99) -- ( 59.89, 71.01);

\path[draw=drawColor,line width= 0.6pt,dash pattern=on 4pt off 4pt ,line join=round] ( 59.95, 11.24) -- ( 59.95, 13.21);

\path[draw=drawColor,line width= 0.6pt,dash pattern=on 4pt off 4pt ,line join=round] ( 71.01, 11.24) -- ( 71.01, 71.01);

\path[draw=drawColor,line width= 0.6pt,dash pattern=on 4pt off 4pt ,line join=round] ( 11.24, 11.24) -- ( 71.01, 11.24);

\path[draw=drawColor,line width= 0.6pt,dash pattern=on 4pt off 4pt ,line join=round] ( 59.95, 13.21) -- ( 71.01, 13.21);

\path[draw=drawColor,line width= 0.6pt,dash pattern=on 4pt off 4pt ,line join=round] ( 11.24, 26.36) -- ( 26.24, 26.36);

\path[draw=drawColor,line width= 0.6pt,dash pattern=on 4pt off 4pt ,line join=round] ( 26.24, 31.02) -- ( 47.10, 31.02);

\path[draw=drawColor,line width= 0.6pt,dash pattern=on 4pt off 4pt ,line join=round] ( 49.79, 58.99) -- ( 59.89, 58.99);

\path[draw=drawColor,line width= 0.6pt,dash pattern=on 4pt off 4pt ,line join=round] ( 11.24, 60.25) -- ( 24.87, 60.25);

\path[draw=drawColor,line width= 0.6pt,dash pattern=on 4pt off 4pt ,line join=round] ( 59.89, 64.13) -- ( 71.01, 64.13);

\path[draw=drawColor,line width= 0.6pt,dash pattern=on 4pt off 4pt ,line join=round] ( 49.79, 68.62) -- ( 59.89, 68.62);

\path[draw=drawColor,line width= 0.6pt,dash pattern=on 4pt off 4pt ,line join=round] ( 24.87, 68.92) -- ( 47.10, 68.92);

\path[draw=drawColor,line width= 0.6pt,dash pattern=on 4pt off 4pt ,line join=round] ( 47.10, 70.29) -- ( 49.79, 70.29);

\path[draw=drawColor,line width= 0.6pt,dash pattern=on 4pt off 4pt ,line join=round] ( 11.24, 71.01) -- ( 47.10, 71.01);

\path[draw=drawColor,line width= 0.6pt,dash pattern=on 4pt off 4pt ,line join=round] ( 49.79, 71.01) -- ( 71.01, 71.01);
\definecolor{drawColor}{gray}{0.20}

\path[draw=drawColor,line width= 0.6pt,line join=round,line cap=round] (  8.25,  8.25) rectangle ( 74.00, 74.00);
\end{scope}
\begin{scope}
\path[clip] (  0.00,  0.00) rectangle ( 79.50, 79.50);
\definecolor{drawColor}{RGB}{0,0,0}

\path[draw=drawColor,line width= 0.6pt,line join=round] (  8.25,  8.25) --
	(  8.25, 74.00);
\end{scope}
\begin{scope}
\path[clip] (  0.00,  0.00) rectangle ( 79.50, 79.50);
\definecolor{drawColor}{RGB}{0,0,0}

\path[draw=drawColor,line width= 0.6pt,line join=round] (  8.25,  8.25) --
	( 74.00,  8.25);
\end{scope}
\end{tikzpicture}
&
\begin{tikzpicture}[x=1pt,y=1pt]
\definecolor{fillColor}{RGB}{255,255,255}
\path[use as bounding box,fill=fillColor,fill opacity=0.00] (0,0) rectangle ( 79.50, 79.50);
\begin{scope}
\path[clip] (  0.00,  0.00) rectangle ( 79.50, 79.50);
\definecolor{drawColor}{RGB}{255,255,255}
\definecolor{fillColor}{RGB}{255,255,255}

\path[draw=drawColor,line width= 0.6pt,line join=round,line cap=round,fill=fillColor] (  0.00,  0.00) rectangle ( 79.50, 79.50);
\end{scope}
\begin{scope}
\path[clip] (  8.25,  8.25) rectangle ( 74.00, 74.00);
\definecolor{drawColor}{RGB}{0,0,0}

\path[draw=drawColor,draw opacity=0.90,line width= 0.6pt,line join=round] ( 11.24, 11.24) -- ( 11.24, 71.01);

\path[draw=drawColor,draw opacity=0.90,line width= 0.6pt,line join=round] ( 17.22, 11.24) -- ( 17.22, 13.03);

\path[draw=drawColor,draw opacity=0.90,line width= 0.6pt,line join=round] ( 17.22, 14.53) -- ( 17.22, 22.30);

\path[draw=drawColor,draw opacity=0.90,line width= 0.6pt,line join=round] ( 19.31, 13.03) -- ( 19.31, 28.87);

\path[draw=drawColor,draw opacity=0.90,line width= 0.6pt,line join=round] ( 33.35, 11.24) -- ( 33.35, 20.80);

\path[draw=drawColor,draw opacity=0.90,line width= 0.6pt,line join=round] ( 33.35, 28.87) -- ( 33.35, 36.94);

\path[draw=drawColor,draw opacity=0.90,line width= 0.6pt,line join=round] ( 33.35, 54.57) -- ( 33.35, 71.01);

\path[draw=drawColor,draw opacity=0.90,line width= 0.6pt,line join=round] ( 45.61, 13.63) -- ( 45.61, 16.32);

\path[draw=drawColor,draw opacity=0.90,line width= 0.6pt,line join=round] ( 45.61, 20.80) -- ( 45.61, 36.94);

\path[draw=drawColor,draw opacity=0.90,line width= 0.6pt,line join=round] ( 45.61, 54.57) -- ( 45.61, 71.01);

\path[draw=drawColor,draw opacity=0.90,line width= 0.6pt,line join=round] ( 69.51, 11.24) -- ( 69.51, 13.63);

\path[draw=drawColor,draw opacity=0.90,line width= 0.6pt,line join=round] ( 69.51, 15.12) -- ( 69.51, 24.09);

\path[draw=drawColor,draw opacity=0.90,line width= 0.6pt,line join=round] ( 69.51, 54.87) -- ( 69.51, 56.36);

\path[draw=drawColor,draw opacity=0.90,line width= 0.6pt,line join=round] ( 69.51, 66.23) -- ( 69.51, 71.01);

\path[draw=drawColor,draw opacity=0.90,line width= 0.6pt,line join=round] ( 71.01, 11.24) -- ( 71.01, 71.01);

\path[draw=drawColor,draw opacity=0.90,line width= 0.6pt,line join=round] ( 11.24, 11.24) -- ( 71.01, 11.24);

\path[draw=drawColor,draw opacity=0.90,line width= 0.6pt,line join=round] ( 17.22, 13.03) -- ( 19.31, 13.03);

\path[draw=drawColor,draw opacity=0.90,line width= 0.6pt,line join=round] ( 45.61, 13.63) -- ( 69.51, 13.63);

\path[draw=drawColor,draw opacity=0.90,line width= 0.6pt,line join=round] ( 11.24, 14.53) -- ( 17.22, 14.53);

\path[draw=drawColor,draw opacity=0.90,line width= 0.6pt,line join=round] ( 45.61, 15.12) -- ( 69.51, 15.12);

\path[draw=drawColor,draw opacity=0.90,line width= 0.6pt,line join=round] ( 33.35, 16.32) -- ( 45.61, 16.32);

\path[draw=drawColor,draw opacity=0.90,line width= 0.6pt,line join=round] ( 11.24, 20.80) -- ( 17.22, 20.80);

\path[draw=drawColor,draw opacity=0.90,line width= 0.6pt,line join=round] ( 33.35, 20.80) -- ( 45.61, 20.80);

\path[draw=drawColor,draw opacity=0.90,line width= 0.6pt,line join=round] ( 11.24, 22.30) -- ( 17.22, 22.30);

\path[draw=drawColor,draw opacity=0.90,line width= 0.6pt,line join=round] ( 45.61, 22.30) -- ( 69.51, 22.30);

\path[draw=drawColor,draw opacity=0.90,line width= 0.6pt,line join=round] ( 45.61, 24.09) -- ( 69.51, 24.09);

\path[draw=drawColor,draw opacity=0.90,line width= 0.6pt,line join=round] ( 19.31, 28.87) -- ( 33.35, 28.87);

\path[draw=drawColor,draw opacity=0.90,line width= 0.6pt,line join=round] ( 33.35, 36.94) -- ( 45.61, 36.94);

\path[draw=drawColor,draw opacity=0.90,line width= 0.6pt,line join=round] ( 33.35, 54.57) -- ( 45.61, 54.57);

\path[draw=drawColor,draw opacity=0.90,line width= 0.6pt,line join=round] ( 69.51, 54.87) -- ( 71.01, 54.87);

\path[draw=drawColor,draw opacity=0.90,line width= 0.6pt,line join=round] ( 69.51, 56.36) -- ( 71.01, 56.36);

\path[draw=drawColor,draw opacity=0.90,line width= 0.6pt,line join=round] ( 69.51, 66.23) -- ( 71.01, 66.23);

\path[draw=drawColor,draw opacity=0.90,line width= 0.6pt,line join=round] ( 11.24, 71.01) -- ( 71.01, 71.01);
\definecolor{drawColor}{RGB}{255,0,0}

\path[draw=drawColor,line width= 0.6pt,dash pattern=on 4pt off 4pt ,line join=round] ( 11.24, 11.24) -- ( 11.24, 71.01);

\path[draw=drawColor,line width= 0.6pt,dash pattern=on 4pt off 4pt ,line join=round] ( 20.44, 11.24) -- ( 20.44, 28.87);

\path[draw=drawColor,line width= 0.6pt,dash pattern=on 4pt off 4pt ,line join=round] ( 33.59, 11.24) -- ( 33.59, 16.62);

\path[draw=drawColor,line width= 0.6pt,dash pattern=on 4pt off 4pt ,line join=round] ( 33.59, 28.87) -- ( 33.59, 36.94);

\path[draw=drawColor,line width= 0.6pt,dash pattern=on 4pt off 4pt ,line join=round] ( 33.59, 52.18) -- ( 33.59, 71.01);

\path[draw=drawColor,line width= 0.6pt,dash pattern=on 4pt off 4pt ,line join=round] ( 45.61, 13.75) -- ( 45.61, 36.94);

\path[draw=drawColor,line width= 0.6pt,dash pattern=on 4pt off 4pt ,line join=round] ( 45.61, 52.18) -- ( 45.61, 70.95);

\path[draw=drawColor,line width= 0.6pt,dash pattern=on 4pt off 4pt ,line join=round] ( 45.61, 11.24) -- ( 45.61, 11.36);

\path[draw=drawColor,line width= 0.6pt,dash pattern=on 4pt off 4pt ,line join=round] ( 45.61, 70.95) -- ( 45.61, 71.01);

\path[draw=drawColor,line width= 0.6pt,dash pattern=on 4pt off 4pt ,line join=round] ( 69.39, 11.24) -- ( 69.39, 13.75);

\path[draw=drawColor,line width= 0.6pt,dash pattern=on 4pt off 4pt ,line join=round] ( 69.81, 15.12) -- ( 69.81, 22.42);

\path[draw=drawColor,line width= 0.6pt,dash pattern=on 4pt off 4pt ,line join=round] ( 71.01, 11.24) -- ( 71.01, 71.01);

\path[draw=drawColor,line width= 0.6pt,dash pattern=on 4pt off 4pt ,line join=round] ( 11.24, 11.24) -- ( 33.59, 11.24);

\path[draw=drawColor,line width= 0.6pt,dash pattern=on 4pt off 4pt ,line join=round] ( 45.61, 11.24) -- ( 71.01, 11.24);

\path[draw=drawColor,line width= 0.6pt,dash pattern=on 4pt off 4pt ,line join=round] ( 33.59, 11.36) -- ( 45.61, 11.36);

\path[draw=drawColor,line width= 0.6pt,dash pattern=on 4pt off 4pt ,line join=round] ( 45.61, 13.75) -- ( 69.39, 13.75);

\path[draw=drawColor,line width= 0.6pt,dash pattern=on 4pt off 4pt ,line join=round] ( 45.61, 15.12) -- ( 69.81, 15.12);

\path[draw=drawColor,line width= 0.6pt,dash pattern=on 4pt off 4pt ,line join=round] ( 33.59, 16.62) -- ( 45.61, 16.62);

\path[draw=drawColor,line width= 0.6pt,dash pattern=on 4pt off 4pt ,line join=round] ( 45.61, 22.42) -- ( 69.81, 22.42);

\path[draw=drawColor,line width= 0.6pt,dash pattern=on 4pt off 4pt ,line join=round] ( 11.24, 23.31) -- ( 20.44, 23.31);

\path[draw=drawColor,line width= 0.6pt,dash pattern=on 4pt off 4pt ,line join=round] ( 20.44, 28.87) -- ( 33.59, 28.87);

\path[draw=drawColor,line width= 0.6pt,dash pattern=on 4pt off 4pt ,line join=round] ( 33.59, 36.94) -- ( 45.61, 36.94);

\path[draw=drawColor,line width= 0.6pt,dash pattern=on 4pt off 4pt ,line join=round] ( 33.59, 52.18) -- ( 45.61, 52.18);

\path[draw=drawColor,line width= 0.6pt,dash pattern=on 4pt off 4pt ,line join=round] ( 33.59, 70.95) -- ( 45.61, 70.95);

\path[draw=drawColor,line width= 0.6pt,dash pattern=on 4pt off 4pt ,line join=round] ( 11.24, 71.01) -- ( 33.59, 71.01);

\path[draw=drawColor,line width= 0.6pt,dash pattern=on 4pt off 4pt ,line join=round] ( 45.61, 71.01) -- ( 71.01, 71.01);
\definecolor{drawColor}{gray}{0.20}

\path[draw=drawColor,line width= 0.6pt,line join=round,line cap=round] (  8.25,  8.25) rectangle ( 74.00, 74.00);
\end{scope}
\begin{scope}
\path[clip] (  0.00,  0.00) rectangle ( 79.50, 79.50);
\definecolor{drawColor}{RGB}{0,0,0}

\path[draw=drawColor,line width= 0.6pt,line join=round] (  8.25,  8.25) --
	(  8.25, 74.00);
\end{scope}
\begin{scope}
\path[clip] (  0.00,  0.00) rectangle ( 79.50, 79.50);
\definecolor{drawColor}{RGB}{0,0,0}

\path[draw=drawColor,line width= 0.6pt,line join=round] (  8.25,  8.25) --
	( 74.00,  8.25);
\end{scope}
\end{tikzpicture}
 & 
\begin{tikzpicture}[x=1pt,y=1pt]
\definecolor{fillColor}{RGB}{255,255,255}
\path[use as bounding box,fill=fillColor,fill opacity=0.00] (0,0) rectangle ( 79.50, 79.50);
\begin{scope}
\path[clip] (  0.00,  0.00) rectangle ( 79.50, 79.50);
\definecolor{drawColor}{RGB}{255,255,255}
\definecolor{fillColor}{RGB}{255,255,255}

\path[draw=drawColor,line width= 0.6pt,line join=round,line cap=round,fill=fillColor] (  0.00,  0.00) rectangle ( 79.50, 79.50);
\end{scope}
\begin{scope}
\path[clip] (  8.25,  8.25) rectangle ( 74.00, 74.00);
\definecolor{drawColor}{RGB}{0,0,0}

\path[draw=drawColor,draw opacity=0.90,line width= 0.6pt,line join=round] ( 11.24, 11.24) -- ( 11.24, 71.01);

\path[draw=drawColor,draw opacity=0.90,line width= 0.6pt,line join=round] ( 13.33, 34.85) -- ( 13.33, 35.74);

\path[draw=drawColor,draw opacity=0.90,line width= 0.6pt,line join=round] ( 19.61, 11.24) -- ( 19.61, 15.72);

\path[draw=drawColor,draw opacity=0.90,line width= 0.6pt,line join=round] ( 24.69, 15.72) -- ( 24.69, 55.77);

\path[draw=drawColor,draw opacity=0.90,line width= 0.6pt,line join=round] ( 38.43, 14.53) -- ( 38.43, 32.76);

\path[draw=drawColor,draw opacity=0.90,line width= 0.6pt,line join=round] ( 38.43, 47.70) -- ( 38.43, 55.77);

\path[draw=drawColor,draw opacity=0.90,line width= 0.6pt,line join=round] ( 71.01, 11.24) -- ( 71.01, 71.01);

\path[draw=drawColor,draw opacity=0.90,line width= 0.6pt,line join=round] ( 11.24, 11.24) -- ( 71.01, 11.24);

\path[draw=drawColor,draw opacity=0.90,line width= 0.6pt,line join=round] ( 38.43, 14.53) -- ( 71.01, 14.53);

\path[draw=drawColor,draw opacity=0.90,line width= 0.6pt,line join=round] ( 19.61, 15.72) -- ( 24.69, 15.72);

\path[draw=drawColor,draw opacity=0.90,line width= 0.6pt,line join=round] ( 24.69, 22.89) -- ( 38.43, 22.89);

\path[draw=drawColor,draw opacity=0.90,line width= 0.6pt,line join=round] ( 38.43, 32.76) -- ( 71.01, 32.76);

\path[draw=drawColor,draw opacity=0.90,line width= 0.6pt,line join=round] ( 11.24, 34.85) -- ( 13.33, 34.85);

\path[draw=drawColor,draw opacity=0.90,line width= 0.6pt,line join=round] ( 11.24, 35.74) -- ( 13.33, 35.74);

\path[draw=drawColor,draw opacity=0.90,line width= 0.6pt,line join=round] ( 38.43, 47.70) -- ( 71.01, 47.70);

\path[draw=drawColor,draw opacity=0.90,line width= 0.6pt,line join=round] ( 24.69, 55.77) -- ( 38.43, 55.77);

\path[draw=drawColor,draw opacity=0.90,line width= 0.6pt,line join=round] ( 11.24, 71.01) -- ( 71.01, 71.01);
\definecolor{drawColor}{RGB}{255,0,0}

\path[draw=drawColor,line width= 0.6pt,dash pattern=on 4pt off 4pt ,line join=round] ( 11.24, 11.24) -- ( 11.24, 71.01);

\path[draw=drawColor,line width= 0.6pt,dash pattern=on 4pt off 4pt ,line join=round] ( 19.55, 11.24) -- ( 19.55, 15.72);

\path[draw=drawColor,line width= 0.6pt,dash pattern=on 4pt off 4pt ,line join=round] ( 24.87, 12.97) -- ( 24.87, 47.70);

\path[draw=drawColor,line width= 0.6pt,dash pattern=on 4pt off 4pt ,line join=round] ( 38.31, 22.89) -- ( 38.31, 32.76);

\path[draw=drawColor,line width= 0.6pt,dash pattern=on 4pt off 4pt ,line join=round] ( 71.01, 11.24) -- ( 71.01, 71.01);

\path[draw=drawColor,line width= 0.6pt,dash pattern=on 4pt off 4pt ,line join=round] ( 11.24, 11.24) -- ( 71.01, 11.24);

\path[draw=drawColor,line width= 0.6pt,dash pattern=on 4pt off 4pt ,line join=round] ( 24.87, 12.97) -- ( 71.01, 12.97);

\path[draw=drawColor,line width= 0.6pt,dash pattern=on 4pt off 4pt ,line join=round] ( 19.55, 15.72) -- ( 24.87, 15.72);

\path[draw=drawColor,line width= 0.6pt,dash pattern=on 4pt off 4pt ,line join=round] ( 24.87, 22.89) -- ( 71.01, 22.89);

\path[draw=drawColor,line width= 0.6pt,dash pattern=on 4pt off 4pt ,line join=round] ( 38.31, 32.76) -- ( 71.01, 32.76);

\path[draw=drawColor,line width= 0.6pt,dash pattern=on 4pt off 4pt ,line join=round] ( 24.87, 47.70) -- ( 71.01, 47.70);

\path[draw=drawColor,line width= 0.6pt,dash pattern=on 4pt off 4pt ,line join=round] ( 11.24, 71.01) -- ( 71.01, 71.01);
\definecolor{drawColor}{gray}{0.20}

\path[draw=drawColor,line width= 0.6pt,line join=round,line cap=round] (  8.25,  8.25) rectangle ( 74.00, 74.00);
\end{scope}
\begin{scope}
\path[clip] (  0.00,  0.00) rectangle ( 79.50, 79.50);
\definecolor{drawColor}{RGB}{0,0,0}

\path[draw=drawColor,line width= 0.6pt,line join=round] (  8.25,  8.25) --
	(  8.25, 74.00);
\end{scope}
\begin{scope}
\path[clip] (  0.00,  0.00) rectangle ( 79.50, 79.50);
\definecolor{drawColor}{RGB}{0,0,0}

\path[draw=drawColor,line width= 0.6pt,line join=round] (  8.25,  8.25) --
	( 74.00,  8.25);
\end{scope}
\end{tikzpicture}

%% file: fig_MISE.tex
\begin{tikzpicture}[x=1pt,y=1pt]
\definecolor{fillColor}{RGB}{255,255,255}
\path[use as bounding box,fill=fillColor,fill opacity=0.00] (0,0) rectangle (332.44,108.41);
\begin{scope}
\path[clip] (  0.00,  0.00) rectangle (332.44,108.41);
\definecolor{drawColor}{RGB}{255,255,255}
\definecolor{fillColor}{RGB}{255,255,255}

\path[draw=drawColor,line width= 0.6pt,line join=round,line cap=round,fill=fillColor] (  0.00,  0.00) rectangle (332.44,108.41);
\end{scope}
\begin{scope}
\path[clip] ( 34.48, 40.66) rectangle (326.94,102.90);
\definecolor{fillColor}{RGB}{255,255,255}

\path[fill=fillColor] ( 34.48, 40.66) rectangle (326.94,102.90);
\definecolor{drawColor}{gray}{0.92}

\path[draw=drawColor,line width= 0.3pt,line join=round] ( 34.48, 49.34) --
	(326.94, 49.34);

\path[draw=drawColor,line width= 0.3pt,line join=round] ( 34.48, 61.04) --
	(326.94, 61.04);

\path[draw=drawColor,line width= 0.3pt,line join=round] ( 34.48, 72.75) --
	(326.94, 72.75);

\path[draw=drawColor,line width= 0.3pt,line join=round] ( 34.48, 84.45) --
	(326.94, 84.45);

\path[draw=drawColor,line width= 0.3pt,line join=round] ( 34.48, 96.15) --
	(326.94, 96.15);

\path[draw=drawColor,line width= 0.3pt,line join=round] ( 34.83, 40.66) --
	( 34.83,102.90);

\path[draw=drawColor,line width= 0.3pt,line join=round] ( 41.30, 40.66) --
	( 41.30,102.90);

\path[draw=drawColor,line width= 0.3pt,line join=round] ( 54.25, 40.66) --
	( 54.25,102.90);

\path[draw=drawColor,line width= 0.3pt,line join=round] ( 67.59, 40.66) --
	( 67.59,102.90);

\path[draw=drawColor,line width= 0.3pt,line join=round] ( 95.60, 40.66) --
	( 95.60,102.90);

\path[draw=drawColor,line width= 0.3pt,line join=round] (126.57, 40.66) --
	(126.57,102.90);

\path[draw=drawColor,line width= 0.3pt,line join=round] (142.88, 40.66) --
	(142.88,102.90);

\path[draw=drawColor,line width= 0.3pt,line join=round] (156.22, 40.66) --
	(156.22,102.90);

\path[draw=drawColor,line width= 0.3pt,line join=round] (184.22, 40.66) --
	(184.22,102.90);

\path[draw=drawColor,line width= 0.3pt,line join=round] (215.19, 40.66) --
	(215.19,102.90);

\path[draw=drawColor,line width= 0.3pt,line join=round] (231.50, 40.66) --
	(231.50,102.90);

\path[draw=drawColor,line width= 0.3pt,line join=round] (244.84, 40.66) --
	(244.84,102.90);

\path[draw=drawColor,line width= 0.3pt,line join=round] (272.85, 40.66) --
	(272.85,102.90);

\path[draw=drawColor,line width= 0.3pt,line join=round] (303.82, 40.66) --
	(303.82,102.90);

\path[draw=drawColor,line width= 0.3pt,line join=round] (320.12, 40.66) --
	(320.12,102.90);

\path[draw=drawColor,line width= 0.3pt,line join=round] (326.60, 40.66) --
	(326.60,102.90);

\path[draw=drawColor,line width= 0.6pt,line join=round] ( 34.48, 43.49) --
	(326.94, 43.49);

\path[draw=drawColor,line width= 0.6pt,line join=round] ( 34.48, 55.19) --
	(326.94, 55.19);

\path[draw=drawColor,line width= 0.6pt,line join=round] ( 34.48, 66.89) --
	(326.94, 66.89);

\path[draw=drawColor,line width= 0.6pt,line join=round] ( 34.48, 78.60) --
	(326.94, 78.60);

\path[draw=drawColor,line width= 0.6pt,line join=round] ( 34.48, 90.30) --
	(326.94, 90.30);

\path[draw=drawColor,line width= 0.6pt,line join=round] ( 34.48,102.01) --
	(326.94,102.01);

\path[draw=drawColor,line width= 0.6pt,line join=round] ( 47.78, 40.66) --
	( 47.78,102.90);

\path[draw=drawColor,line width= 0.6pt,line join=round] ( 60.73, 40.66) --
	( 60.73,102.90);

\path[draw=drawColor,line width= 0.6pt,line join=round] ( 74.46, 40.66) --
	( 74.46,102.90);

\path[draw=drawColor,line width= 0.6pt,line join=round] (116.74, 40.66) --
	(116.74,102.90);

\path[draw=drawColor,line width= 0.6pt,line join=round] (136.40, 40.66) --
	(136.40,102.90);

\path[draw=drawColor,line width= 0.6pt,line join=round] (149.35, 40.66) --
	(149.35,102.90);

\path[draw=drawColor,line width= 0.6pt,line join=round] (163.08, 40.66) --
	(163.08,102.90);

\path[draw=drawColor,line width= 0.6pt,line join=round] (205.36, 40.66) --
	(205.36,102.90);

\path[draw=drawColor,line width= 0.6pt,line join=round] (225.02, 40.66) --
	(225.02,102.90);

\path[draw=drawColor,line width= 0.6pt,line join=round] (237.98, 40.66) --
	(237.98,102.90);

\path[draw=drawColor,line width= 0.6pt,line join=round] (251.70, 40.66) --
	(251.70,102.90);

\path[draw=drawColor,line width= 0.6pt,line join=round] (293.99, 40.66) --
	(293.99,102.90);

\path[draw=drawColor,line width= 0.6pt,line join=round] (313.65, 40.66) --
	(313.65,102.90);
\definecolor{drawColor}{RGB}{0,0,0}
\definecolor{fillColor}{RGB}{0,0,0}

\path[draw=drawColor,line width= 0.4pt,line join=round,line cap=round,fill=fillColor] ( 47.78, 67.64) circle (  0.89);

\path[draw=drawColor,line width= 0.4pt,line join=round,line cap=round,fill=fillColor] ( 60.73, 63.33) circle (  0.89);

\path[draw=drawColor,line width= 0.4pt,line join=round,line cap=round,fill=fillColor] ( 74.46, 57.16) circle (  0.89);

\path[draw=drawColor,line width= 0.4pt,line join=round,line cap=round,fill=fillColor] (116.74, 48.35) circle (  0.89);

\path[draw=drawColor,line width= 0.4pt,line join=round,line cap=round,fill=fillColor] (136.40, 46.91) circle (  0.89);

\path[draw=drawColor,line width= 0.4pt,line join=round,line cap=round,fill=fillColor] (149.35, 45.95) circle (  0.89);

\path[draw=drawColor,line width= 0.4pt,line join=round,line cap=round,fill=fillColor] (163.08, 45.21) circle (  0.89);

\path[draw=drawColor,line width= 0.4pt,line join=round,line cap=round,fill=fillColor] (205.36, 44.35) circle (  0.89);

\path[draw=drawColor,line width= 0.4pt,line join=round,line cap=round,fill=fillColor] (225.02, 44.24) circle (  0.89);

\path[draw=drawColor,line width= 0.4pt,line join=round,line cap=round,fill=fillColor] (237.98, 44.29) circle (  0.89);

\path[draw=drawColor,line width= 0.4pt,line join=round,line cap=round,fill=fillColor] (251.70, 44.16) circle (  0.89);

\path[draw=drawColor,line width= 0.4pt,line join=round,line cap=round,fill=fillColor] (293.99, 44.13) circle (  0.89);

\path[draw=drawColor,line width= 0.4pt,line join=round,line cap=round,fill=fillColor] (313.65, 44.08) circle (  0.89);

\path[draw=drawColor,line width= 0.6pt,line join=round] ( 47.78, 67.64) --
	( 60.73, 63.33) --
	( 74.46, 57.16) --
	(116.74, 48.35) --
	(136.40, 46.91) --
	(149.35, 45.95) --
	(163.08, 45.21) --
	(205.36, 44.35) --
	(225.02, 44.24) --
	(237.98, 44.29) --
	(251.70, 44.16) --
	(293.99, 44.13) --
	(313.65, 44.08);
\definecolor{fillColor}{RGB}{135,206,235}

\path[fill=fillColor,fill opacity=0.70] ( 47.78,100.08) --
	( 60.73, 86.17) --
	( 74.46, 74.09) --
	(116.74, 55.17) --
	(136.40, 52.08) --
	(149.35, 49.77) --
	(163.08, 47.67) --
	(205.36, 46.04) --
	(225.02, 45.54) --
	(237.98, 45.59) --
	(251.70, 46.03) --
	(293.99, 46.39) --
	(313.65, 45.91) --
	(313.65, 43.49) --
	(293.99, 43.49) --
	(251.70, 43.49) --
	(237.98, 43.49) --
	(225.02, 43.52) --
	(205.36, 43.54) --
	(163.08, 43.74) --
	(149.35, 43.92) --
	(136.40, 44.08) --
	(116.74, 44.24) --
	( 74.46, 45.73) --
	( 60.73, 46.53) --
	( 47.78, 46.85) --
	cycle;
\definecolor{drawColor}{gray}{0.20}

\path[draw=drawColor,line width= 0.6pt,line join=round,line cap=round] ( 34.48, 40.66) rectangle (326.94,102.90);
\end{scope}
\begin{scope}
\path[clip] (  0.00,  0.00) rectangle (332.44,108.41);
\definecolor{drawColor}{gray}{0.30}

\node[text=drawColor,anchor=base east,inner sep=0pt, outer sep=0pt, scale=  0.80] at ( 29.53, 40.73) {0.00};

\node[text=drawColor,anchor=base east,inner sep=0pt, outer sep=0pt, scale=  0.80] at ( 29.53, 52.43) {0.01};

\node[text=drawColor,anchor=base east,inner sep=0pt, outer sep=0pt, scale=  0.80] at ( 29.53, 64.14) {0.02};

\node[text=drawColor,anchor=base east,inner sep=0pt, outer sep=0pt, scale=  0.80] at ( 29.53, 75.84) {0.03};

\node[text=drawColor,anchor=base east,inner sep=0pt, outer sep=0pt, scale=  0.80] at ( 29.53, 87.55) {0.04};

\node[text=drawColor,anchor=base east,inner sep=0pt, outer sep=0pt, scale=  0.80] at ( 29.53, 99.25) {0.05};
\end{scope}
\begin{scope}
\path[clip] (  0.00,  0.00) rectangle (332.44,108.41);
\definecolor{drawColor}{gray}{0.20}

\path[draw=drawColor,line width= 0.6pt,line join=round] ( 31.73, 43.49) --
	( 34.48, 43.49);

\path[draw=drawColor,line width= 0.6pt,line join=round] ( 31.73, 55.19) --
	( 34.48, 55.19);

\path[draw=drawColor,line width= 0.6pt,line join=round] ( 31.73, 66.89) --
	( 34.48, 66.89);

\path[draw=drawColor,line width= 0.6pt,line join=round] ( 31.73, 78.60) --
	( 34.48, 78.60);

\path[draw=drawColor,line width= 0.6pt,line join=round] ( 31.73, 90.30) --
	( 34.48, 90.30);

\path[draw=drawColor,line width= 0.6pt,line join=round] ( 31.73,102.01) --
	( 34.48,102.01);
\end{scope}
\begin{scope}
\path[clip] (  0.00,  0.00) rectangle (332.44,108.41);
\definecolor{drawColor}{gray}{0.20}

\path[draw=drawColor,line width= 0.6pt,line join=round] ( 47.78, 37.91) --
	( 47.78, 40.66);

\path[draw=drawColor,line width= 0.6pt,line join=round] ( 60.73, 37.91) --
	( 60.73, 40.66);

\path[draw=drawColor,line width= 0.6pt,line join=round] ( 74.46, 37.91) --
	( 74.46, 40.66);

\path[draw=drawColor,line width= 0.6pt,line join=round] (116.74, 37.91) --
	(116.74, 40.66);

\path[draw=drawColor,line width= 0.6pt,line join=round] (136.40, 37.91) --
	(136.40, 40.66);

\path[draw=drawColor,line width= 0.6pt,line join=round] (149.35, 37.91) --
	(149.35, 40.66);

\path[draw=drawColor,line width= 0.6pt,line join=round] (163.08, 37.91) --
	(163.08, 40.66);

\path[draw=drawColor,line width= 0.6pt,line join=round] (205.36, 37.91) --
	(205.36, 40.66);

\path[draw=drawColor,line width= 0.6pt,line join=round] (225.02, 37.91) --
	(225.02, 40.66);

\path[draw=drawColor,line width= 0.6pt,line join=round] (237.98, 37.91) --
	(237.98, 40.66);

\path[draw=drawColor,line width= 0.6pt,line join=round] (251.70, 37.91) --
	(251.70, 40.66);

\path[draw=drawColor,line width= 0.6pt,line join=round] (293.99, 37.91) --
	(293.99, 40.66);

\path[draw=drawColor,line width= 0.6pt,line join=round] (313.65, 37.91) --
	(313.65, 40.66);
\end{scope}
\begin{scope}
\path[clip] (  0.00,  0.00) rectangle (332.44,108.41);
\definecolor{drawColor}{gray}{0.30}

\node[text=drawColor,rotate= 45.00,anchor=base,inner sep=0pt, outer sep=0pt, scale=  0.80] at ( 51.67, 31.81) {5e+03};

\node[text=drawColor,rotate= 45.00,anchor=base,inner sep=0pt, outer sep=0pt, scale=  0.80] at ( 64.62, 31.81) {7e+03};

\node[text=drawColor,rotate= 45.00,anchor=base,inner sep=0pt, outer sep=0pt, scale=  0.80] at ( 78.35, 31.81) {1e+04};

\node[text=drawColor,rotate= 45.00,anchor=base,inner sep=0pt, outer sep=0pt, scale=  0.80] at (120.64, 31.81) {3e+04};

\node[text=drawColor,rotate= 45.00,anchor=base,inner sep=0pt, outer sep=0pt, scale=  0.80] at (140.30, 31.81) {5e+04};

\node[text=drawColor,rotate= 45.00,anchor=base,inner sep=0pt, outer sep=0pt, scale=  0.80] at (153.25, 31.81) {7e+04};

\node[text=drawColor,rotate= 45.00,anchor=base,inner sep=0pt, outer sep=0pt, scale=  0.80] at (166.98, 31.81) {1e+05};

\node[text=drawColor,rotate= 45.00,anchor=base,inner sep=0pt, outer sep=0pt, scale=  0.80] at (209.26, 31.81) {3e+05};

\node[text=drawColor,rotate= 45.00,anchor=base,inner sep=0pt, outer sep=0pt, scale=  0.80] at (228.92, 31.81) {5e+05};

\node[text=drawColor,rotate= 45.00,anchor=base,inner sep=0pt, outer sep=0pt, scale=  0.80] at (241.87, 31.81) {7e+05};

\node[text=drawColor,rotate= 45.00,anchor=base,inner sep=0pt, outer sep=0pt, scale=  0.80] at (255.60, 31.81) {1e+06};

\node[text=drawColor,rotate= 45.00,anchor=base,inner sep=0pt, outer sep=0pt, scale=  0.80] at (297.88, 31.81) {3e+06};

\node[text=drawColor,rotate= 45.00,anchor=base,inner sep=0pt, outer sep=0pt, scale=  0.80] at (317.54, 31.81) {5e+06};
\end{scope}
\begin{scope}
\path[clip] (  0.00,  0.00) rectangle (332.44,108.41);
\definecolor{drawColor}{RGB}{0,0,0}

\node[text=drawColor,anchor=base,inner sep=0pt, outer sep=0pt, scale=  0.80] at (180.71,  15.06) {sample size};
\end{scope}
\begin{scope}
\path[clip] (  0.00,  0.00) rectangle (332.44,108.41);
\definecolor{drawColor}{RGB}{0,0,0}

\node[text=drawColor,rotate= 90.00,anchor=base,inner sep=0pt, outer sep=0pt, scale=  0.80] at ( 11.01, 71.78) {MISE};
\end{scope}
\end{tikzpicture}

%% file: fig_gauss.tex
\begin{tikzpicture}[x=1pt,y=1pt]
\definecolor{fillColor}{RGB}{255,255,255}
\path[use as bounding box,fill=fillColor,fill opacity=0.00] (0,0) rectangle (346.90,144.54);
\begin{scope}
\path[clip] (  0.00,  0.00) rectangle (346.90,144.54);
\definecolor{drawColor}{RGB}{255,255,255}
\definecolor{fillColor}{RGB}{255,255,255}

\path[draw=drawColor,line width= 0.6pt,line join=round,line cap=round,fill=fillColor] (  0.00,  0.00) rectangle (346.90,144.54);
\end{scope}
\begin{scope}
\path[clip] ( 38.48, 27.33) rectangle (156.96,122.47);
\definecolor{fillColor}{RGB}{255,255,255}

\path[fill=fillColor] ( 38.48, 27.33) rectangle (156.96,122.47);
\definecolor{drawColor}{gray}{0.92}

\path[draw=drawColor,line width= 0.3pt,line join=round] ( 38.48, 30.28) --
	(156.96, 30.28);

\path[draw=drawColor,line width= 0.3pt,line join=round] ( 38.48, 32.66) --
	(156.96, 32.66);

\path[draw=drawColor,line width= 0.3pt,line join=round] ( 38.48, 35.04) --
	(156.96, 35.04);

\path[draw=drawColor,line width= 0.3pt,line join=round] ( 38.48, 37.41) --
	(156.96, 37.41);

\path[draw=drawColor,line width= 0.3pt,line join=round] ( 38.48, 39.79) --
	(156.96, 39.79);

\path[draw=drawColor,line width= 0.3pt,line join=round] ( 38.48, 42.16) --
	(156.96, 42.16);

\path[draw=drawColor,line width= 0.3pt,line join=round] ( 38.48, 44.54) --
	(156.96, 44.54);

\path[draw=drawColor,line width= 0.3pt,line join=round] ( 38.48, 46.91) --
	(156.96, 46.91);

\path[draw=drawColor,line width= 0.3pt,line join=round] ( 38.48, 49.29) --
	(156.96, 49.29);

\path[draw=drawColor,line width= 0.3pt,line join=round] ( 38.48, 54.04) --
	(156.96, 54.04);

\path[draw=drawColor,line width= 0.3pt,line join=round] ( 38.48, 56.42) --
	(156.96, 56.42);

\path[draw=drawColor,line width= 0.3pt,line join=round] ( 38.48, 58.79) --
	(156.96, 58.79);

\path[draw=drawColor,line width= 0.3pt,line join=round] ( 38.48, 61.17) --
	(156.96, 61.17);

\path[draw=drawColor,line width= 0.3pt,line join=round] ( 38.48, 63.55) --
	(156.96, 63.55);

\path[draw=drawColor,line width= 0.3pt,line join=round] ( 38.48, 65.92) --
	(156.96, 65.92);

\path[draw=drawColor,line width= 0.3pt,line join=round] ( 38.48, 68.30) --
	(156.96, 68.30);

\path[draw=drawColor,line width= 0.3pt,line join=round] ( 38.48, 70.67) --
	(156.96, 70.67);

\path[draw=drawColor,line width= 0.3pt,line join=round] ( 38.48, 73.05) --
	(156.96, 73.05);

\path[draw=drawColor,line width= 0.3pt,line join=round] ( 38.48, 77.80) --
	(156.96, 77.80);

\path[draw=drawColor,line width= 0.3pt,line join=round] ( 38.48, 80.18) --
	(156.96, 80.18);

\path[draw=drawColor,line width= 0.3pt,line join=round] ( 38.48, 82.55) --
	(156.96, 82.55);

\path[draw=drawColor,line width= 0.3pt,line join=round] ( 38.48, 84.93) --
	(156.96, 84.93);

\path[draw=drawColor,line width= 0.3pt,line join=round] ( 38.48, 87.30) --
	(156.96, 87.30);

\path[draw=drawColor,line width= 0.3pt,line join=round] ( 38.48, 89.68) --
	(156.96, 89.68);

\path[draw=drawColor,line width= 0.3pt,line join=round] ( 38.48, 92.05) --
	(156.96, 92.05);

\path[draw=drawColor,line width= 0.3pt,line join=round] ( 38.48, 94.43) --
	(156.96, 94.43);

\path[draw=drawColor,line width= 0.3pt,line join=round] ( 38.48, 96.81) --
	(156.96, 96.81);

\path[draw=drawColor,line width= 0.3pt,line join=round] ( 38.48,101.56) --
	(156.96,101.56);

\path[draw=drawColor,line width= 0.3pt,line join=round] ( 38.48,103.93) --
	(156.96,103.93);

\path[draw=drawColor,line width= 0.3pt,line join=round] ( 38.48,106.31) --
	(156.96,106.31);

\path[draw=drawColor,line width= 0.3pt,line join=round] ( 38.48,108.69) --
	(156.96,108.69);

\path[draw=drawColor,line width= 0.3pt,line join=round] ( 38.48,111.06) --
	(156.96,111.06);

\path[draw=drawColor,line width= 0.3pt,line join=round] ( 38.48,113.44) --
	(156.96,113.44);

\path[draw=drawColor,line width= 0.3pt,line join=round] ( 38.48,115.81) --
	(156.96,115.81);

\path[draw=drawColor,line width= 0.3pt,line join=round] ( 38.48,118.19) --
	(156.96,118.19);

\path[draw=drawColor,line width= 0.3pt,line join=round] ( 38.48,120.56) --
	(156.96,120.56);

\path[draw=drawColor,line width= 0.3pt,line join=round] ( 52.84, 27.33) --
	( 52.84,122.47);

\path[draw=drawColor,line width= 0.3pt,line join=round] ( 70.80, 27.33) --
	( 70.80,122.47);

\path[draw=drawColor,line width= 0.3pt,line join=round] ( 88.75, 27.33) --
	( 88.75,122.47);

\path[draw=drawColor,line width= 0.3pt,line join=round] (106.70, 27.33) --
	(106.70,122.47);

\path[draw=drawColor,line width= 0.3pt,line join=round] (124.65, 27.33) --
	(124.65,122.47);

\path[draw=drawColor,line width= 0.3pt,line join=round] (142.60, 27.33) --
	(142.60,122.47);

\path[draw=drawColor,line width= 0.6pt,line join=round] ( 38.48, 27.91) --
	(156.96, 27.91);

\path[draw=drawColor,line width= 0.6pt,line join=round] ( 38.48, 51.67) --
	(156.96, 51.67);

\path[draw=drawColor,line width= 0.6pt,line join=round] ( 38.48, 75.42) --
	(156.96, 75.42);

\path[draw=drawColor,line width= 0.6pt,line join=round] ( 38.48, 99.18) --
	(156.96, 99.18);

\path[draw=drawColor,line width= 0.6pt,line join=round] ( 43.87, 27.33) --
	( 43.87,122.47);

\path[draw=drawColor,line width= 0.6pt,line join=round] ( 61.82, 27.33) --
	( 61.82,122.47);

\path[draw=drawColor,line width= 0.6pt,line join=round] ( 79.77, 27.33) --
	( 79.77,122.47);

\path[draw=drawColor,line width= 0.6pt,line join=round] ( 97.72, 27.33) --
	( 97.72,122.47);

\path[draw=drawColor,line width= 0.6pt,line join=round] (115.68, 27.33) --
	(115.68,122.47);

\path[draw=drawColor,line width= 0.6pt,line join=round] (133.63, 27.33) --
	(133.63,122.47);

\path[draw=drawColor,line width= 0.6pt,line join=round] (151.58, 27.33) --
	(151.58,122.47);
\definecolor{drawColor}{RGB}{248,118,109}
\definecolor{fillColor}{RGB}{248,118,109}

\path[draw=drawColor,line width= 0.4pt,line join=round,line cap=round,fill=fillColor] ( 43.87, 46.77) circle (  1.43);

\path[draw=drawColor,line width= 0.4pt,line join=round,line cap=round,fill=fillColor] ( 61.82, 87.76) circle (  1.43);

\path[draw=drawColor,line width= 0.4pt,line join=round,line cap=round,fill=fillColor] ( 79.77, 63.40) circle (  1.43);

\path[draw=drawColor,line width= 0.4pt,line join=round,line cap=round,fill=fillColor] ( 97.72, 53.19) circle (  1.43);

\path[draw=drawColor,line width= 0.4pt,line join=round,line cap=round,fill=fillColor] (115.68, 45.14) circle (  1.43);

\path[draw=drawColor,line width= 0.4pt,line join=round,line cap=round,fill=fillColor] (133.63, 39.58) circle (  1.43);

\path[draw=drawColor,line width= 0.4pt,line join=round,line cap=round,fill=fillColor] (151.58, 64.56) circle (  1.43);
\definecolor{drawColor}{RGB}{0,186,56}
\definecolor{fillColor}{RGB}{0,186,56}

\path[draw=drawColor,line width= 0.4pt,line join=round,line cap=round,fill=fillColor] ( 43.87, 41.60) circle (  1.43);

\path[draw=drawColor,line width= 0.4pt,line join=round,line cap=round,fill=fillColor] ( 61.82, 76.18) circle (  1.43);

\path[draw=drawColor,line width= 0.4pt,line join=round,line cap=round,fill=fillColor] ( 79.77, 55.15) circle (  1.43);

\path[draw=drawColor,line width= 0.4pt,line join=round,line cap=round,fill=fillColor] ( 97.72, 47.31) circle (  1.43);

\path[draw=drawColor,line width= 0.4pt,line join=round,line cap=round,fill=fillColor] (115.68, 40.68) circle (  1.43);

\path[draw=drawColor,line width= 0.4pt,line join=round,line cap=round,fill=fillColor] (133.63, 36.35) circle (  1.43);

\path[draw=drawColor,line width= 0.4pt,line join=round,line cap=round,fill=fillColor] (151.58, 55.46) circle (  1.43);
\definecolor{drawColor}{RGB}{97,156,255}
\definecolor{fillColor}{RGB}{97,156,255}

\path[draw=drawColor,line width= 0.4pt,line join=round,line cap=round,fill=fillColor] ( 43.87, 34.53) circle (  1.43);

\path[draw=drawColor,line width= 0.4pt,line join=round,line cap=round,fill=fillColor] ( 61.82, 54.52) circle (  1.43);

\path[draw=drawColor,line width= 0.4pt,line join=round,line cap=round,fill=fillColor] ( 79.77, 42.22) circle (  1.43);

\path[draw=drawColor,line width= 0.4pt,line join=round,line cap=round,fill=fillColor] ( 97.72, 37.79) circle (  1.43);

\path[draw=drawColor,line width= 0.4pt,line join=round,line cap=round,fill=fillColor] (115.68, 34.30) circle (  1.43);

\path[draw=drawColor,line width= 0.4pt,line join=round,line cap=round,fill=fillColor] (133.63, 31.90) circle (  1.43);

\path[draw=drawColor,line width= 0.4pt,line join=round,line cap=round,fill=fillColor] (151.58, 41.18) circle (  1.43);
\definecolor{drawColor}{RGB}{248,118,109}

\path[draw=drawColor,line width= 0.6pt,line join=round] ( 43.87, 46.77) --
	( 61.82, 87.76) --
	( 79.77, 63.40) --
	( 97.72, 53.19) --
	(115.68, 45.14) --
	(133.63, 39.58) --
	(151.58, 64.56);
\definecolor{drawColor}{RGB}{0,186,56}

\path[draw=drawColor,line width= 0.6pt,line join=round] ( 43.87, 41.60) --
	( 61.82, 76.18) --
	( 79.77, 55.15) --
	( 97.72, 47.31) --
	(115.68, 40.68) --
	(133.63, 36.35) --
	(151.58, 55.46);
\definecolor{drawColor}{RGB}{97,156,255}

\path[draw=drawColor,line width= 0.6pt,line join=round] ( 43.87, 34.53) --
	( 61.82, 54.52) --
	( 79.77, 42.22) --
	( 97.72, 37.79) --
	(115.68, 34.30) --
	(133.63, 31.90) --
	(151.58, 41.18);
\definecolor{drawColor}{RGB}{255,255,255}
\definecolor{fillColor}{RGB}{248,118,109}

\path[draw=drawColor,line width= 0.6pt,line join=round,line cap=round,fill=fillColor,fill opacity=0.20] ( 43.87, 51.40) --
	( 61.82,101.96) --
	( 79.77, 69.38) --
	( 97.72, 56.93) --
	(115.68, 47.32) --
	(133.63, 40.53) --
	(151.58, 70.12) --
	(151.58, 59.01) --
	(133.63, 38.63) --
	(115.68, 42.97) --
	( 97.72, 49.46) --
	( 79.77, 57.42) --
	( 61.82, 73.56) --
	( 43.87, 42.14) --
	cycle;
\definecolor{fillColor}{RGB}{0,186,56}

\path[draw=drawColor,line width= 0.6pt,line join=round,line cap=round,fill=fillColor,fill opacity=0.20] ( 43.87, 42.75) --
	( 61.82, 82.09) --
	( 79.77, 59.34) --
	( 97.72, 50.22) --
	(115.68, 42.40) --
	(133.63, 37.06) --
	(151.58, 58.00) --
	(151.58, 52.92) --
	(133.63, 35.64) --
	(115.68, 38.95) --
	( 97.72, 44.40) --
	( 79.77, 50.95) --
	( 61.82, 70.28) --
	( 43.87, 40.44) --
	cycle;
\definecolor{fillColor}{RGB}{97,156,255}

\path[draw=drawColor,line width= 0.6pt,line join=round,line cap=round,fill=fillColor,fill opacity=0.20] ( 43.87, 34.92) --
	( 61.82, 57.45) --
	( 79.77, 43.74) --
	( 97.72, 38.79) --
	(115.68, 34.88) --
	(133.63, 32.15) --
	(151.58, 42.13) --
	(151.58, 40.22) --
	(133.63, 31.65) --
	(115.68, 33.71) --
	( 97.72, 36.80) --
	( 79.77, 40.69) --
	( 61.82, 51.60) --
	( 43.87, 34.14) --
	cycle;
\definecolor{drawColor}{gray}{0.20}

\path[draw=drawColor,line width= 0.6pt,line join=round,line cap=round] ( 38.48, 27.33) rectangle (156.96,122.47);
\end{scope}
\begin{scope}
\path[clip] (162.46, 27.33) rectangle (280.95,122.47);
\definecolor{fillColor}{RGB}{255,255,255}

\path[fill=fillColor] (162.46, 27.33) rectangle (280.95,122.47);
\definecolor{drawColor}{gray}{0.92}

\path[draw=drawColor,line width= 0.3pt,line join=round] (162.46, 30.28) --
	(280.95, 30.28);

\path[draw=drawColor,line width= 0.3pt,line join=round] (162.46, 32.66) --
	(280.95, 32.66);

\path[draw=drawColor,line width= 0.3pt,line join=round] (162.46, 35.04) --
	(280.95, 35.04);

\path[draw=drawColor,line width= 0.3pt,line join=round] (162.46, 37.41) --
	(280.95, 37.41);

\path[draw=drawColor,line width= 0.3pt,line join=round] (162.46, 39.79) --
	(280.95, 39.79);

\path[draw=drawColor,line width= 0.3pt,line join=round] (162.46, 42.16) --
	(280.95, 42.16);

\path[draw=drawColor,line width= 0.3pt,line join=round] (162.46, 44.54) --
	(280.95, 44.54);

\path[draw=drawColor,line width= 0.3pt,line join=round] (162.46, 46.91) --
	(280.95, 46.91);

\path[draw=drawColor,line width= 0.3pt,line join=round] (162.46, 49.29) --
	(280.95, 49.29);

\path[draw=drawColor,line width= 0.3pt,line join=round] (162.46, 54.04) --
	(280.95, 54.04);

\path[draw=drawColor,line width= 0.3pt,line join=round] (162.46, 56.42) --
	(280.95, 56.42);

\path[draw=drawColor,line width= 0.3pt,line join=round] (162.46, 58.79) --
	(280.95, 58.79);

\path[draw=drawColor,line width= 0.3pt,line join=round] (162.46, 61.17) --
	(280.95, 61.17);

\path[draw=drawColor,line width= 0.3pt,line join=round] (162.46, 63.55) --
	(280.95, 63.55);

\path[draw=drawColor,line width= 0.3pt,line join=round] (162.46, 65.92) --
	(280.95, 65.92);

\path[draw=drawColor,line width= 0.3pt,line join=round] (162.46, 68.30) --
	(280.95, 68.30);

\path[draw=drawColor,line width= 0.3pt,line join=round] (162.46, 70.67) --
	(280.95, 70.67);

\path[draw=drawColor,line width= 0.3pt,line join=round] (162.46, 73.05) --
	(280.95, 73.05);

\path[draw=drawColor,line width= 0.3pt,line join=round] (162.46, 77.80) --
	(280.95, 77.80);

\path[draw=drawColor,line width= 0.3pt,line join=round] (162.46, 80.18) --
	(280.95, 80.18);

\path[draw=drawColor,line width= 0.3pt,line join=round] (162.46, 82.55) --
	(280.95, 82.55);

\path[draw=drawColor,line width= 0.3pt,line join=round] (162.46, 84.93) --
	(280.95, 84.93);

\path[draw=drawColor,line width= 0.3pt,line join=round] (162.46, 87.30) --
	(280.95, 87.30);

\path[draw=drawColor,line width= 0.3pt,line join=round] (162.46, 89.68) --
	(280.95, 89.68);

\path[draw=drawColor,line width= 0.3pt,line join=round] (162.46, 92.05) --
	(280.95, 92.05);

\path[draw=drawColor,line width= 0.3pt,line join=round] (162.46, 94.43) --
	(280.95, 94.43);

\path[draw=drawColor,line width= 0.3pt,line join=round] (162.46, 96.81) --
	(280.95, 96.81);

\path[draw=drawColor,line width= 0.3pt,line join=round] (162.46,101.56) --
	(280.95,101.56);

\path[draw=drawColor,line width= 0.3pt,line join=round] (162.46,103.93) --
	(280.95,103.93);

\path[draw=drawColor,line width= 0.3pt,line join=round] (162.46,106.31) --
	(280.95,106.31);

\path[draw=drawColor,line width= 0.3pt,line join=round] (162.46,108.69) --
	(280.95,108.69);

\path[draw=drawColor,line width= 0.3pt,line join=round] (162.46,111.06) --
	(280.95,111.06);

\path[draw=drawColor,line width= 0.3pt,line join=round] (162.46,113.44) --
	(280.95,113.44);

\path[draw=drawColor,line width= 0.3pt,line join=round] (162.46,115.81) --
	(280.95,115.81);

\path[draw=drawColor,line width= 0.3pt,line join=round] (162.46,118.19) --
	(280.95,118.19);

\path[draw=drawColor,line width= 0.3pt,line join=round] (162.46,120.56) --
	(280.95,120.56);

\path[draw=drawColor,line width= 0.3pt,line join=round] (176.83, 27.33) --
	(176.83,122.47);

\path[draw=drawColor,line width= 0.3pt,line join=round] (194.78, 27.33) --
	(194.78,122.47);

\path[draw=drawColor,line width= 0.3pt,line join=round] (212.73, 27.33) --
	(212.73,122.47);

\path[draw=drawColor,line width= 0.3pt,line join=round] (230.68, 27.33) --
	(230.68,122.47);

\path[draw=drawColor,line width= 0.3pt,line join=round] (248.63, 27.33) --
	(248.63,122.47);

\path[draw=drawColor,line width= 0.3pt,line join=round] (266.59, 27.33) --
	(266.59,122.47);

\path[draw=drawColor,line width= 0.6pt,line join=round] (162.46, 27.91) --
	(280.95, 27.91);

\path[draw=drawColor,line width= 0.6pt,line join=round] (162.46, 51.67) --
	(280.95, 51.67);

\path[draw=drawColor,line width= 0.6pt,line join=round] (162.46, 75.42) --
	(280.95, 75.42);

\path[draw=drawColor,line width= 0.6pt,line join=round] (162.46, 99.18) --
	(280.95, 99.18);

\path[draw=drawColor,line width= 0.6pt,line join=round] (167.85, 27.33) --
	(167.85,122.47);

\path[draw=drawColor,line width= 0.6pt,line join=round] (185.80, 27.33) --
	(185.80,122.47);

\path[draw=drawColor,line width= 0.6pt,line join=round] (203.75, 27.33) --
	(203.75,122.47);

\path[draw=drawColor,line width= 0.6pt,line join=round] (221.71, 27.33) --
	(221.71,122.47);

\path[draw=drawColor,line width= 0.6pt,line join=round] (239.66, 27.33) --
	(239.66,122.47);

\path[draw=drawColor,line width= 0.6pt,line join=round] (257.61, 27.33) --
	(257.61,122.47);

\path[draw=drawColor,line width= 0.6pt,line join=round] (275.56, 27.33) --
	(275.56,122.47);
\definecolor{drawColor}{RGB}{248,118,109}
\definecolor{fillColor}{RGB}{248,118,109}

\path[draw=drawColor,line width= 0.4pt,line join=round,line cap=round,fill=fillColor] (167.85, 54.73) circle (  1.43);

\path[draw=drawColor,line width= 0.4pt,line join=round,line cap=round,fill=fillColor] (185.80,106.96) circle (  1.43);

\path[draw=drawColor,line width= 0.4pt,line join=round,line cap=round,fill=fillColor] (203.75, 74.36) circle (  1.43);

\path[draw=drawColor,line width= 0.4pt,line join=round,line cap=round,fill=fillColor] (221.71, 61.40) circle (  1.43);

\path[draw=drawColor,line width= 0.4pt,line join=round,line cap=round,fill=fillColor] (239.66, 50.53) circle (  1.43);

\path[draw=drawColor,line width= 0.4pt,line join=round,line cap=round,fill=fillColor] (257.61, 42.72) circle (  1.43);

\path[draw=drawColor,line width= 0.4pt,line join=round,line cap=round,fill=fillColor] (275.56, 76.97) circle (  1.43);
\definecolor{drawColor}{RGB}{0,186,56}
\definecolor{fillColor}{RGB}{0,186,56}

\path[draw=drawColor,line width= 0.4pt,line join=round,line cap=round,fill=fillColor] (167.85, 47.15) circle (  1.43);

\path[draw=drawColor,line width= 0.4pt,line join=round,line cap=round,fill=fillColor] (185.80, 93.36) circle (  1.43);

\path[draw=drawColor,line width= 0.4pt,line join=round,line cap=round,fill=fillColor] (203.75, 64.64) circle (  1.43);

\path[draw=drawColor,line width= 0.4pt,line join=round,line cap=round,fill=fillColor] (221.71, 53.88) circle (  1.43);

\path[draw=drawColor,line width= 0.4pt,line join=round,line cap=round,fill=fillColor] (239.66, 44.85) circle (  1.43);

\path[draw=drawColor,line width= 0.4pt,line join=round,line cap=round,fill=fillColor] (257.61, 38.83) circle (  1.43);

\path[draw=drawColor,line width= 0.4pt,line join=round,line cap=round,fill=fillColor] (275.56, 66.88) circle (  1.43);
\definecolor{drawColor}{RGB}{97,156,255}
\definecolor{fillColor}{RGB}{97,156,255}

\path[draw=drawColor,line width= 0.4pt,line join=round,line cap=round,fill=fillColor] (167.85, 36.88) circle (  1.43);

\path[draw=drawColor,line width= 0.4pt,line join=round,line cap=round,fill=fillColor] (185.80, 64.74) circle (  1.43);

\path[draw=drawColor,line width= 0.4pt,line join=round,line cap=round,fill=fillColor] (203.75, 47.75) circle (  1.43);

\path[draw=drawColor,line width= 0.4pt,line join=round,line cap=round,fill=fillColor] (221.71, 41.67) circle (  1.43);

\path[draw=drawColor,line width= 0.4pt,line join=round,line cap=round,fill=fillColor] (239.66, 36.73) circle (  1.43);

\path[draw=drawColor,line width= 0.4pt,line join=round,line cap=round,fill=fillColor] (257.61, 33.21) circle (  1.43);

\path[draw=drawColor,line width= 0.4pt,line join=round,line cap=round,fill=fillColor] (275.56, 47.27) circle (  1.43);
\definecolor{drawColor}{RGB}{248,118,109}

\path[draw=drawColor,line width= 0.6pt,line join=round] (167.85, 54.73) --
	(185.80,106.96) --
	(203.75, 74.36) --
	(221.71, 61.40) --
	(239.66, 50.53) --
	(257.61, 42.72) --
	(275.56, 76.97);
\definecolor{drawColor}{RGB}{0,186,56}

\path[draw=drawColor,line width= 0.6pt,line join=round] (167.85, 47.15) --
	(185.80, 93.36) --
	(203.75, 64.64) --
	(221.71, 53.88) --
	(239.66, 44.85) --
	(257.61, 38.83) --
	(275.56, 66.88);
\definecolor{drawColor}{RGB}{97,156,255}

\path[draw=drawColor,line width= 0.6pt,line join=round] (167.85, 36.88) --
	(185.80, 64.74) --
	(203.75, 47.75) --
	(221.71, 41.67) --
	(239.66, 36.73) --
	(257.61, 33.21) --
	(275.56, 47.27);
\definecolor{drawColor}{RGB}{255,255,255}
\definecolor{fillColor}{RGB}{248,118,109}

\path[draw=drawColor,line width= 0.6pt,line join=round,line cap=round,fill=fillColor,fill opacity=0.20] (167.85, 57.12) --
	(185.80,118.14) --
	(203.75, 81.00) --
	(221.71, 65.92) --
	(239.66, 53.33) --
	(257.61, 43.94) --
	(275.56, 81.42) --
	(275.56, 72.52) --
	(257.61, 41.49) --
	(239.66, 47.74) --
	(221.71, 56.88) --
	(203.75, 67.72) --
	(185.80, 95.77) --
	(167.85, 52.34) --
	cycle;
\definecolor{fillColor}{RGB}{0,186,56}

\path[draw=drawColor,line width= 0.6pt,line join=round,line cap=round,fill=fillColor,fill opacity=0.20] (167.85, 48.25) --
	(185.80,101.46) --
	(203.75, 69.36) --
	(221.71, 57.16) --
	(239.66, 46.63) --
	(257.61, 39.67) --
	(275.56, 71.17) --
	(275.56, 62.59) --
	(257.61, 37.99) --
	(239.66, 43.07) --
	(221.71, 50.60) --
	(203.75, 59.92) --
	(185.80, 85.27) --
	(167.85, 46.05) --
	cycle;
\definecolor{fillColor}{RGB}{97,156,255}

\path[draw=drawColor,line width= 0.6pt,line join=round,line cap=round,fill=fillColor,fill opacity=0.20] (167.85, 37.28) --
	(185.80, 68.67) --
	(203.75, 49.71) --
	(221.71, 42.98) --
	(239.66, 37.50) --
	(257.61, 33.54) --
	(275.56, 48.81) --
	(275.56, 45.73) --
	(257.61, 32.89) --
	(239.66, 35.96) --
	(221.71, 40.37) --
	(203.75, 45.79) --
	(185.80, 60.80) --
	(167.85, 36.49) --
	cycle;
\definecolor{drawColor}{gray}{0.20}

\path[draw=drawColor,line width= 0.6pt,line join=round,line cap=round] (162.46, 27.33) rectangle (280.95,122.47);
\end{scope}
\begin{scope}
\path[clip] ( 38.48,122.47) rectangle (156.96,139.04);
\definecolor{drawColor}{gray}{0.20}
\definecolor{fillColor}{gray}{0.85}

\path[draw=drawColor,line width= 0.6pt,line join=round,line cap=round,fill=fillColor] ( 38.48,122.47) rectangle (156.96,139.04);
\definecolor{drawColor}{gray}{0.10}

\node[text=drawColor,anchor=base,inner sep=0pt, outer sep=0pt, scale=  0.88] at ( 97.72,127.72) {Independent: $N[\begin{psmallmatrix}0\\ 0\end{psmallmatrix},\begin{psmallmatrix}1 & 0\\ 0 & 1\end{psmallmatrix}]$};
\end{scope}
\begin{scope}
\path[clip] (162.46,122.47) rectangle (280.95,139.04);
\definecolor{drawColor}{gray}{0.20}
\definecolor{fillColor}{gray}{0.85}

\path[draw=drawColor,line width= 0.6pt,line join=round,line cap=round,fill=fillColor] (162.46,122.47) rectangle (280.95,139.04);
\definecolor{drawColor}{gray}{0.10}

\node[text=drawColor,anchor=base,inner sep=0pt, outer sep=0pt, scale=  0.88] at (221.71,127.72) {Dependent: $N[\begin{psmallmatrix}0\\ 0\end{psmallmatrix},\begin{psmallmatrix}1 & 0.5\\ 0.5 & 1\end{psmallmatrix}]$};
\end{scope}
\begin{scope}
\path[clip] (  0.00,  0.00) rectangle (346.90,144.54);
\definecolor{drawColor}{gray}{0.20}

\path[draw=drawColor,line width= 0.6pt,line join=round] ( 43.87, 24.58) --
	( 43.87, 27.33);

\path[draw=drawColor,line width= 0.6pt,line join=round] ( 61.82, 24.58) --
	( 61.82, 27.33);

\path[draw=drawColor,line width= 0.6pt,line join=round] ( 79.77, 24.58) --
	( 79.77, 27.33);

\path[draw=drawColor,line width= 0.6pt,line join=round] ( 97.72, 24.58) --
	( 97.72, 27.33);

\path[draw=drawColor,line width= 0.6pt,line join=round] (115.68, 24.58) --
	(115.68, 27.33);

\path[draw=drawColor,line width= 0.6pt,line join=round] (133.63, 24.58) --
	(133.63, 27.33);

\path[draw=drawColor,line width= 0.6pt,line join=round] (151.58, 24.58) --
	(151.58, 27.33);
\end{scope}
\begin{scope}
\path[clip] (  0.00,  0.00) rectangle (346.90,144.54);
\definecolor{drawColor}{gray}{0.30}

\node[text=drawColor,anchor=base,inner sep=0pt, outer sep=0pt, scale=  0.80] at ( 43.87, 16.87) {PALM};

\node[text=drawColor,anchor=base,inner sep=0pt, outer sep=0pt, scale=  0.80] at ( 61.82, 16.87) {1x};

\node[text=drawColor,anchor=base,inner sep=0pt, outer sep=0pt, scale=  0.80] at ( 79.77, 16.87) {2x};

\node[text=drawColor,anchor=base,inner sep=0pt, outer sep=0pt, scale=  0.80] at ( 97.72, 16.87) {3x};

\node[text=drawColor,anchor=base,inner sep=0pt, outer sep=0pt, scale=  0.80] at (115.68, 16.87) {5x};

\node[text=drawColor,anchor=base,inner sep=0pt, outer sep=0pt, scale=  0.80] at (133.63, 16.87) {10x};

\node[text=drawColor,anchor=base,inner sep=0pt, outer sep=0pt, scale=  0.80] at (151.58, 16.87) {*1x};
\end{scope}
\begin{scope}
\path[clip] (  0.00,  0.00) rectangle (346.90,144.54);
\definecolor{drawColor}{gray}{0.20}

\path[draw=drawColor,line width= 0.6pt,line join=round] (167.85, 24.58) --
	(167.85, 27.33);

\path[draw=drawColor,line width= 0.6pt,line join=round] (185.80, 24.58) --
	(185.80, 27.33);

\path[draw=drawColor,line width= 0.6pt,line join=round] (203.75, 24.58) --
	(203.75, 27.33);

\path[draw=drawColor,line width= 0.6pt,line join=round] (221.71, 24.58) --
	(221.71, 27.33);

\path[draw=drawColor,line width= 0.6pt,line join=round] (239.66, 24.58) --
	(239.66, 27.33);

\path[draw=drawColor,line width= 0.6pt,line join=round] (257.61, 24.58) --
	(257.61, 27.33);

\path[draw=drawColor,line width= 0.6pt,line join=round] (275.56, 24.58) --
	(275.56, 27.33);
\end{scope}
\begin{scope}
\path[clip] (  0.00,  0.00) rectangle (346.90,144.54);
\definecolor{drawColor}{gray}{0.30}

\node[text=drawColor,anchor=base,inner sep=0pt, outer sep=0pt, scale=  0.80] at (167.85, 16.87) {PALM};

\node[text=drawColor,anchor=base,inner sep=0pt, outer sep=0pt, scale=  0.80] at (185.80, 16.87) {1x};

\node[text=drawColor,anchor=base,inner sep=0pt, outer sep=0pt, scale=  0.80] at (203.75, 16.87) {2x};

\node[text=drawColor,anchor=base,inner sep=0pt, outer sep=0pt, scale=  0.80] at (221.71, 16.87) {3x};

\node[text=drawColor,anchor=base,inner sep=0pt, outer sep=0pt, scale=  0.80] at (239.66, 16.87) {5x};

\node[text=drawColor,anchor=base,inner sep=0pt, outer sep=0pt, scale=  0.80] at (257.61, 16.87) {10x};

\node[text=drawColor,anchor=base,inner sep=0pt, outer sep=0pt, scale=  0.80] at (275.56, 16.87) {*1x};
\end{scope}
\begin{scope}
\path[clip] (  0.00,  0.00) rectangle (346.90,144.54);
\definecolor{drawColor}{gray}{0.30}

\node[text=drawColor,anchor=base east,inner sep=0pt, outer sep=0pt, scale=  0.80] at ( 33.53, 25.15) {0.000};

\node[text=drawColor,anchor=base east,inner sep=0pt, outer sep=0pt, scale=  0.80] at ( 33.53, 48.91) {0.005};

\node[text=drawColor,anchor=base east,inner sep=0pt, outer sep=0pt, scale=  0.80] at ( 33.53, 72.67) {0.010};

\node[text=drawColor,anchor=base east,inner sep=0pt, outer sep=0pt, scale=  0.80] at ( 33.53, 96.43) {0.015};
\end{scope}
\begin{scope}
\path[clip] (  0.00,  0.00) rectangle (346.90,144.54);
\definecolor{drawColor}{gray}{0.20}

\path[draw=drawColor,line width= 0.6pt,line join=round] ( 35.73, 27.91) --
	( 38.48, 27.91);

\path[draw=drawColor,line width= 0.6pt,line join=round] ( 35.73, 51.67) --
	( 38.48, 51.67);

\path[draw=drawColor,line width= 0.6pt,line join=round] ( 35.73, 75.42) --
	( 38.48, 75.42);

\path[draw=drawColor,line width= 0.6pt,line join=round] ( 35.73, 99.18) --
	( 38.48, 99.18);
\end{scope}
\begin{scope}
\path[clip] (  0.00,  0.00) rectangle (346.90,144.54);
\definecolor{drawColor}{RGB}{0,0,0}

\node[text=drawColor,anchor=base,inner sep=0pt, outer sep=0pt, scale=  0.80] at (159.71,  7.06) {number of cells};
\end{scope}
\begin{scope}
\path[clip] (  0.00,  0.00) rectangle (346.90,144.54);
\definecolor{drawColor}{RGB}{0,0,0}

\node[text=drawColor,rotate= 90.00,anchor=base,inner sep=0pt, outer sep=0pt, scale=  0.80] at ( 11.01, 74.90) {MISE};
\end{scope}
\begin{scope}
\path[clip] (  0.00,  0.00) rectangle (346.90,144.54);
\definecolor{fillColor}{RGB}{255,255,255}

\path[fill=fillColor] (291.95, 42.19) rectangle (341.40,107.61);
\end{scope}
\begin{scope}
\path[clip] (  0.00,  0.00) rectangle (346.90,144.54);
\definecolor{drawColor}{RGB}{0,0,0}

\node[text=drawColor,anchor=base west,inner sep=0pt, outer sep=0pt, scale=  0.80] at (297.45, 95.83) {sample size};
\end{scope}
\begin{scope}
\path[clip] (  0.00,  0.00) rectangle (346.90,144.54);
\definecolor{fillColor}{RGB}{255,255,255}

\path[fill=fillColor] (297.45, 76.59) rectangle (311.90, 91.05);
\end{scope}
\begin{scope}
\path[clip] (  0.00,  0.00) rectangle (346.90,144.54);
\definecolor{drawColor}{RGB}{248,118,109}
\definecolor{fillColor}{RGB}{248,118,109}

\path[draw=drawColor,line width= 0.4pt,line join=round,line cap=round,fill=fillColor] (304.67, 83.82) circle (  1.43);
\end{scope}
\begin{scope}
\path[clip] (  0.00,  0.00) rectangle (346.90,144.54);
\definecolor{drawColor}{RGB}{248,118,109}

\path[draw=drawColor,line width= 0.6pt,line join=round] (298.89, 83.82) -- (310.46, 83.82);
\end{scope}
\begin{scope}
\path[clip] (  0.00,  0.00) rectangle (346.90,144.54);
\definecolor{drawColor}{RGB}{255,255,255}
\definecolor{fillColor}{RGB}{248,118,109}

\path[draw=drawColor,line width= 0.6pt,line cap=rect,fill=fillColor,fill opacity=0.20] (298.16, 77.31) rectangle (311.19, 90.34);
\end{scope}
\begin{scope}
\path[clip] (  0.00,  0.00) rectangle (346.90,144.54);
\definecolor{fillColor}{RGB}{255,255,255}

\path[fill=fillColor] (297.45, 62.14) rectangle (311.90, 76.59);
\end{scope}
\begin{scope}
\path[clip] (  0.00,  0.00) rectangle (346.90,144.54);
\definecolor{drawColor}{RGB}{0,186,56}
\definecolor{fillColor}{RGB}{0,186,56}

\path[draw=drawColor,line width= 0.4pt,line join=round,line cap=round,fill=fillColor] (304.67, 69.37) circle (  1.43);
\end{scope}
\begin{scope}
\path[clip] (  0.00,  0.00) rectangle (346.90,144.54);
\definecolor{drawColor}{RGB}{0,186,56}

\path[draw=drawColor,line width= 0.6pt,line join=round] (298.89, 69.37) -- (310.46, 69.37);
\end{scope}
\begin{scope}
\path[clip] (  0.00,  0.00) rectangle (346.90,144.54);
\definecolor{drawColor}{RGB}{255,255,255}
\definecolor{fillColor}{RGB}{0,186,56}

\path[draw=drawColor,line width= 0.6pt,line cap=rect,fill=fillColor,fill opacity=0.20] (298.16, 62.85) rectangle (311.19, 75.88);
\end{scope}
\begin{scope}
\path[clip] (  0.00,  0.00) rectangle (346.90,144.54);
\definecolor{fillColor}{RGB}{255,255,255}

\path[fill=fillColor] (297.45, 47.69) rectangle (311.90, 62.14);
\end{scope}
\begin{scope}
\path[clip] (  0.00,  0.00) rectangle (346.90,144.54);
\definecolor{drawColor}{RGB}{97,156,255}
\definecolor{fillColor}{RGB}{97,156,255}

\path[draw=drawColor,line width= 0.4pt,line join=round,line cap=round,fill=fillColor] (304.67, 54.91) circle (  1.43);
\end{scope}
\begin{scope}
\path[clip] (  0.00,  0.00) rectangle (346.90,144.54);
\definecolor{drawColor}{RGB}{97,156,255}

\path[draw=drawColor,line width= 0.6pt,line join=round] (298.89, 54.91) -- (310.46, 54.91);
\end{scope}
\begin{scope}
\path[clip] (  0.00,  0.00) rectangle (346.90,144.54);
\definecolor{drawColor}{RGB}{255,255,255}
\definecolor{fillColor}{RGB}{97,156,255}

\path[draw=drawColor,line width= 0.6pt,line cap=rect,fill=fillColor,fill opacity=0.20] (298.16, 48.40) rectangle (311.19, 61.43);
\end{scope}
\begin{scope}
\path[clip] (  0.00,  0.00) rectangle (346.90,144.54);
\definecolor{drawColor}{RGB}{0,0,0}

\node[text=drawColor,anchor=base west,inner sep=0pt, outer sep=0pt, scale=  0.80] at (315.90, 81.07) {5000};
\end{scope}
\begin{scope}
\path[clip] (  0.00,  0.00) rectangle (346.90,144.54);
\definecolor{drawColor}{RGB}{0,0,0}

\node[text=drawColor,anchor=base west,inner sep=0pt, outer sep=0pt, scale=  0.80] at (315.90, 66.61) {10000};
\end{scope}
\begin{scope}
\path[clip] (  0.00,  0.00) rectangle (346.90,144.54);
\definecolor{drawColor}{RGB}{0,0,0}

\node[text=drawColor,anchor=base west,inner sep=0pt, outer sep=0pt, scale=  0.80] at (315.90, 52.16) {50000};
\end{scope}
\end{tikzpicture}

%% file: main.bbl
\begin{thebibliography}{43}
\providecommand{\natexlab}[1]{#1}
\providecommand{\url}[1]{{#1}}
\providecommand{\urlprefix}{URL }
\expandafter\ifx\csname urlstyle\endcsname\relax
  \providecommand{\doi}[1]{DOI~\discretionary{}{}{}#1}\else
  \providecommand{\doi}{DOI~\discretionary{}{}{}\begingroup
  \urlstyle{rm}\Url}\fi
\providecommand{\eprint}[2][]{\url{#2}}

\bibitem[{Bay(2001)}]{bay2001multivariate}
Bay SD (2001) Multivariate discretization for set mining. Knowledge and
  Information Systems 3(4):491--512

\bibitem[{Biba et~al.(2007)Biba, Esposito, Ferilli, Di~Mauro, and
  Basile}]{biba2007unsupervised}
Biba M, Esposito F, Ferilli S, Di~Mauro N, Basile TMA (2007) Unsupervised
  discretization using kernel density estimation. In: Proceedings of the 20th
  International Joint Conference on Artifical Intelligence, Morgan Kaufmann
  Publishers Inc., San Francisco, CA, USA, IJCAI'07, p 696–701

\bibitem[{Boulle(2004)}]{boulle2004khiops}
Boulle M (2004) Khiops: A statistical discretization method of continuous
  attributes. Machine learning 55(1):53--69

\bibitem[{Boull{\'e}(2006)}]{boulle2006modl}
Boull{\'e} M (2006) Modl: a bayes optimal discretization method for continuous
  attributes. Machine learning 65(1):131--165

\bibitem[{Cuevas et~al.(1997)Cuevas, Fraiman et~al.}]{cuevas1997plug}
Cuevas A, Fraiman R, et~al. (1997) A plug-in approach to support estimation.
  The Annals of Statistics 25(6):2300--2312

\bibitem[{Duong and Hazelton(2003)}]{duong2003plug}
Duong T, Hazelton M (2003) Plug-in bandwidth matrices for bivariate kernel
  density estimation. Journal of Nonparametric Statistics 15(1):17--30

\bibitem[{Duong et~al.(2007)}]{duong2007ks}
Duong T, et~al. (2007) ks: Kernel density estimation and kernel discriminant
  analysis for multivariate data in {R}. Journal of Statistical Software
  21(7):1--16

\bibitem[{Fayyad and Irani(1993)}]{fayyad1993multi}
Fayyad U, Irani K (1993) Multi-interval discretization of continuous-valued
  attributes for classification learning. In: Proceedings of the 13th
  International Joint Conference on Artificial Intelligence (IJCAI-93), pp
  1022--1027

\bibitem[{Ferrandiz and Boull{\'e}(2005)}]{ferrandiz2005multivariate}
Ferrandiz S, Boull{\'e} M (2005) Multivariate discretization by recursive
  supervised bipartition of graph. In: International Workshop on Machine
  Learning and Data Mining in Pattern Recognition, Springer, pp 253--264

\bibitem[{Friedman et~al.(2001)Friedman, Hastie, and
  Tibshirani}]{friedman2001elements}
Friedman J, Hastie T, Tibshirani R (2001) The elements of statistical learning,
  vol~1. Springer series in statistics New York

\bibitem[{Galbrun(2020)}]{galbrun2020minimum}
Galbrun E (2020) The minimum description length principle for pattern mining: A
  survey. arXiv preprint arXiv:200714009

\bibitem[{Gasparini(1996)}]{gasparini1996bayesian}
Gasparini M (1996) Bayesian density estimation via {D}irichlet density
  processes. Journal of Nonparametric Statistics 6(4):355--366

\bibitem[{Gr{\"u}nwald(2004)}]{grunwald2004tutorial}
Gr{\"u}nwald P (2004) A tutorial introduction to the minimum description length
  principle. arXiv preprint math/0406077

\bibitem[{Gr{\"u}nwald and Roos(2019)}]{grunwald2019minimum}
Gr{\"u}nwald P, Roos T (2019) Minimum description length revisited. arXiv
  preprint arXiv:190808484

\bibitem[{Gr{\"u}nwald(2007)}]{grunwald2007minimum}
Gr{\"u}nwald PD (2007) The minimum description length principle. MIT press

\bibitem[{Gupta et~al.(2010)Gupta, Mehrotra, and Mohan}]{gupta2010clustering}
Gupta A, Mehrotra KG, Mohan C (2010) A clustering-based discretization for
  supervised learning. Statistics \& probability letters 80(9-10):816--824

\bibitem[{Hansen and Yu(2001)}]{hansen2001model}
Hansen MH, Yu B (2001) Model selection and the principle of minimum description
  length. Journal of the American Statistical Association 96(454):746--774

\bibitem[{Jin et~al.(2009)Jin, Breitbart, and Muoh}]{jin2009data}
Jin R, Breitbart Y, Muoh C (2009) Data discretization unification. Knowledge
  and Information Systems 19(1):1--29

\bibitem[{Kahle and Wickham(2013)}]{KahleGGmap}
Kahle D, Wickham H (2013) ggmap: Spatial visualization with ggplot2. The R
  Journal 5(1):144--161,
  \urlprefix\url{https://journal.r-project.org/archive/2013-1/kahle-wickham.pdf}

\bibitem[{Kameya(2011)}]{kameya2011time}
Kameya Y (2011) Time series discretization via {MDL}-based histogram density
  estimation. In: 2011 IEEE 23rd International Conference on Tools with
  Artificial Intelligence, IEEE, pp 732--739

\bibitem[{Kang et~al.(2006)Kang, Wang, Liu, Lai, Wang, and Miao}]{kang2006ica}
Kang Y, Wang S, Liu X, Lai H, Wang H, Miao B (2006) An {ICA}-based multivariate
  discretization algorithm. In: International Conference on Knowledge Science,
  Engineering and Management, Springer, pp 556--562

\bibitem[{Kerber(1992)}]{kerber1992chimerge}
Kerber R (1992) Chimerge: Discretization of numeric attributes. In: Proceedings
  of the tenth national conference on Artificial intelligence, pp 123--128

\bibitem[{Kontkanen and Myllym{\"a}ki(2007)}]{kontkanen2007linear}
Kontkanen P, Myllym{\"a}ki P (2007) A linear-time algorithm for computing the
  multinomial stochastic complexity. Information Processing Letters
  103(6):227--233

\bibitem[{Kontkanen and Myllymäki(2007)}]{kontkanen2007mdl}
Kontkanen P, Myllymäki P (2007) {MDL} histogram density estimation. In: Meila
  M, Shen X (eds) Proceedings of the Eleventh International Conference on
  Artificial Intelligence and Statistics, PMLR, Proceedings of Machine Learning
  Research, vol~2, pp 219--226

\bibitem[{Kontkanen et~al.(1997)Kontkanen, Myllym{\"a}ki, Silander, and
  Tirri}]{kontkanen1997bayesian}
Kontkanen P, Myllym{\"a}ki P, Silander T, Tirri H (1997) A {B}ayesian approach
  to discretization. In: Proceedings of the European symposium on intelligent
  techniques, Citeseer

\bibitem[{Kotsiantis and Kanellopoulos(2006)}]{kotsiantis2006discretization}
Kotsiantis S, Kanellopoulos D (2006) Discretization techniques: A recent
  survey. GESTS International Transactions on Computer Science and Engineering
  32(1):47--58

\bibitem[{Kurgan and Cios(2004)}]{kurgan2004caim}
Kurgan LA, Cios KJ (2004) Caim discretization algorithm. IEEE transactions on
  Knowledge and Data Engineering 16(2):145--153

\bibitem[{Kwedlo and Kretowski(1999)}]{kwedlo1999evolutionary}
Kwedlo W, Kretowski M (1999) An evolutionary algorithm using multivariate
  discretization for decision rule induction. In: European Conference on
  Principles of Data Mining and Knowledge Discovery, Springer, pp 392--397

\bibitem[{Liu and Wong(2014)}]{liu2014multivariate}
Liu L, Wong WH (2014) Multivariate density estimation based on adaptive
  partitioning: Convergence rate, variable selection and spatial adaptation.
  arXiv preprint arXiv:14012597

\bibitem[{Lu et~al.(2013)Lu, Jiang, and Wong}]{lu2013multivariate}
Lu L, Jiang H, Wong WH (2013) Multivariate density estimation by {B}ayesian
  sequential partitioning. Journal of the American Statistical Association
  108(504):1402--1410

\bibitem[{Lud and Widmer(2000)}]{lud2000relative}
Lud MC, Widmer G (2000) Relative unsupervised discretization for association
  rule mining. In: European conference on principles of data mining and
  knowledge discovery, Springer, pp 148--158

\bibitem[{Marx et~al.(2021)Marx, Yang, and van Leeuwen}]{marx2021estimating}
Marx A, Yang L, van Leeuwen M (2021) Estimating conditional mutual information
  for discrete-continuous mixtures using multi-dimensional adaptive histograms.
  In: Proceedings of the 2021 SIAM International Conference on Data Mining
  (SDM), SIAM, pp 387--395

\bibitem[{Mehta et~al.(2005)Mehta, Parthasarathy, and Yang}]{mehta2005toward}
Mehta S, Parthasarathy S, Yang H (2005) Toward unsupervised correlation
  preserving discretization. IEEE Transactions on Knowledge and Data
  Engineering 17(9):1174--1185

\bibitem[{Nguyen et~al.(2014)Nguyen, M{\"u}ller, Vreeken, and
  B{\"o}hm}]{nguyen2014unsupervised}
Nguyen HV, M{\"u}ller E, Vreeken J, B{\"o}hm K (2014) Unsupervised
  interaction-preserving discretization of multivariate data. Data Mining and
  Knowledge Discovery 28(5-6):1366--1397

\bibitem[{Pfahringer(1995)}]{pfahringer1995compression}
Pfahringer B (1995) Compression-based discretization of continuous attributes.
  In: Machine Learning Proceedings 1995, Elsevier, pp 456--463

\bibitem[{Ram and Gray(2011)}]{ram2011density}
Ram P, Gray AG (2011) Density estimation trees. In: Proceedings of the 17th ACM
  SIGKDD international conference on Knowledge discovery and data mining, pp
  627--635

\bibitem[{Rissanen(1978)}]{rissanen1978modeling}
Rissanen J (1978) Modeling by shortest data description. Automatica
  14(5):465--471

\bibitem[{Schmidberger and Frank(2005)}]{schmidberger2005unsupervised}
Schmidberger G, Frank E (2005) Unsupervised discretization using tree-based
  density estimation. In: European Conference on Principles of Data Mining and
  Knowledge Discovery, Springer, pp 240--251

\bibitem[{Scott(2015)}]{scott2015multivariate}
Scott DW (2015) Multivariate density estimation: theory, practice, and
  visualization. John Wiley \& Sons

\bibitem[{Scricciolo(2007)}]{scricciolo2007rates}
Scricciolo C (2007) On rates of convergence for bayesian density estimation.
  Scandinavian Journal of Statistics 34(3):626--642

\bibitem[{Van Der~Pas and Rockov{\'a}(2017)}]{van2017bayesian}
Van Der~Pas S, Rockov{\'a} V (2017) Bayesian dyadic trees and histograms for
  regression. arXiv preprint arXiv:170800078

\bibitem[{Yang and Wong(2014)}]{yang2014density}
Yang K, Wong WH (2014) Density estimation via adaptive partition and
  discrepancy control. arXiv preprint arXiv:14041425

\bibitem[{Zhang et~al.(2007)Zhang, Wu, Lu, and Jiang}]{zhang2007discretization}
Zhang XH, Wu J, Lu TJ, Jiang Y (2007) A discretization algorithm based on gini
  criterion. In: 2007 International Conference on Machine Learning and
  Cybernetics, IEEE, vol~5, pp 2557--2561

\end{thebibliography}
